\documentclass{article}

\PassOptionsToPackage{square,numbers,sort}{natbib}
\usepackage[dblblindworkshop, final]{neurips_2025}
\workshoptitle{The First Workshop on Generative and Protective AI for Content Creation}

\usepackage{amsmath}
\usepackage{amssymb}
\usepackage{mathtools}
\usepackage{amsthm}
\usepackage{array}
\usepackage{caption}
\usepackage{algorithm}
\usepackage{algorithmic}
\usepackage{graphicx}
\usepackage{xcolor}
\usepackage{multirow}
\usepackage{amsfonts}
\usepackage{colortbl}
\usepackage{makecell}
\usepackage{float}
\usepackage{diagbox} 

\definecolor{darkgreen}{rgb}{0.0, 0.5, 0.0}
\definecolor{text_red}{RGB}{220,20,60}

\newcommand{\G}[1]{\textcolor{darkgreen}{$\kern 0.15em ^{#1}$}}
\newcommand{\GG}[2]{\textbf{#1}\textcolor{darkgreen}{$\kern 0.15em ^{#2}$}}

\theoremstyle{plain}

\theoremstyle{definition}

\theoremstyle{remark}

\usepackage[textsize=tiny]{todonotes}

\usepackage[utf8]{inputenc} 
\usepackage[T1]{fontenc}    
\usepackage{hyperref}       
\usepackage{url}            
\usepackage{booktabs}       
\usepackage{amsfonts}       
\usepackage{nicefrac}       
\usepackage{microtype}      
\usepackage{xcolor}         
\usepackage{wrapfig}

\usepackage{capt-of}
\usepackage{pifont}
\usepackage{cuted}
\usepackage[capitalise,noabbrev]{cleveref}

\title{FreeBlend: Advancing Concept Blending with Staged Feedback-Driven Interpolation Diffusion}

\author{%
  Yufan Zhou$^{1*}$ \quad Haoyu Shen$^{2*}$ \quad Huan Wang$^3$ \\
  $^1$Harbin Institute of Technology \\
  $^2$University of Science and Technology of China \\
  $^3$Westlake University \\
  $^*$equal contribution
}

\begin{document}

\begingroup
\renewcommand\thefootnote{\textasteriskcentered}
\footnotetext{These authors contributed equally to this work.}
\endgroup

\maketitle
\captionsetup{hypcap=false}  
{\renewcommand\twocolumn[1][]{#1}
\begin{center}
    \centering
    \renewcommand{\arraystretch}{0.05} %
    \vspace{-8mm}
    \begin{tabular}{c}
        \includegraphics[width = 0.98\linewidth]{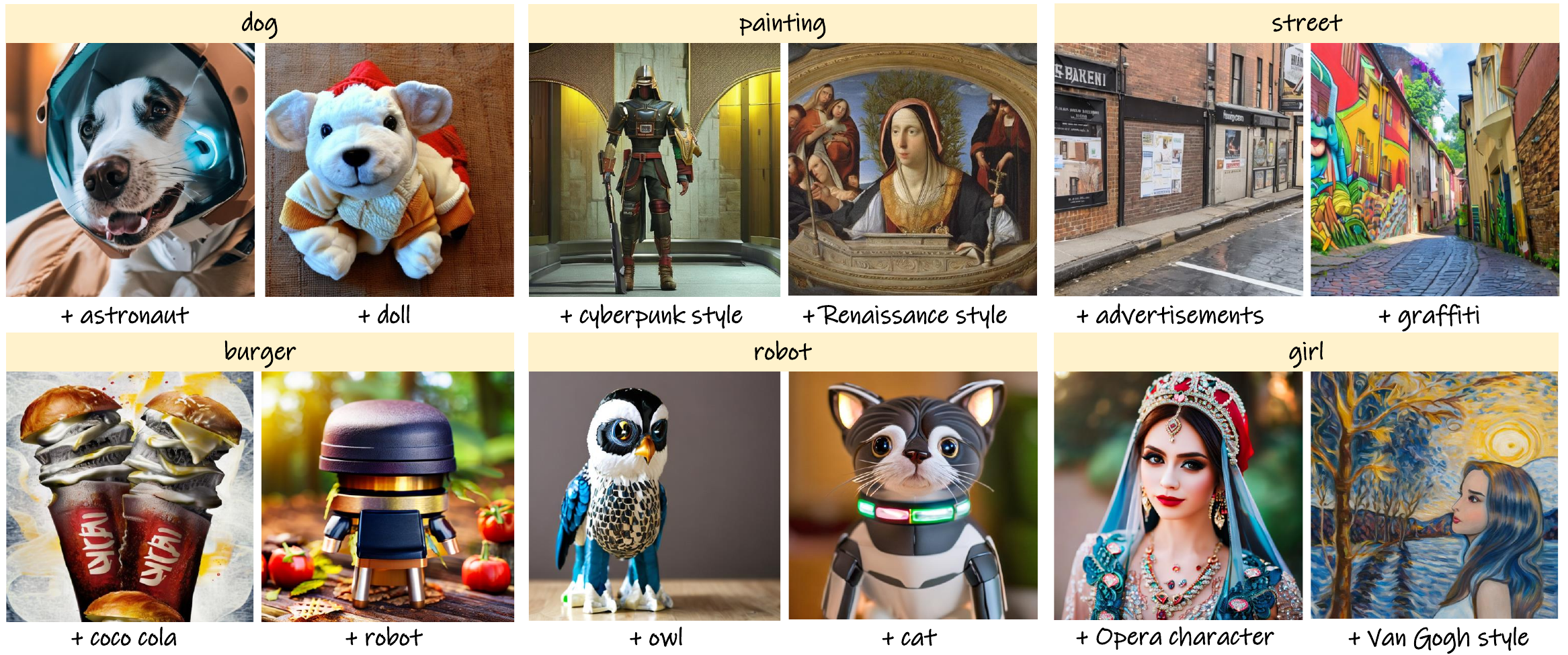}
        \end{tabular}
        \captionof{figure}{We introduce \textbf{FreeBlend}, a training-free approach that effectively blends two concepts to generate new objects through feedback interpolation and auxiliary inference. This method consistently produces visually coherent and harmonious blends, enabling users to create customized images with diverse combinations of concepts.}
    \label{fig:teaser}
\end{center}
}
\begin{abstract}
Have you ever imagined a creature combining the features of a cat and a car? In this work, we tackle the problem known as concept blending, which represents a fascinating yet underexplored area in generative models. While recent approaches, such as embedding mixing and latent modification based on structural sketches, have been proposed, they often suffer from incompatible semantic information and discrepancies in shape and appearance. In this work, we introduce FreeBlend, an effective, training-free framework designed to address these challenges. To mitigate cross-modal loss and enhance feature details, we leverage transferred image embeddings as conditional inputs. We also design a stepwise increasing interpolation strategy between latents to seamlessly integrate auxiliary features. Moreover, we introduce a feedback-driven mechanism that updates the auxiliary latents in reverse order, facilitating global blending and preventing unnatural outputs. Extensive experiments demonstrate that our method significantly improves both the semantic coherence and visual quality of blended images.
The project homepage is available at \url{https://petershen-csworld.github.io/FreeBlend/}.
\end{abstract}

\section{Introduction}

With the breakneck development of diffusion models, the field of image generation has witnessed rapid progress~\citep{rombach2022high, podellsdxl,esser2024scaling,saharia2022photorealistic}. These models exhibit strong instruction-following capabilities and high-fidelity synthesis, achieving remarkable performance across a variety of downstream tasks, including local inpainting~\citep{lugmayr2022repaint, corneanu2024latentpaint}, personalized image generation~\citep{ruiz2023dreambooth, shi2023instantbooth}, and more~\citep{chen2024anydoor, zhao2023uni, li2024photomaker}.

Such advances have paved the way for a spectrum of novel and creative applications. Among them, concept blending stands out as an interesting direction previously overlooked by the community. For example, one could create a creature combining the features of an orange and a teddy bear, resulting in a surprising yet coherent visual concept. More concept pair blends are shown in \Cref{fig:teaser}.

Formally, concept blending involves blending two distinct \textit{text} concepts to create a new one that retains the defining characteristics of its components~\citep{fauconnier1998conceptual, ritchie2004lost}. It can be viewed as a way of integrating elements from different domains into a novel and meaningful output and creating new objects, scenes, or alterations that are coherent and creatively synthesized. Although current models can generate realistic combinations of individual concepts~\citep{liu2022compositional, chefer2023attend}, more sophisticated concept blending techniques are needed to ensure that the generated images not only blend features but also maintain their semantic consistency and visual appeal.

Early methods of concept blending rely on a simple, ``black-box'' approach. These methods achieve coarse concept blending by manipulating text embeddings from the input side, which typically involves adding or reassembling the embeddings~\citep{melzi2023does, olearo2024blend}. However, these approaches often lead to inaccurate representations and a lack of correspondence between visual and semantic features due to cross-modal discrepancies. MagicMix~\citep{liew2022magicmix} interpolates the original class latent into another latent space corresponding to a text prompt. While it creatively introduces latent interpolation for mixing, it struggles with shape-mismatch issues and lacks flexibility in visual transformations. ConceptLab~\citep{richardson2024conceptlab} utilizes VLMs and latent space manipulation, but the constraints imposed during training the additional module limit the flexibility of its application. ATIH~\citep{xiong2024novel}, based on MagicMix, injects trainable parameters into the denoising process and enforces similarity constraints to harmonize the fusing of texts and images. However, the limitations of its model structure, similar to those of MagicMix, combined with its inability to address mismatched shapes or semantically irrelevant features, hinder its overall blending performance. 

In this paper, we introduce FreeBlend, a training-free method for concept blending in image generation. As shown in \Cref{architecture}, FreeBlend smoothly interpolates between auxiliary latents by incorporating a feedback-driven mechanism into the diffusion model's denoising process. As denoising advances, the influence of auxiliary latents diminishes, allowing the blending latent to take greater control, enhancing the blending performance and leveraging the creative power of diffusion models.

Specifically, FreeBlend consists of three core components: transferred unCLIP~\citep{ramesh2022hierarchical} image conditions for Stable Diffusion~\citep{rombach2022high}, a stepwise increasing interpolation strategy, and a feedback-driven mechanism of the denoising process. Instead of using traditional text-based conditions, we employ images generated from text as conditions to guide the generation process via the unCLIP model. This approach reduces the uncertainty caused by cross-modal differences. In addition to this, we divide the denoising process into three stages: the initialization stage, the blending stage, and the refinement stage. At the initialization stage, the pretrained Stable Diffusion model starts with random noise sampled from a standard normal distribution, which is then denoised under the guidance of unCLIP image condition. At the blending stage, we add noise to the auxiliary latents derived from the condition images to ensure to be in the same period as the blending latent. The blending and auxiliary processes are all denoised simultaneously. Ultimately, in the final refinement stage, only the unCLIP image condition is used to provide additional information, enabling the model to generate images with greater clarity and finer details. 

The core contributions of this paper are  (i) We propose a feedback-driven latent interpolation approach that leverages diffusion models to address the concept blending problem. Our method is designed to be training-free and computationally efficient.
(ii) Our approach incorporates unCLIP to use images as conditions, along with a stepwise increasing interpolation strategy and a feedback-driven denoising process to effectively blend different concepts.
(iii) We conduct extensive qualitative and quantitative experiments to validate the effectiveness of our proposed method. Multiple evaluation metrics, including our newly introduced blending score metric, CLIP-BS, confirm that our approach achieves state-of-the-art performance in generating blended concepts.

\section{Related Work}

\textbf{Image Editing and Training-free Guidance. }Generative models like Stable Diffusion~\citep{rombach2022high}, Imagen~\citep{saharia2022photorealistic}, and DALL-E~\citep{ramesh2021zero} have advanced text-to-image synthesis~\citep{podellsdxl, zhang2023adding, gu2022vector, zhou2023shifted} and diffusion models have significantly advanced image editing~\citep{couairon2023zero, couairon2023zero, balaji2022ediff, couairon2022diffedit, brooks2023instructpix2pix, shi2024dragdiffusion}, image interpolation~\citep{wang2023interpolating, qiyuan2024aid}, style transfer~\citep{hamazaspyan2023diffusion, wang2023stylediffusion}, and 3D generation~\citep{xiang2024structured}. DiffEdit~\citep{couairon2022diffedit} and InstructPix2Pix~\citep{brooks2023instructpix2pix} primarily focus on semantic image editing. DiffEdit leverages mask guidance to facilitate intuitive modifications, while InstructPix2Pix enables efficient editing through natural language instructions. In training-free guidance, Structure Diffusion~\citep{feng2022training} introduces structured diffusion guidance to tackle compositional challenges, enhancing the handling of multiple objects. FreeDoM~\citep{yu2023freedom} is a training-free method using energy-guided conditional diffusion for dynamic editing with conditions like segmentation maps. FreeControl~\citep{mo2024freecontrol} offers zero-shot spatial control over pretrained models. These methods enhance the flexibility and precision of diffusion models for effective, retraining-free image editing. Our work builds on this intuition and introduces a novel method for achieving a unique blending style of image editing through training-free guidance.

\textbf{Concept Composition and Blending. }The field of concept composition has made significant advancements, with the combination of multiple concepts to generate complex, multi-faceted images~\citep{liu2021learning, liu2022compositional, wang2024compositional, kumari2023multi, chefer2023attend}. Composable Diffusion~\citep{liu2022compositional} introduces a compositional approach that utilizes multiple models to generate distinct components of an image, effectively addressing challenges associated with complex object compositions and their interrelationships. Custom Diffusion~\citep{kumari2023multi} explores multi-concept customization, enabling the generation of unified, intricate images that seamlessly blend multiple concepts. In contrast, while concept blending holds substantial promise, its practical applications remain more limited and underexplored when compared to compositional approaches. MagicMix~\citep{liew2022magicmix} addresses semantic mixing while preserving spatial layout, though its use is more restricted due to shape limitations. Melzi et al.~\citep{melzi2023does} and Olearo et al.~\citep{olearo2024blend} investigate concept blending by manipulating the relationships between different prompts, with the former focusing on prompt ratios and the latter experimenting with various mechanisms for blending text embeddings. ConceptLab~\citep{richardson2024conceptlab} leverages Diffusion Prior models to generate novel concepts within a category through CLIP-based constraints, enabling the creation of unique hybrid concepts. ATIH~\citep{xiong2024novel} further advances the field by introducing adaptive text-image harmony, merging text and image inputs to generate novel objects while maintaining their original layout. Although these studies contribute to the domain of concept blending, their overall impact has been more limited compared to concept composition, which has inspired the completion of our work.

\section{Method}
\label{method_sec}

\subsection{Preliminaries}

Our method is based on Stable Diffusion, a prominent application of the latent diffusion model~\citep{rombach2022high}. It operates within a compressed latent space, improving sampling and denoising efficiency compared to DDPM~\citep{ho2020denoising} and DDIM~\citep{dhariwal2021diffusion}. A VAE~\citep{kingma2013auto} is used to compress image representations, enhancing computational efficiency. During training, at timestep $t$, a noisy latent vector $\mathbf{z}_t$ is generated by adding noise to the latent representation $\mathbf{z}_0$ (the original clean image latent). This noisy latent is progressively denoised over several timesteps to recover the original image.

Besides, Stable Diffusion employs classifier-free guidance ~\citep{ho2021classifier} to control the synthesized image content with $\mathbf{c}$ representing given additional conditions like text prompts, images, or other customized properties. The model \( \epsilon_\theta \) could be trained via the following objective:
\begin{equation}
\mathcal{L} = \mathbb{E}_{\mathbf{z} \sim \mathcal{E}(\mathbf{x}), \mathbf{c}, \epsilon \sim \mathcal{N}(0, I), t} 
\left[ \left\| \epsilon - \hat{\epsilon}_\theta(\mathbf{z}_t, t, \mathbf{c}) \right\|_2^2 \right],
\end{equation}
where $\epsilon_\theta$, typically implemented by U-Net~\citep{ronneberger2015u} or DiT~\citep{peebles2023scalable}, represents the denoiser.

During inference, a latent \( \mathbf{z}_t \) is sampled from the standard normal distribution \( \mathcal{N}(0, I) \) and the trained denoiser is used to iteratively remove the noise \(\epsilon_t = \hat{\epsilon}_\theta(\mathbf{z}_t, t, \mathbf{c}) \) in \( \mathbf{z}_t \) to produce \( \mathbf{z}_0 \). This process is expressed as:
\begin{equation}
    \hat{\epsilon}_\theta(\mathbf{z}_t, t, \mathbf{c}) \approx -\sigma_t \nabla_{\mathbf{z}_t} \log p_t(\mathbf{z}_t | \mathbf{c}),
\end{equation}
where \( \hat{\epsilon}_\theta(\mathbf{z}_t, t, \mathbf{c}) \) represents the noise removal at timestep \( t \), and the gradient of the log marginal distribution provides the direction for noise reduction.
In the end, the latent \( \mathbf{z}_0 \) is passed to the decoder \( \mathcal{D} \) to generate the output image \( \tilde{\mathbf{x}} \).

\begin{figure*}[t]
\vskip 0.2in
\begin{center}
\centerline{
\includegraphics[width=1.0\textwidth]{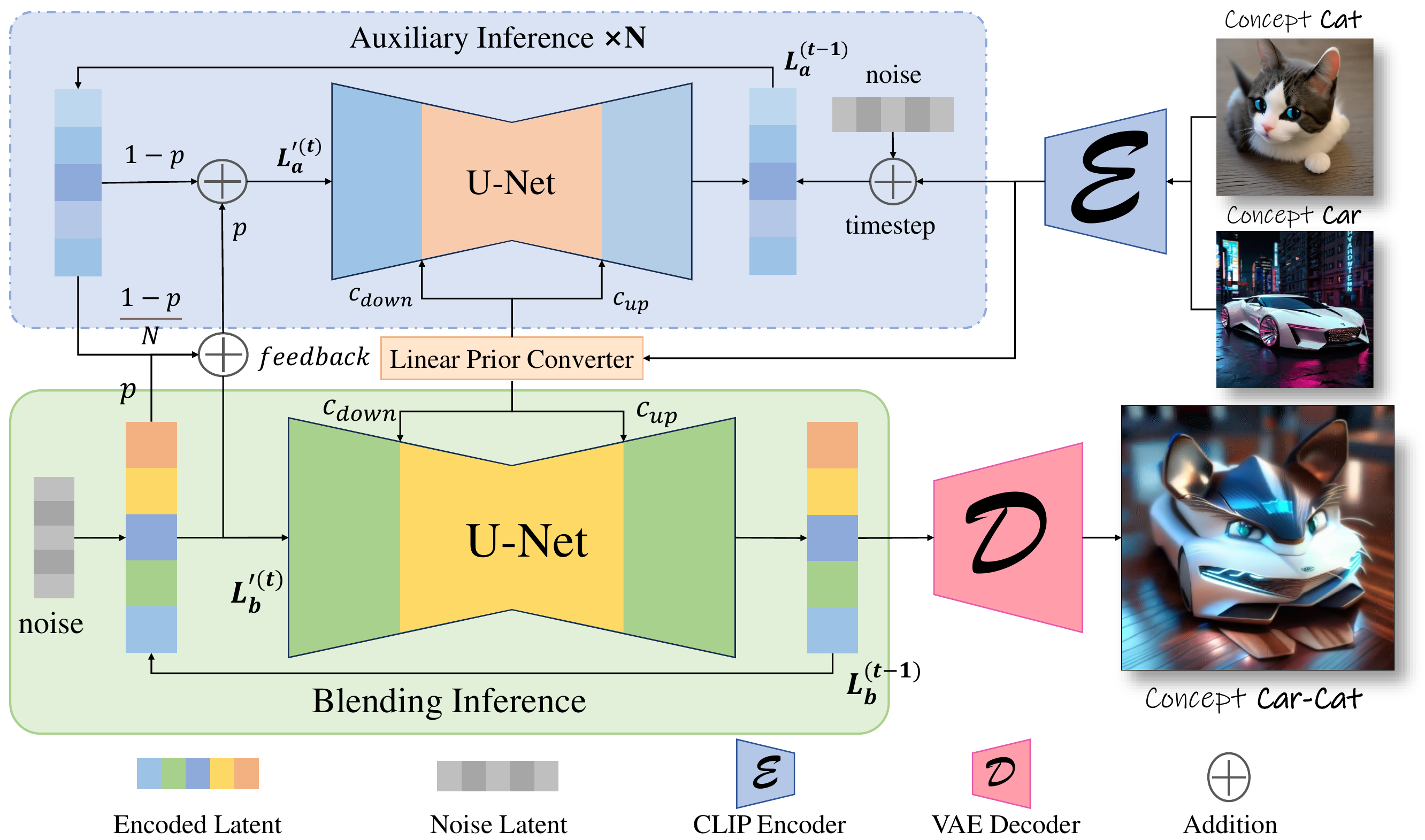}}   
\caption{\textbf{Overview of our method. }Two input images, generated by Stable Diffusion with respective concepts, are encoded into CLIP embeddings and mapped to a shared text space via the Linear Prior Converter from unCLIP~\citep{ramesh2022hierarchical}. These embeddings condition the U-Net, one for downsampling and the other for upsampling. The blending latent $L_b$ is initialized with Gaussian noise and processed during initialization. The module within the dashed box is used only in the blending stage. Noise $\epsilon$ is added to the image embeddings to generate initial auxiliary latents, which are interpolated into $L^{(t)}_b$ for feedback. The latent $L^{(t)}_a$ is combined with $L'^{(t)}_b$ by proportion $p$. Updated latents $L'^{(t)}_a$ are refined in auxiliary inference using unCLIP embeddings to preserve original features, while $L'^{(t)}_b$ is denoised in the blending inference. Finally, the blending latent is refined and passed to the VAE decoder to generate the final image.}
\label{architecture}
\end{center}
\vspace{-12mm}
\end{figure*}


\subsection{Concept Blending with unCLIP Condition}

In multimedia production, the innovation of novel entities is crucial, particularly in the development of anthropomorphic characters. A prevalent method for conceptual innovation involves the fusion of distinct ideas. For instance, numerous animated characters are conceived by combining elements from various sources, such as the "Iron Man" suit and a traditional Japanese samurai's armor. 

In this section, we present a novel image generation task called concept blending. This task aims to merge two distinct text concepts and generate blended images that retain the original shape, color, texture, and other relevant details. To achieve this task, we first attempt to directly use texts as condition~\citep{melzi2023does, olearo2024blend}, but due to the low quality and visual uncertainty due to the gap of vision and text, we decide to choose images as condition~\citep{zhan2024conditional, ye2023ip, mou2024t2i}. 

We adopt the \textit{Linear Prior Converter} from unCLIP~\citep{ramesh2022hierarchical}, which, like DALL-E~\citep{ramesh2021zero}, enables images as conditional inputs. This converter maps image embeddings to the text embedding space, and the resulting embeddings are fed into the U-Net's intermediate layers via cross-attention, guiding the denoising process across all stages. While mapping image embeddings to the text space, the transformed embeddings preserve visual-like information, similar to textual inversion~\citep{gal2022image}, providing more deterministic features and richer details than the original text embeddings.

Our model uses two types of embeddings at different phases: one during the downsampling phase and another during the upsampling phase. Simply averaging the embeddings~\citep{melzi2023does} could lead to the loss of essential details, such as fine textures or spatial positioning. By interacting with the text embedding space through unCLIP, the model captures the semantic essence of the concepts while maintaining precise control over visual attributes like shape and color.

\subsection{Staged Denoising Process}

To enable more flexible and precise control over the image generation process, and inspired by previous works~\citep{liew2022magicmix, lin2024dreamsalon, ackermann2022high}, we propose dividing the generation process into three stages: the initialization stage, the blending stage, and the refinement stage. We introduce two coefficients, \( t_s \) and \( t_e \), to represent the start and end timesteps of the blending stage, respectively.

At the initialization stage, the denoising process begins with random noise, $ \epsilon \sim \mathcal{N}(0, I) $. In this stage, the blending latent is iteratively updated based on the input unCLIP image embeddings, which define the basic layout of the subject and background. The process starts with $ L^{(T)}_{b} = \epsilon $, and the blending latent is updated according to the following equation:
\begin{equation}
L^{(t-1)}_{b} = \epsilon_{\theta}(L^{(t)}_{b}, t, \varnothing) + w \cdot \left[ \epsilon_{\theta}(L^{(t)}_{b}, t, \mathbf{c}_{\text{stage}}) - \epsilon_{\theta}(L_b^{(t)}, t, \varnothing) \right], 
\label{denoise_eq}
\end{equation}
where $w$ denotes the classifier-free guidance~\citep{ho2021classifier} scale, and $\epsilon_{\theta}(L^{(t)}_{b}, t, \mathbf{c})$ represents the pre-trained U-Net model with the conditional input $\mathbf{c}$ and timestep $t$.

In the blending stage, the focus shifts to incorporating original features into the initially formed latent. This process is guided by methods described in~\Cref{sec_stepwise} and \Cref{sec_feed}, which enable the effective blending and filtering of features, ensuring the integration of relevant information.

Finally, at the refinement stage, the general structure of the latent has been established, and the model’s focus moves toward enhancing finer details. The objective here is to improve the overall appearance, making the image more natural and cohesive, while addressing any disjointedness or artifacts from earlier stages. This phase continues to refine and improve the final image output, building on the layout established in the initialization stage.

\subsection{Stepwise Increasing Interpolation Strategy}
\label{sec_stepwise}

In this section, we elaborate on how our method ensures semantic coherence and consistency. We interpolate two image latents with the current denoising latents to encode semantic information from both sources together, which shares a similar intuition with MagicMix~\citep{liew2022magicmix}. The key challenge is to select an appropriate blending ratio at each timestep. One issue of using a constant blending ratio is that we may result in blurring and unclear images, dramatically decreasing the generation quality. To address this, for a denoising process with \( T\) timesteps, at timestep \( t \), we apply a stepwise declining blending ratio $p$, which could be expressed as 
\begin{equation}
p = 1 - \frac{t}{T},
\end{equation}
and the proportion of the $k$-th auxiliary latent $L^{(k)}_a$ is 
\begin{equation}
    \lambda = \frac{1-p}{N},
\end{equation}
where $N$ is the number of concepts needed to blend and we set it two. Another challenge is the potential for biased or catastrophic forgetting of content. Specifically, the model should account for the varying influence of different concepts. For instance, when subject concepts such as ``animals'' are blended with background concepts like ``plants'', the model might give more weight to the subject concepts, resulting in generated images with a higher representation of animals. 

To address this, we introduce an additional hyperparameter, denoted as $\gamma$, which regulates the contribution of each image embedding in the latent space. In the case of a blending task involving $N$ images, the interpolation process can be expressed as follows:
\begin{equation} 
L'^{(t)}_{b} = p \cdot L^{(t)}_{b}  + \lambda \cdot \sum_{n=1}^{N}  \gamma_n  \cdot L_a^{(t,n)},
\end{equation}
where $\gamma_n$ denotes the interpolation strength of $n$-th image, $L^{(t)}_b$ refers to the blending latent, and $L_a^{(t,n)}$ specifies the $n$-th auxiliary latent at the $t$-th timestep. 

\subsection{Feedback-Driven Mechanism}
\label{sec_feed}
In terms of synthesis quality, setting the time proportion of the blending stage too high can lead to overlapping issues, particularly when the blending stage is extended too long. This indicates that the interpolation methods used in previous approaches are unable to handle significant changes, resulting in incomplete processing and causing overlapping effects. In contrast, earlier methods~\citep{liew2022magicmix} terminate the blending stage prematurely, which restricts further blending and fine-tuning. To address this, we propose a feedback-driven mechanism that strikes an optimal balance between these two extremes.

Specifically, $L^{(t_{s},k)}_{a}$ is initialized using the initial auxiliary latent $L^{(0,k)}_{a}$ as follows:
\begin{equation}
    L^{(t_{s},k)}_{a} = \sqrt{\bar{\alpha}_t} L^{(0,k)}_{a} + \sqrt{1-\bar{\alpha}_t} \epsilon,
\end{equation}
where $L^{(0,k)}_{a}$ is the embedding encoded from image. Once $L^{(t)}_b$ is interpolated with $L^{(t,k)}_a$, the latter should simultaneously undergo a feedback update using $L'^{(t)}_b$ as follows:
\begin{equation}
\begin{aligned}
    L'^{(t,k)}_{a} &=  p \cdot L'^{(t)}_{b} + (1-p) \cdot L_a^{(t,k)} \\
    &= \underbrace{(1-p) \cdot L_a^{(t,k)}}_{\text{inherits the original appearance}} + \underbrace{ p^2 \cdot L^{(t)}_{b} }_{\text{integrates into the subject}} + \underbrace{ p \cdot (1-p) \cdot \left[ \frac{1}{N} \cdot  \sum_{n=1}^{N} \gamma_n \cdot L_a^{(t,n)}  \right] }_{\text{maintains balance}}.
\end{aligned}
\label{main_eq}
\end{equation}
After interpolation, $L_a$ should also be denoised normally, with $\mathbf{c}_{\text{stage}}$ being replaced by $\mathbf{c}_{\text{up}}$ or $\mathbf{c}_{\text{down}}$ according to the category of it.

As shown in \Cref{main_eq}, as the timesteps decrease, $p$ increases, which causes $L_b$ to take up a larger proportion of $L_a$. Consequently, in the later stages of blending, the interpolated $L_a$ increasingly resembles $L_b$, thus maintaining consistency over time. The average of all previous $L_a$ values contributes to the incorporation of general features, particularly during the middle stages of blending, due to the coefficient $p \cdot (1-p)$. Regarding $L^{(t,k)}_a$, it gradually declines in the final stages, thereby reducing the proportion of the original features. Through this approach, the auxiliary latents are updated by incorporating both the preceding auxiliary latents and the main blending latent.
\section{Experimental Results}

\subsection{Experimental Settings}
\label{expset_sec}

\textbf{Datasets Construction. }To ensure a fair comparison between text and image-conditioned methods while minimizing the deviation caused by external datasets, we create the Concept Text-Image Reference (CTIR) dataset. This dataset consists of 20 categories, each containing 30 images. The categories have been carefully selected to represent a wide variety of real-world objects, showcasing the model's ability to generate content across different species and types. For evaluation purposes, we develop the Concept Text-Image Blending (CTIB) dataset, which includes 5,700 text-image pairs. These pairs are drawn from a set of 190 text prompts, distributed across the 20 categories, with each category containing 30 images. 

\textbf{Implementation Details. }We use SD-2-1\footnotemark[1] as the backbone model for the baseline comparisons. For the pipeline, we employ SD-2-1-unCLIP\footnotemark[2], which enables image-based conditioning as an input. To maintain a uniform resolution throughout the experiments, all images are resized to $768\times768$ pixels. The experiments are conducted on a node equipped with 4 NVIDIA GeForce RTX A100 GPUs.

\footnotetext[1]{\url{https://huggingface.co/stabilityai/stable-diffusion-2-1}}
\footnotetext[2]{\url{https://huggingface.co/stabilityai/stable-diffusion-2-1-unclip}}

\textbf{Metrics. }Given the unique nature of our task compared to other generation tasks, we use four key metrics to evaluate the blending performance: (1) \textbf{CLIP Blending Similarity (CLIP-BS)} is the primary metric, which measures the CLIP~\citep{radford2021learning} distance and similarity between the blending results and the original concepts. (2) \textbf{DINO Blending Similarity (DINO-BS)}~\citep{liu2025grounding} assesses the detection score of blending objects or the combination of one object with features from another. (3) \textbf{CLIP Image Quality Assessment (CLIP-IQA)}~\citep{wang2023exploring} evaluates the image quality and the degree of match for the blending objects. (4) \textbf{Human Preference Score (HPS)}~\citep{wu2023human} measures human preferences for the blending results based on blending object prompts. Further explanation is provided in \Cref{more_about_metrics}.

\begin{figure*}[ht]
\begin{center}
\centerline{\includegraphics[width=1.0\textwidth]{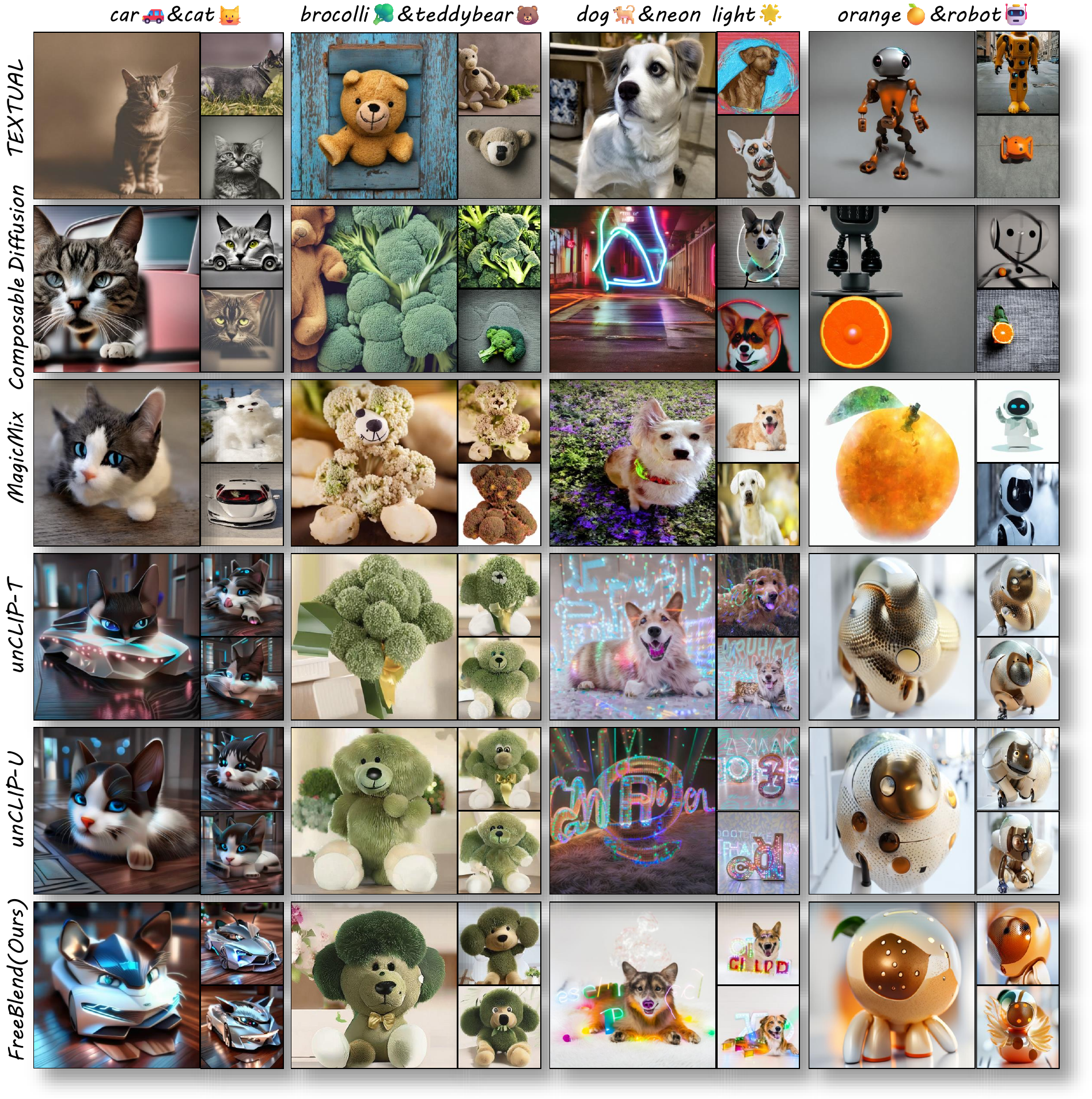}}  
\caption{At the top are the concepts, and on the left are the methods we compare. Each row shows the results, with three images per method and concept pair, evaluating our method against five blending methods. Unlike others, which can suffer from rigid splicing, discordant compositions, or concept bias, our method smoothly integrates features from different concepts into a cohesive new object.}
\label{compare_methods}
\end{center}
\vspace{-8mm}
\end{figure*}

\subsection{Qualitative Comparisons}

The qualitative comparison of our method with five existing approaches applicable to the concept blending task is shown in \Cref{compare_methods}. We observe that interpolated text embeddings~\citep{melzi2023does} tend to favor one category, failing to integrate both concepts effectively. Composable Diffusion~\citep{liu2022compositional} introduces both categories but often results in them co-occurring rather than blending harmoniously. MagicMix~\citep{liew2022magicmix} heavily relies on shape similarity between the reference image and the categories, struggling when the categories are dissimilar, such as ``dog-neon light'' or ``orange-robot''. unCLIP-T and unCLIP-U~\citep{ramesh2022hierarchical} blend the two concepts well, but their blending direction and extent are suboptimal, failing to generate a meaningful blend when the categories lack visual similarity. In contrast, our method effectively merges both categories, producing well-integrated results even with highly distinct concepts like ``car-cat'' or ``dog-neon light''.

\subsection{Quantitative Comparisons}

We evaluate our method on our designed CTIB dataset with four key metrics, with the results presented in \Cref{main_table}. The results demonstrate that our method significantly outperforms all other approaches across all metrics. This indicates that the quality and the blending effect of our synthetic images achieved by our method is the most visually pleasing. 

\begin{table*}[t]
\caption{Quantitative comparisons with other methods. Our method outperforms all other methods across all metrics, including the main blending effect (CLIP-BS), the blending objects detection metric DINO-BS, image quality (CLIP-IQA), and better human preference (HPS).}
\label{main_table}
\vspace{-1mm} 
\begin{center}
\scalebox{1.06}{ 
\begin{small}
\begin{tabular}{lcccc}
\toprule
Methods & CLIP-BS($\uparrow$) & DINO-BS($\uparrow$) & CLIP-IQA($\uparrow$) & HPS($\uparrow$)  \\
\midrule
MagicMix~\citep{liew2022magicmix}& 8.3063\textsubscript{$\pm$ 2.4757} & 
 \cellcolor{orange!40} 0.2485\textsubscript{$\pm$ 0.1575}
& 0.4435\textsubscript{$\pm$ 0.0980}  & 0.2710\textsubscript{$\pm$ 0.0290} \\

Composable Diffusion~\citep{liu2022compositional}    & 6.1374\textsubscript{$\pm$ 1.9454}& 0.2441\textsubscript{$\pm$ 0.1655}& 0.4270\textsubscript{$\pm$ 0.1082} &\cellcolor{orange!40} 0.2903\textsubscript{$\pm$ 0.0277}\\

unCLIP-T~\citep{ramesh2022hierarchical} & \cellcolor{orange!40} 8.7433\textsubscript{$\pm$ 3.1892}	&0.2214\textsubscript{$\pm$ 0.1689} & 0.4436\textsubscript{$\pm$ 0.1073}&0.2384\textsubscript{$\pm$ 0.0285} \\

unCLIP-U~\citep{ramesh2022hierarchical} & 8.7346\textsubscript{$\pm$ 3.1577}&0.2190\textsubscript{$\pm$ 0.1703}&\cellcolor{orange!40} 0.4450\textsubscript{$\pm$ 0.1061}&0.2385\textsubscript{$\pm$ 0.0280}\\

TEXTUAL~\citep{melzi2023does}    & 7.8102\textsubscript{$\pm$ 2.6852} & 0.2366\textsubscript{$\pm$ 0.1537} &0.4164\textsubscript{$\pm$ 0.1143} & 0.2399\textsubscript{$\pm$ 0.0230}\\

UNET~\citep{olearo2024blend}  & 8.5080\textsubscript{$\pm$ 2.7110}& 0.2481\textsubscript{$\pm$ 0.1628} &0.3544\textsubscript{$\pm$ 0.2785} &0.2405\textsubscript{$\pm$ 0.0326} \\

AID~\citep{qiyuan2024aid} & 7.0438\textsubscript{$\pm$ 3.0395}& 0.2355\textsubscript{$\pm$ 0.1655} &0.4213\textsubscript{$\pm$ 0.1132} &0.2551\textsubscript{$\pm$ 0.0309} \\

\textbf{FreeBlend(Ours)} & \cellcolor{red!40} \textbf{9.1555\textsubscript{$\pm$ 2.7134}}& \cellcolor{red!40} \textbf{0.2743\textsubscript{$\pm$ 0.1586}}& \cellcolor{red!40} \textbf{0.5238\textsubscript{$\pm$ 0.0975}}& \cellcolor{red!40} \textbf{0.2932\textsubscript{$\pm$ 0.0316}}\\

\bottomrule
\end{tabular}
\end{small}
}
\end{center}
\vspace{-6mm}  
\end{table*}

\begin{table}[t]
\caption{Ablation study with different interpolation strategies. The results demonstrate that our increase strategy outperforms both the invariant and decline strategies. }
\label{ablation2}
\begin{center}
\scalebox{1.0}{
\begin{small}
\begin{tabular}{ p{0.07\textwidth}p{0.07\textwidth}p{0.07\textwidth}cccc}

\toprule
Increase & Invariant & Decline & \multicolumn{1}{c}{CLIP-BS($\uparrow$)}  & \multicolumn{1}{c}{DINO-BS($\uparrow$)}  & \multicolumn{1}{c}{CLIP-IQA($\uparrow$)} & \multicolumn{1}{c}{HPS($\uparrow$)} \\ 
\midrule
   \centering \checkmark &  &  & \cellcolor{red!40} \textbf{9.1555\textsubscript{$\pm$ 2.7134}} & \cellcolor{red!40} \textbf{0.2743\textsubscript{$\pm$ 0.1586}} & \cellcolor{red!40} \textbf{0.5238\textsubscript{$\pm$ 0.0975}} & \cellcolor{red!40} \textbf{0.2932\textsubscript{$\pm$ 0.0316}} \\
   &  \centering \checkmark& & 7.8891\textsubscript{$\pm$ 2.8732} & 0.1970\textsubscript{$\pm$ 0.1489} &  0.4861\textsubscript{$\pm$ 0.0990} &  \cellcolor{orange!40} 0.2769\textsubscript{$\pm$ 0.0264} \\
   &  & \centering \checkmark&  \cellcolor{orange!40} 8.5355\textsubscript{$\pm$ 3.0644} &  \cellcolor{orange!40} 0.2222\textsubscript{$\pm$ 0.1622} & \cellcolor{orange!40} 0.4981\textsubscript{$\pm$ 0.0836} &  0.2712\textsubscript{$\pm$ 0.0308} \\
\bottomrule
\end{tabular}
\end{small}
}
\end{center}
\vspace{-4mm} 
\end{table}

\begin{table}[t]
\vspace{-2mm}
\caption{Ablation study with staged denoising process. The results demonstrate that our staged denoising process effectively enhances both the blending quality and the scores. Both the initialization and refinement stages contribute to the overall improvement.}
\label{ablation_three_stages}
\begin{center}
\scalebox{0.95}{
\begin{small}
\begin{tabular}{cccccccc}
\toprule
Initialization & Blending & Refinement & \multicolumn{1}{c}{CLIP-BS($\uparrow$)}  & \multicolumn{1}{c}{DINO-BS($\uparrow$)} & \multicolumn{1}{c}{CLIP-IQA($\uparrow$)} & \multicolumn{1}{c}{HPS($\uparrow$)} \\ 
\midrule
  &   \centering \checkmark & & 8.3716\textsubscript{$\pm$ 2.8559} & 0.2520\textsubscript{$\pm$ 0.1542} & \cellcolor{red!40}  \textbf{0.5288\textsubscript{$\pm$ 0.0965}}  & 8.9552\textsubscript{$\pm$ 2.7259}  \\
  &   \centering \checkmark &   \centering \checkmark & 8.3382\textsubscript{$\pm$ 2.8628} & \cellcolor{orange!40}  0.2655\textsubscript{$\pm$ 0.1562} & \cellcolor{orange!40}  0.5282\textsubscript{$\pm$ 0.0934}  & 8.9832\textsubscript{$\pm$ 2.8083}  \\
  \centering \checkmark  &  \centering \checkmark &  & \cellcolor{orange!40} 8.5026\textsubscript{$\pm$ 2.8505} & 0.2617\textsubscript{$\pm$ 0.1501} & 0.5222\textsubscript{$\pm$ 0.1006} & \cellcolor{orange!40} 9.1313\textsubscript{$\pm$ 2.6941} \\
  \centering \checkmark & \centering \checkmark & \centering \checkmark & \cellcolor{red!40} \textbf{9.1555\textsubscript{$\pm$ 2.7134}} & \cellcolor{red!40} \textbf{0.2743\textsubscript{$\pm$ 0.1586}} & 0.5238\textsubscript{$\pm$ 0.0975} & \cellcolor{red!40} \textbf{9.1555\textsubscript{$\pm$ 2.7134}} \\
\bottomrule
\end{tabular}
\end{small}
}
\end{center}
\vspace{-6mm}
\end{table}

\begin{figure*}[ht]
\vskip -0.1in
\begin{center}
\includegraphics[width=1.0\textwidth]{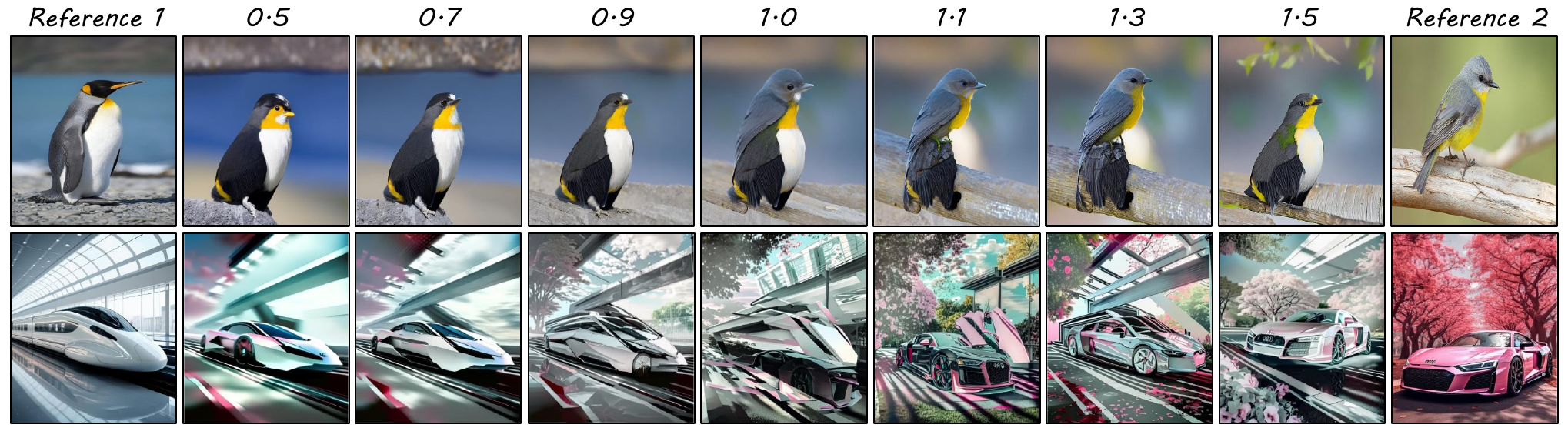}  
\caption{Ablation study on the impact of \( \gamma \). The results show that, in the first row, better blending is achieved on the left side, while the right side appears more spliced. In the second row, both sides exhibit a more visually appealing effect. The blending process, however, is inherently subjective, and users can adjust the parameter \( \gamma \) to tailor the output according to their preferences. By adjusting \( \gamma \), users can control the contribution of each concept, thereby mitigating associated biases.}
\label{gamma_images}
\end{center}
\vskip -0.21in
\end{figure*}

\subsection{Ablation Study}

\textbf{Mitigation of bias. }Different concepts inherently exhibit biases in Stable Diffusion, likely due to the training datasets and their relevance to specific concepts or content. So we introduce an additional parameter, $\gamma$, to reduce the bias toward specific categories in the generated output image. As shown in \Cref{gamma_images}, we vary $\gamma$ within the range $[0.5,1.5]$. Our observations show that as \( \gamma \) moves from 0 to 2, the generated image gradually transitions from resembling reference 1 to favoring reference 2. This provides us with the flexibility to adjust the interpolation strength, allowing for precise control over the image's characteristics and enabling the achievement of optimal results. 

\textbf{Interpolation Strategy. }\Cref{ablation2} illustrates the impact of different interpolation strategies. In the blending stage, we vary the blending ratio between the interpolated auxiliary latents and the blending latent. From this table, we observe that the ``increase'' strategy yields a higher CLIP-BS and CLIP-IQA score. As $p$ increases, the blending latent gains more weight relative to the auxiliary latents, resulting in the final output exhibiting features of both kinds. However, the ``invariant'' method leads to a rigid adjustment, yielding the lowest performance. On the other hand, the ``decline'' strategy prioritizes the auxiliary latent space and avoids blending, which clearly results in ineffective blending.

\textbf{Staged Denoising Process. }We validate the impact of our staged denoising process in \Cref{ablation_three_stages}. The results demonstrate that both the initialization and refinement stages enhance the stability and quality of the generation. This highlights the effectiveness of the initialization stage in shaping the structure of the initial latents, while the refinement stage successfully adds finer details to the images.

\begin{wrapfigure}{r}{0.65\textwidth} 
\vspace{-6mm}
\begin{center}
\includegraphics[width=0.65\textwidth]{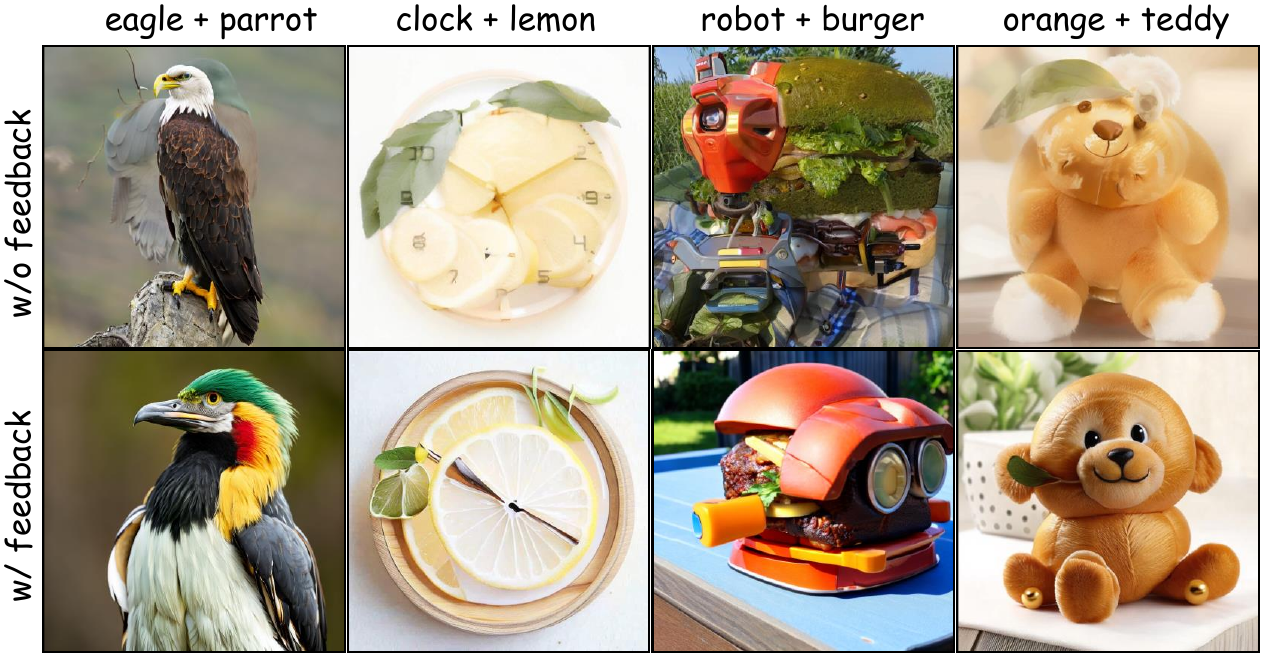}  
\caption{Ablation study of the feedback mechanism: removing the feedback module causes image overlap, disrupting blending and preventing integration.}
\label{feedback_images}
\end{center}
\vspace{-7mm}
\end{wrapfigure}

\textbf{Feedback-Driven Mechanism. }We conduct an ablation study in \Cref{ablation_feedback} to validate the effect of the feedback-driven mechanism. The results demonstrate the impact of our feedback mechanism in blending different concepts. The DINO-BS score, however, is not optimal, which may be due to the surprising overlap of blended objects, thus boosting the score. We also present an intuitive visual comparison in \Cref{feedback_images}, where our effect is clearly significant.

\subsection{User Study} 

We also conduct a user study to evaluate human preference for the blending results with images generated by different methods. Our survey comprises 50 sets of blending pairs. With the active participation of 26 volunteers, we successfully collected a total of 1300 votes. The results are displayed in \Cref{user_study_comparison}. These findings not only affirm the effectiveness of our approach but also provide valuable insights into user preferences in the context of concept blending.

\begin{table*}[ht]
\vspace{-0.5mm}
\caption{User study results comparing FreeBlend with three other methods. Higher values indicate better user preference.}
\label{user_study_comparison}
\begin{center}
\scalebox{0.95}{
\begin{small}
\begin{tabular}{lccccc}
\toprule
Method & MagicMix & TEXTUAL & Composable Diffusion & FreeBlend(Ours) & Total \\
\midrule
User Votes ($\uparrow$) & 49(3.77\%) & 42(3.23\%) & 127(9.77\%) & \textbf{1082(83.2\%)} & 1300 \\
\bottomrule
\end{tabular}
\end{small}
}
\end{center}
\label{user_study}
\vspace{-6mm}
\end{table*}

\section{Conclusion}

Blending multiple concepts into a single object based on diffusion models is both interesting and valuable for 2D content creation. However, it is quite challenging, as it requires controlling the diffusion model to generate objects that have never been seen before. In this paper, we introduce \textit{FreeBlend}, a \textit{training-free} method to blend concepts with stark inherent semantic dissimilarities. At its core, FreeBlend employs the unCLIP model with images as conditions and utilizes a novel proposed interpolation strategy with feedback. Extensive qualitative and quantitative studies demonstrate that our method effectively blends disparate concepts, significantly surpassing the prior SoTA counterparts. Moving forward, exploring alternative ways to address the preference dynamics between different concepts may further improve the performance.

\bibliographystyle{IEEEtran}
\bibliography{reference}

\newpage
\appendix
\onecolumn

\section{T-SNE Analysis}
\label{t_SNE_analysis_paragraph}

In this section, we explore the CLIP embedding space of blended images. Given the high dimensionality of the embeddings, we apply t-SNE~\citep{van2008visualizing} for dimensionality reduction and visualize the embeddings in 2D to illustrate their spatial relationships with the corresponding concepts.  
\begin{figure*}[ht]
\begin{center}
\begin{minipage}{0.48\textwidth}
    \centering
    \includegraphics[width=\linewidth, height=4.6cm]{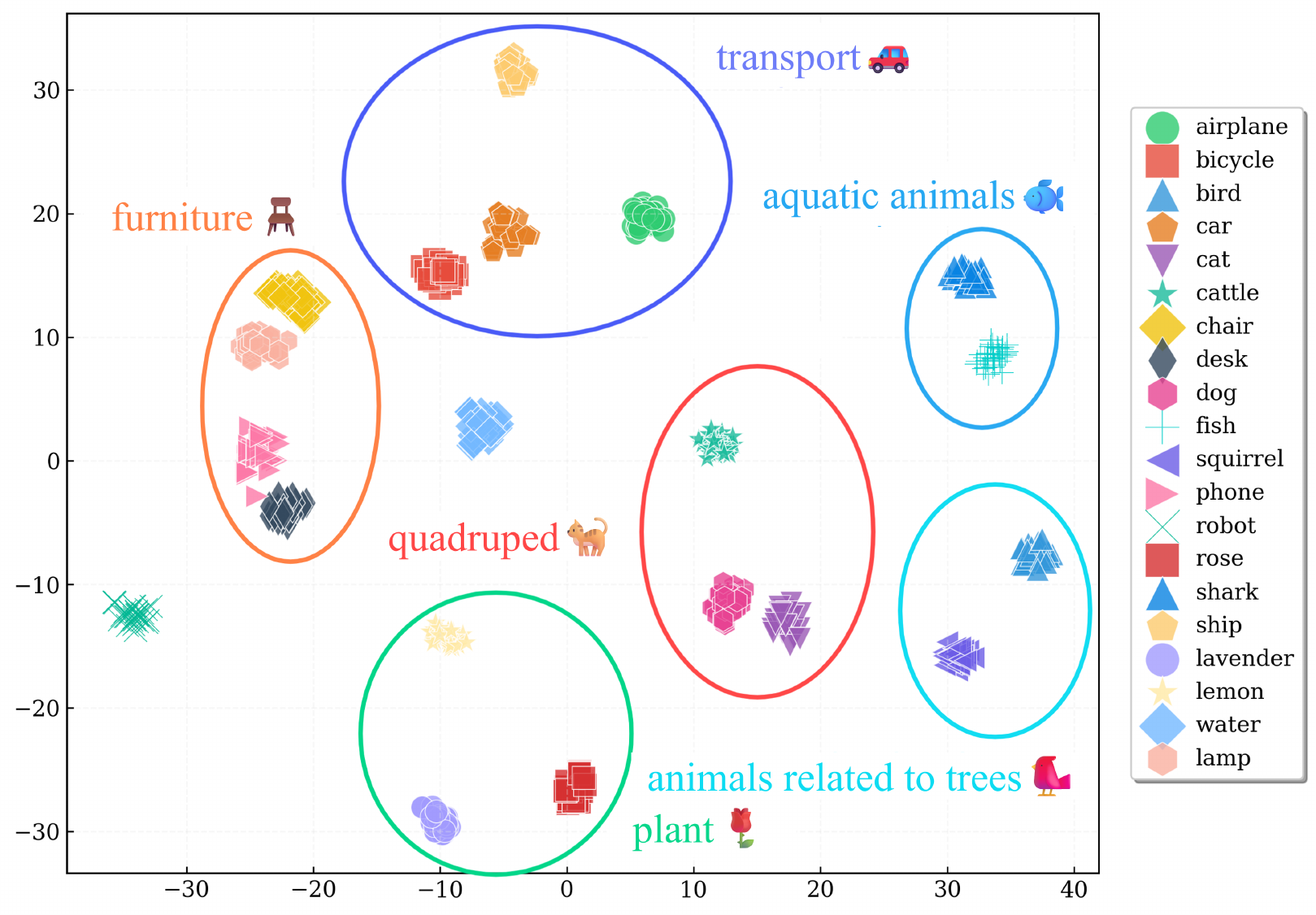}
    \caption{Visualization of the CLIP feature distribution for various image categories after dimensionality reduction via t-SNE.}
    \label{sne_a}
\end{minipage}
\hfill
\begin{minipage}{0.48\textwidth}
    \centering
    \includegraphics[width=\linewidth, height=4.6cm]{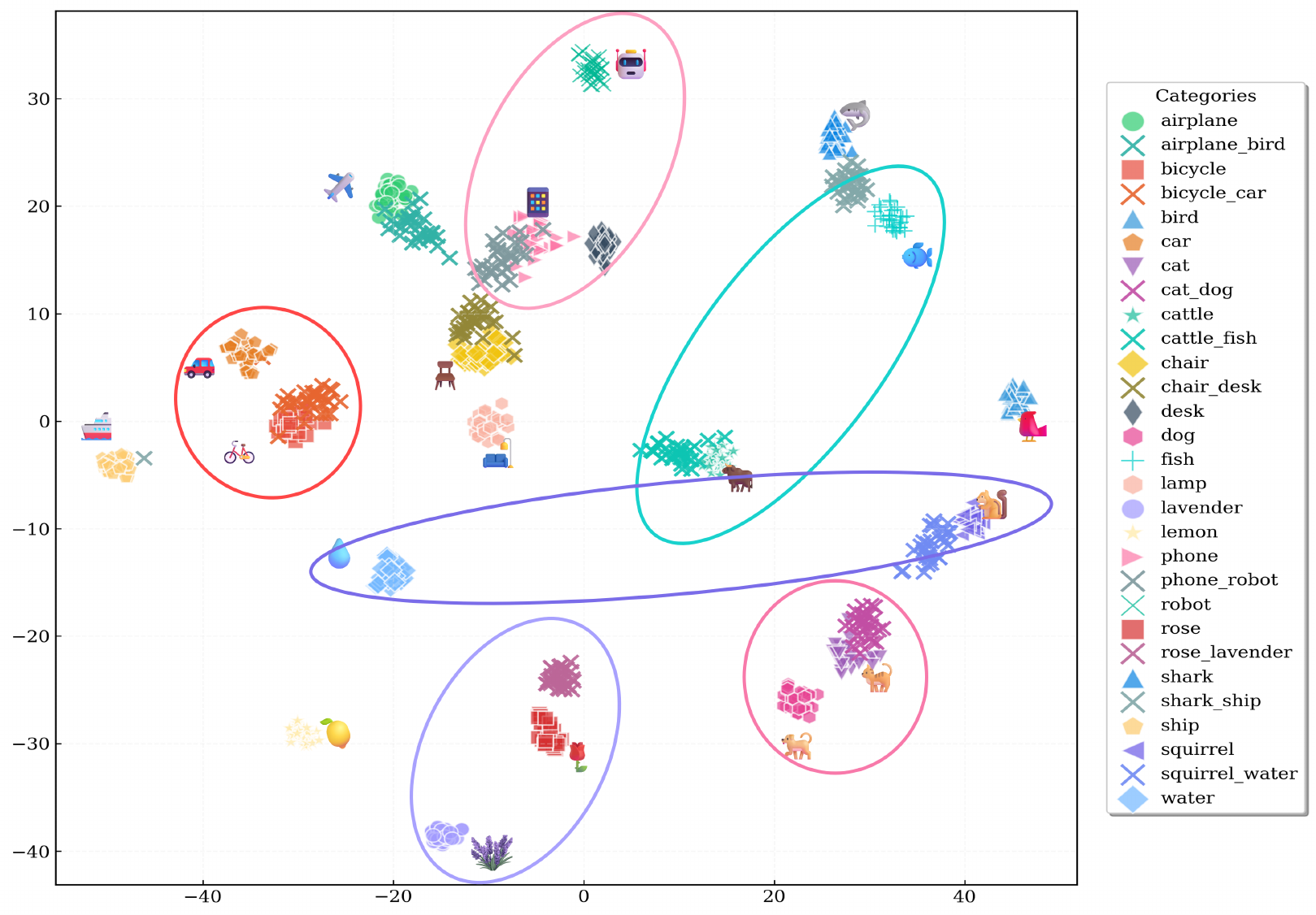}
    \caption{Visualization of the CLIP feature distribution for original class images and blending class images after dimensionality reduction via t-SNE.}
    \label{sne_b}
\end{minipage}
\end{center}
\vspace{-5mm}
\end{figure*}

In \Cref{sne_a}, it is evident that semantically related categories tend to be positioned closer to each other, while unrelated categories are spaced farther apart. This is reasonable given the contrastive training objective of CLIP. In \Cref{sne_b}, we observe that the blending images exhibit a strong relationship with the original classes. Notably, these blending images are generally biased towards one of the original classes, which reflects the inherent bias in the concept blending task. 

Specifically, we notice that the blending images are not located at the midpoint between the original classes. We hypothesize that this observation can be attributed to the following factors: a) t-SNE is a non-linear dimensionality reduction technique that aims to map similar points in high-dimensional space to nearby points in low-dimensional space, while preserving local structures rather than global relationships. Consequently, the position of points in the t-SNE plot may not necessarily correspond to the relative position of categories in the original feature space. b) The feature vector of the blended class may not be positioned directly between the two original classes. This could be due to: i) The blended class containing a mixture of features from original classes, with the resulting feature vector being closer to one of the original classes in the feature space. ii) The blended feature may have a direction in the feature space that is relatively distant from both original classes. In certain cases, this could cause the blended class's feature vector to extend further from the center, leading to its placement behind the original classes in the t-SNE plot, rather than being positioned between them.

\section{Comparison on TIF Dataset}

To provide a convincing comparison with other image-conditioned methods, we conduct experiments on the TIF dataset~\citep{xiong2024novel}. \Cref{main_table_TIF} demonstrates that our method consistently achieves superior results.

\begin{table*}[ht]

\caption{Quantitative comparisons with other image-conditioned methods on TIF dataset.}
\label{main_table_TIF}
\begin{center}
\scalebox{1}{
\begin{small}
\begin{tabular}{lcccc}
\toprule
Methods & CLIP-BS($\uparrow$) & DINO-BS($\uparrow$) & CLIP-IQA($\uparrow$) & HPS($\uparrow$)  \\
\midrule
unCLIP-T~\citep{ramesh2022hierarchical}&11.2156\textsubscript{$\pm$ 4.3520}&\cellcolor{red!40}0.2277\textsubscript{$\pm$ 0.0415}&0.4162\textsubscript{$\pm$ 0.1594}&0.2020\textsubscript{$\pm$ 0.1954}\\

unCLIP-U~\citep{ramesh2022hierarchical}&\cellcolor{orange!40}11.2767\textsubscript{$\pm$ 4.1601}&0.1687\textsubscript{$\pm$ 0.1889}&\cellcolor{orange!40}0.4399\textsubscript{$\pm$ 0.1611}&0.2163\textsubscript{$\pm$ 0.0445}\\

AID~\citep{qiyuan2024aid} & 11.2288\textsubscript{$\pm$ 3.6646}& 0.1378\textsubscript{$\pm$ 0.1792} &0.4378\textsubscript{$\pm$ 0.1630} &\cellcolor{orange!40} 0.2191\textsubscript{$\pm$ 0.0438} \\

\textbf{FreeBlend(Ours)} & \cellcolor{red!40} \textbf{11.9740\textsubscript{$\pm$ 4.0038}}&  \cellcolor{orange!40}\textbf{0.1784\textsubscript{$\pm$ 0.1715}}& \cellcolor{red!40} \textbf{0.5029\textsubscript{$\pm$ 0.1663}}& \cellcolor{red!40} \textbf{0.2500\textsubscript{$\pm$ 0.0465}}\\

\bottomrule
\end{tabular}
\end{small}
}
\end{center}
\vspace{-6mm}  
\end{table*}

\section{More Ablation Study}
\textbf{Conditions for Denoising Process. }\Cref{ablation1} presents the results of our ablation study under different conditions. For text conditions, we use the TEXTUAL and UNET text prompt blending methods for comparison. For image conditions, we apply the unCLIP method along with either unCLIP-T (TEXTUAL-like) or unCLIP-U (UNET-like) image blending methods. We also explore the method of simultaneously using text and images. The results demonstrate that the unCLIP-U method yields the best performance compared to other conditions, while text-based conditions tend to reduce the effectiveness of the generation.
\begin{table}[ht]
\vspace{-2mm} 
\caption{Ablation study for our method with different conditions. We find that the unCLIP-U method, which incorporates two embeddings—one for downsampling and the other for upsampling—achieves the best blending results.}
\label{ablation1}
\begin{center}
\scalebox{1.0}{
\begin{small}
\begin{tabular}{ p{0.08\textwidth}p{0.05\textwidth}p{0.1\textwidth}p{0.1\textwidth}c}
\toprule
TEXTUAL & UNET & unCLIP-T & unCLIP-U & \multicolumn{1}{c}{CLIP-BS($\uparrow$)} \\ 
\midrule
&  & \centering \checkmark  &  &  8.5577\textsubscript{$\pm$ 2.6566} \\

&  &  & \centering \checkmark  & \cellcolor{red!40} \textbf{9.1555\textsubscript{$\pm$ 2.7134}}  \\

\centering \checkmark &  & & & 7.5187\textsubscript{$\pm$ 2.5852}  \\

\centering \checkmark &  &\centering \checkmark & & 7.9888\textsubscript{$\pm$ 2.5823} \\
  
\centering \checkmark &  & & \centering \checkmark& 8.8170\textsubscript{$\pm$ 2.7085}\\

& \centering \checkmark & & & 7.4068\textsubscript{$\pm$ 2.5855}  \\
   
& \centering \checkmark &\centering \checkmark & & 6.9255\textsubscript{$\pm$ 2.4354}  \\
   
& \centering \checkmark & & \centering \checkmark & \cellcolor{orange!40} 9.0260\textsubscript{$\pm$ 2.7408}\\
   
\bottomrule
\end{tabular}
\end{small}
}
\end{center}
\vspace{-7mm} 
\end{table}

\begin{figure*}[ht]
\vskip 0.2in
\begin{center}
\includegraphics[width=0.98\textwidth, height=8.3cm]{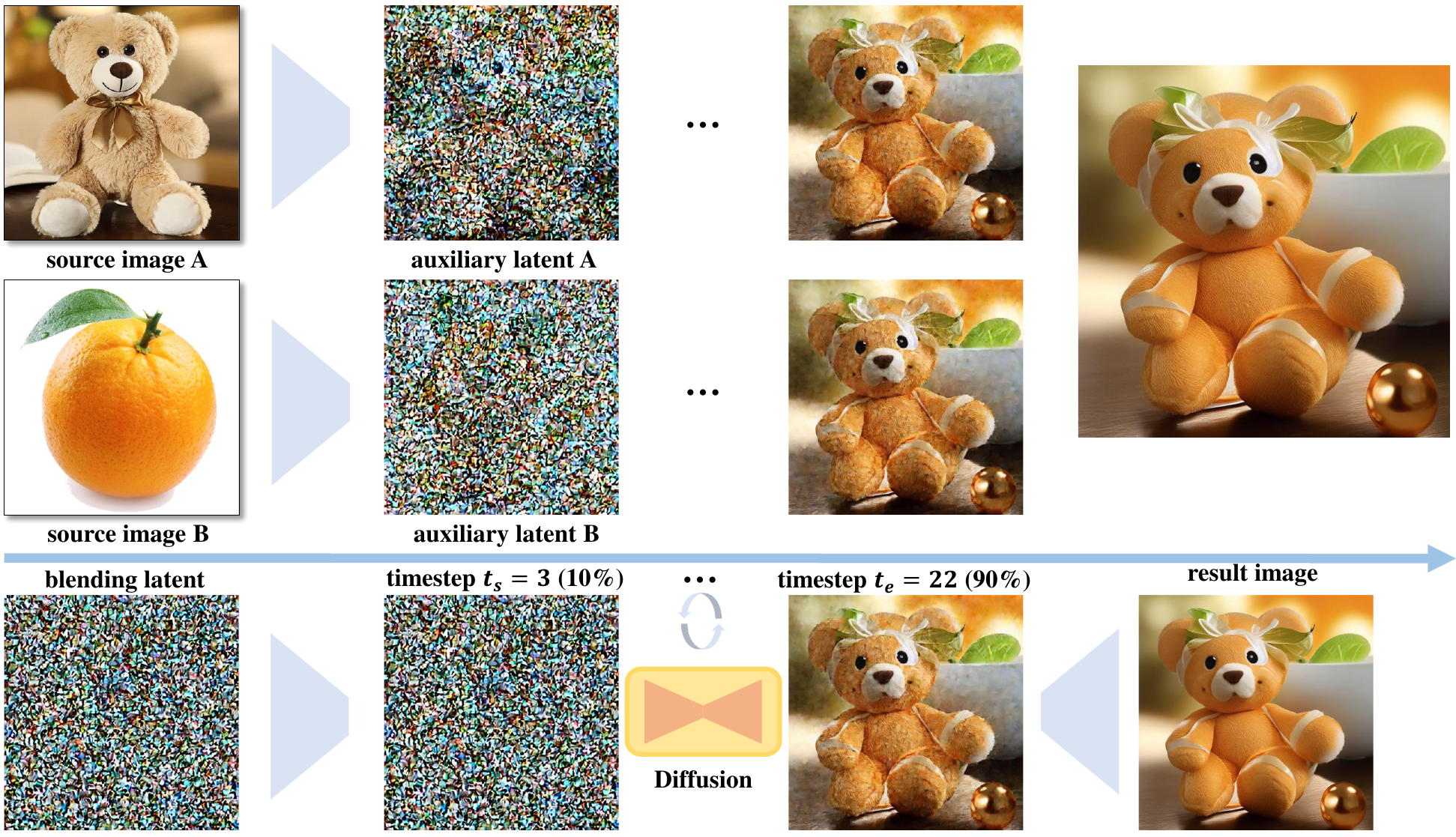}  
\caption{Visualization of our staged denoising process. From the image, we can observe the effects of the three stages. From the start to \( t_s \), the image establishes the basic layout of the blending. During the blending stage, both the blending latent and the auxiliary latents are iteratively updated. Finally, in the refinement stage, the quality of the blended image is significantly improved. }
\label{visualize_latent}
\end{center}
\end{figure*}

\textbf{Visualization of Staged Denoising Process. }Here, we provide a more detailed description of the staged denoising process employed in our method. To facilitate understanding, we visualize the intermediate latent representations, including the blending latent \( L_b \) and the auxiliary latents \( L_a^{(t,n)} \) (\( n=0,1,2,...,N \)) at various stages of the process. The results are presented in \Cref{visualize_latent}. Specifically, we focus on timesteps 3 and 22, as they correspond to the initial and final stages where our method begins to exhibit its effects. These visualizations demonstrate how the latents evolve over timesteps, gradually converging to a consistent structure. Through the feedback interpolation module, the blending latent and auxiliary latents seamlessly merge, ultimately resulting in a coherent output image.

\begin{table}[ht]
\vspace{-4mm}
\caption{Ablation study with our feedback mechanism. }
\label{ablation_feedback}
\begin{center}
\scalebox{1}{ 
\begin{small}
\begin{tabular}{ p{0.07\textwidth}p{0.14\textwidth}p{0.14\textwidth}p{0.14\textwidth}p{0.14\textwidth}}
\toprule
Feedback  & \multicolumn{1}{c}{CLIP-BS($\uparrow$)}  & \multicolumn{1}{c}{DINO-BS($\uparrow$)}  & \multicolumn{1}{c}{CLIP-IQA($\uparrow$)}  & \multicolumn{1}{c}{HPS($\uparrow$)}   \\ 
\midrule 
& 8.9211\textsubscript{$\pm$ 3.0076} & \textbf{0.2860\textsubscript{$\pm$ 0.1663}} & 0.4874\textsubscript{$\pm$ 0.1035} & 0.2866\textsubscript{$\pm$ 0.0305}  \\

\centering \checkmark & \textbf{9.1555\textsubscript{$\pm$ 2.7134}}& 0.2743\textsubscript{$\pm$ 0.1586}& \textbf{0.5238\textsubscript{$\pm$ 0.0975}}& \textbf{0.2932\textsubscript{$\pm$ 0.0316}}\\

\bottomrule
\end{tabular}
\end{small}
}
\end{center}
\vspace{-4mm}
\end{table}

\section{Details of Dataset Construction}
\label{dataset_construction_sec}
\begin{table}[ht]
\caption{Categories of Concepts. Categories of these selected concepts for blending are designed to encompass the primary objects commonly found in the real world.}
\vspace{3mm}
\centering
\begin{tabular}{|c|c|}
\hline
\textbf{Category} & \textbf{Names} \\ \hline
Transport & airplane, bicycle, car, 
 ship\\ \hline
Animals & bird, cat, cattle, dog, fish, squirrel, shark  \\ \hline
Common objects & chair, desk, phone, robot, lamp \\ \hline
Nature & rose, lavender, lemon, water \\ \hline
\end{tabular}
\label{categories_table}
\end{table}

The designed category list can be divided into four distinct groups based on their characteristics, as shown in \Cref{categories_table}. This categorization provides a clear organization of items based on their respective categories and reflects common concepts encountered in daily life.

\section{Why Vanilla Stable Diffusion Fails}

\begin{figure*}[ht]
\vskip 0.2in
\begin{center}
\includegraphics[width=1.0\textwidth, height=8.3cm]{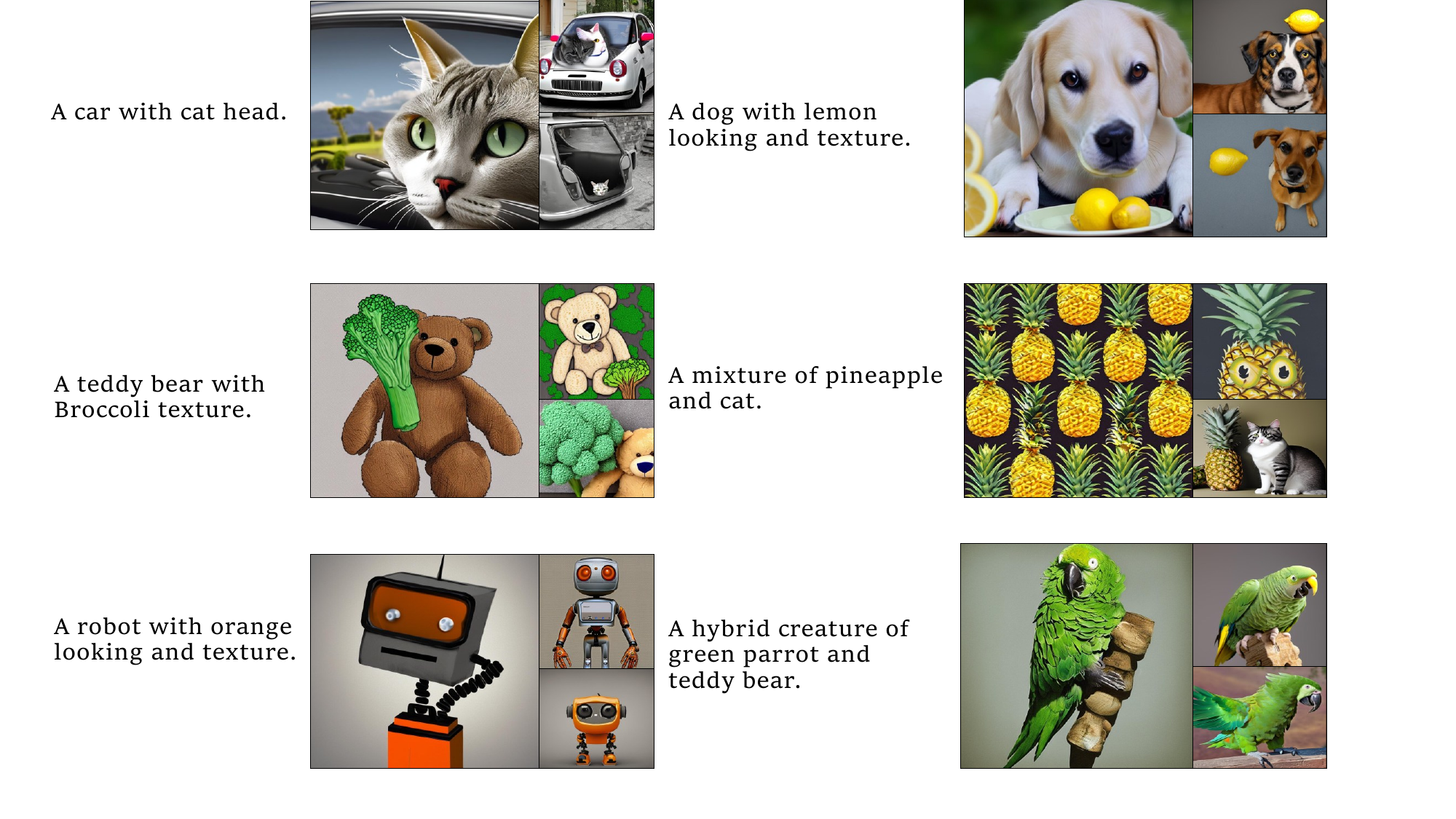}  
\caption{Relying solely on prompts for concept blending, we find that Stable Diffusion 2.1 often fails to effectively merge the concepts as instructed. Instead, it typically depicts the two concepts as separate, coexisting elements within the image.}
\label{sd_prompt}
\end{center}
\vskip -0.2in
\end{figure*}

\section{More Style Tasks}

\begin{figure*}[ht]
\vskip 0.2in
\begin{center}
\includegraphics[width=0.98\textwidth]{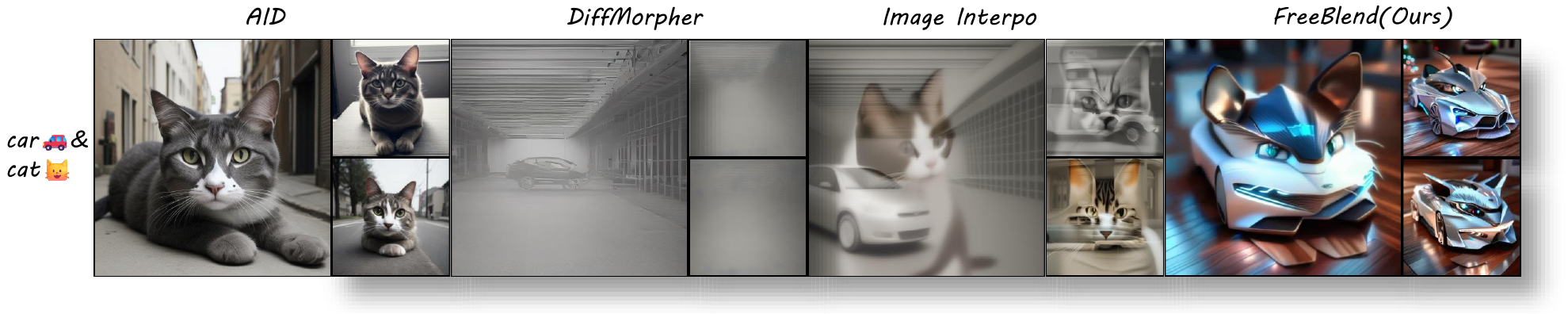}  
\caption{When compared to other image interpolation methods, we find that AID~\citep{qiyuan2024aid} exhibits a stronger attention bias towards maintaining the quality of generated images, but it does not effectively blend concepts together. DiffMorpher~\citep{zhang2024diffmorpher} is designed to smoothly transfer one image to another, but when applied to unrelated concepts, it generates white noise. On the other hand, Image Interpo~\citep{wang2023interpolating} interpolates between two different images but suffers from significant overlapping issues.}
\label{image_interpolation}
\end{center}
\vskip -0.2in
\end{figure*}

\textbf{Image Interpolation Task. }\Cref{image_interpolation} presents a qualitative comparison with other image interpolation methods, demonstrating that our method achieves the best results.

\begin{figure*}[ht]
\begin{center}
\includegraphics[width=0.6\textwidth]{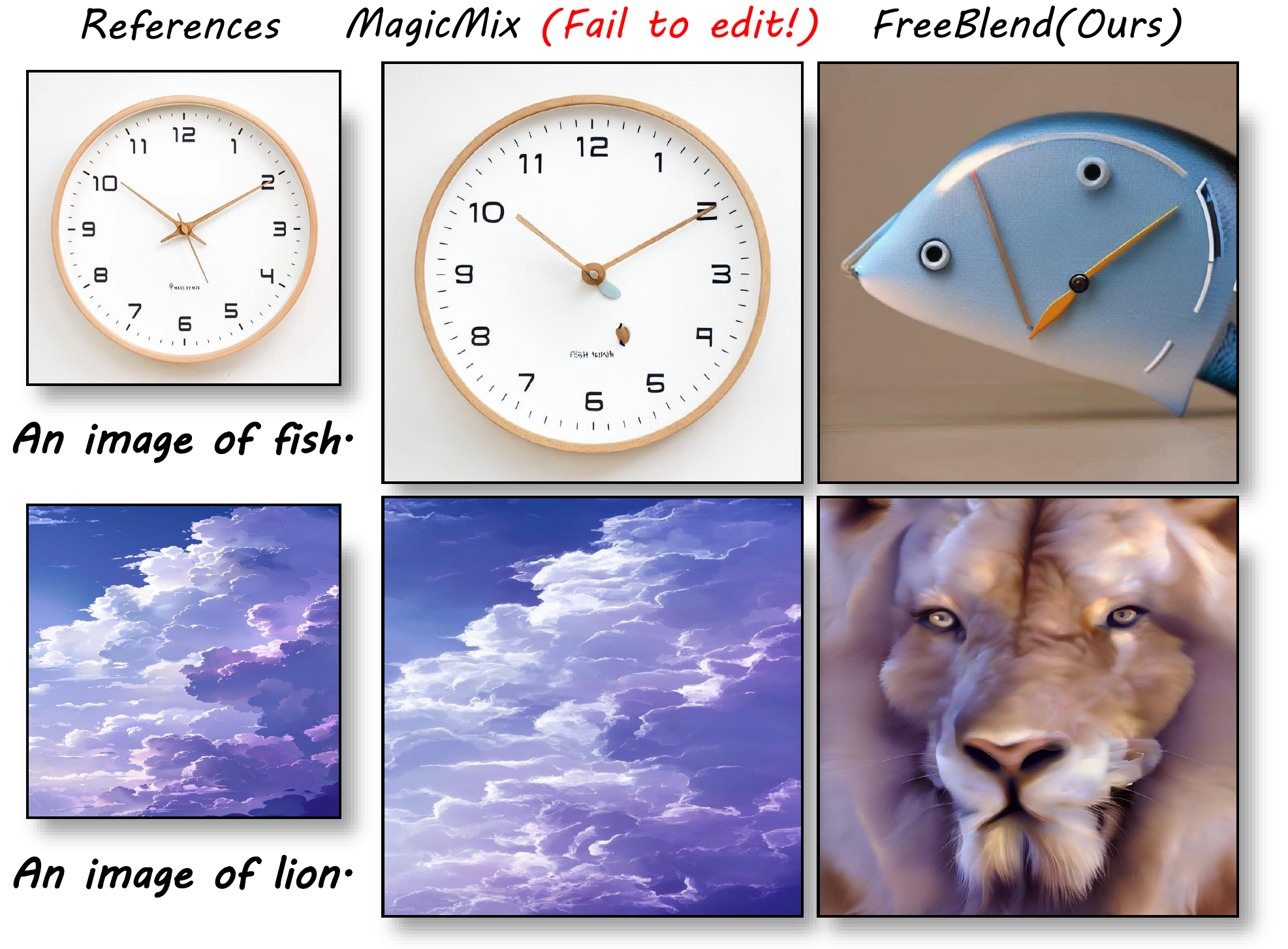} 
\caption{Comparison with the setting of MagicMix~\citep{liew2022magicmix}. Given a reference image and a prompt, the model can transfer the reference image to another concept. From the results, we observe that when there is a significant semantic and shape gap between the original and target concepts, MagicMix struggles to modify the original image while preserving its features, without introducing new ones. This highlights that our method is capable of handling concepts with substantial semantic and shape differences.}
\label{image_switching}
\end{center}
\vskip -0.2in
\end{figure*}

\textbf{Switching One Condition from Image to Text. }\Cref{image_switching} presents the qualitative comparison between MagicMix~\citep{liew2022magicmix} and our method.

\begin{figure*}[htbp]
\vskip 0.2in
\begin{center}
\includegraphics[width=1.0\textwidth, height=9cm]{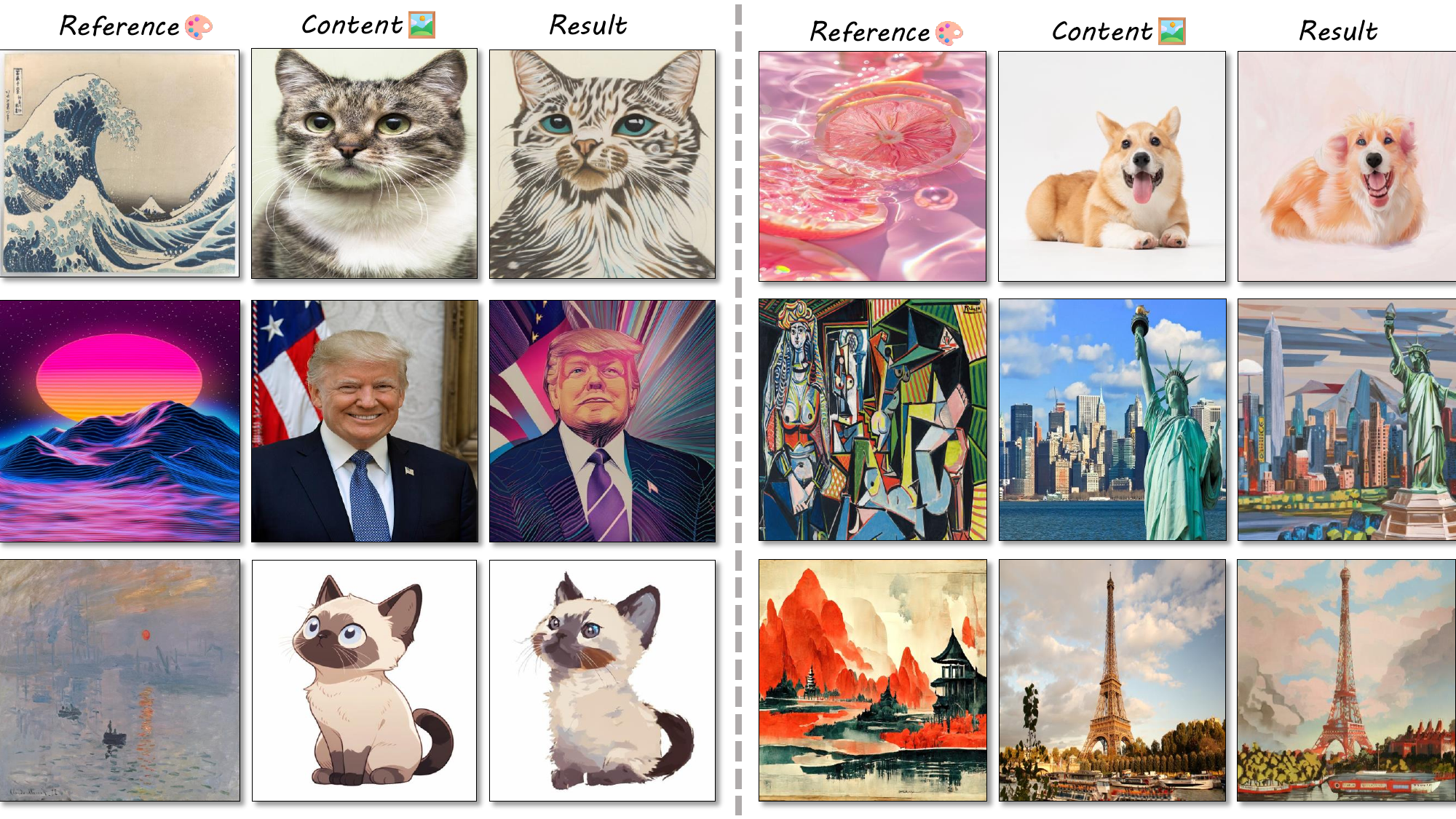}  
\caption{Adapting our method to the style transfer task.}
\label{style_transfer_images}
\end{center}
\vspace{-8mm}
\end{figure*}

\textbf{Style Transfer Task. }\Cref{style_transfer_images} illustrates the impressive visual performance of our method on the style transfer task. Given a reference image and a content image, our training-free approach effortlessly transfers the content into the style of the reference, demonstrating both the adaptability and compatibility of our method for this task.

\section{Details of Metrics}
\label{more_about_metrics}

We calculated the corresponding scores for each image within a category, then computed the average score for that category, and finally determined the overall average score for the entire dataset.

\textbf{CLIP-BS. }For class $A$ and class $B$, we calculate the CLIP blending score using the following equation:
\begin{equation}
    \text{Score}_{\text{CLIP-BS}} = \sum_{i=1}^{n} \left[ \text{Score}(\text{Image}_{\text{blending}}, \text{Prompt}_{i}) - \text{Score}(\text{Image}_{\text{original}_{i}}, \text{Prompt}_{i}) \right],
\end{equation}
where $\text{Score}(\cdot, \cdot)$ is the cosine similarity score computed by the CLIP model using the input image and prompt, $\text{Image}_{\text{blending}}$ represents the generated blended images, $\text{Image}_{\text{original}_{i}}$ denotes the original images of class $i$,  and $n$ is the number of concepts blended. $\text{Prompt}_{i}$ refers to the prompt for class $i$, formatted as ``a photo of a \{class\}'' to minimize noise in the image generation process. This metric can also be interpreted as the high-dimensional distance between the blended concepts and the original ones. It represents the total distance between the blended concept and each of the original concepts. A larger score generally indicates the dimensional center of the original concepts. Note that irrelevant content could potentially receive a high score using this metric due to unrelated features. However, this scenario is unlikely to occur, as our generated images are based on relevant concepts.

\textbf{DINO-BS~\citep{liu2025grounding}. }Grounding DINO is an open-set object detection model designed to improve the ability to interpret human language inputs through tight modality fusion and large-scale grounded pre-training. The model excels at detecting unseen objects and can perform effective zero-shot transfer learning, even for categories not encountered during training, by leveraging language prompts. To guide the model in detecting blended objects, we designed two specific prompts: ``a \{concept\_a\} with blending features from \{concept\_b\}'' and ``a \{concept\_b\} with blending features from \{concept\_a\}''. These prompts help the model appropriately recognize and identify the blended objects.

\textbf{CLIP-IQA~\citep{wang2023exploring}. }This metric employs a thoughtfully crafted prompt pairing strategy to reduce linguistic ambiguity and effectively harness CLIP's pretrained knowledge for assessing visual perception. In our task, we design specific prompt pairs like (``mixed'', ``dull''), (``blending features from two different objects'', ``natural object from one object'').

\textbf{HPS~\citep{wu2023human}. }This metric more accurately predicts human preferences for generated images. The designed prompt is ``a photo of a blended object combining mixed features from \{concept\_a\} and \{concept\_b\}''.

\section{Baseline Methods}
In this section, we describe the baseline methods used in our study:
\begin{itemize}
    \item \textbf{MagicMix~\citep{liew2022magicmix}. }The original MagicMix method modifies the input image using a text prompt from a different class to transfer the image’s characteristics. In our implementation, we first generate the original image using a class text prompt with Stable Diffusion, and then pass it into the original MagicMix pipeline.
    \item \textbf{Composable Diffusion~\citep{liu2022compositional}. }This method decomposes the text description into components, each processed by a different encoder to produce a latent vector capturing its semantic information. Multiple diffusion models are then used to independently generate images for each component. The Conjunction operation combines these components into a single image. This approach excels in zero-shot compositional generation, allowing the model to create novel combinations \textit{but not blending} of objects and their relationships, even without prior training on such combinations.
    \item \textbf{unCLIP-T~\citep{wang2023exploring}. }The unCLIP model leverages images as conditioning inputs, replacing traditional text prompts, by training several linear layer adapters. In this implementation, image embeddings are treated similarly to text embeddings, with their average embedding being computed to serve as the conditions.
    \item \textbf{unCLIP-U~\citep{wang2023exploring}. }This implementation is akin to unCLIP-T, with the key difference being that the image embeddings are integrated using the UNET method.
    \item \textbf{TEXTUAL~\citep{melzi2023does}. }This method encodes the text prompts into text embeddings and computes their average embedding to input into Stable Diffusion for generating blended images. However, it is unstable within the embedding space and constrained by the semantic scope of the concepts involved.
    \item \textbf{UNET~\citep{olearo2024blend}. }This method uses the first text embedding for the conditions of downsampling of U-Net~\citep{ronneberger2015u} and the second one for the upsampling of the U-Net.
    \item \textbf{AID~\citep{qiyuan2024aid}. }This paper introduces Attention Interpolation via Diffusion (AID), a training-free method for improving conditional image interpolation in text-to-image models. AID enhances consistency and smoothness, with a variant (PAID) enabling user-guided interpolation. Experiments show AID outperforms traditional methods and enhances control for image editing.
    
\end{itemize}

\section{User Study}
\begin{figure*}[ht]
\begin{center}
\begin{minipage}{0.48\textwidth}
    \centering
    \includegraphics[width=\textwidth, height=6cm]{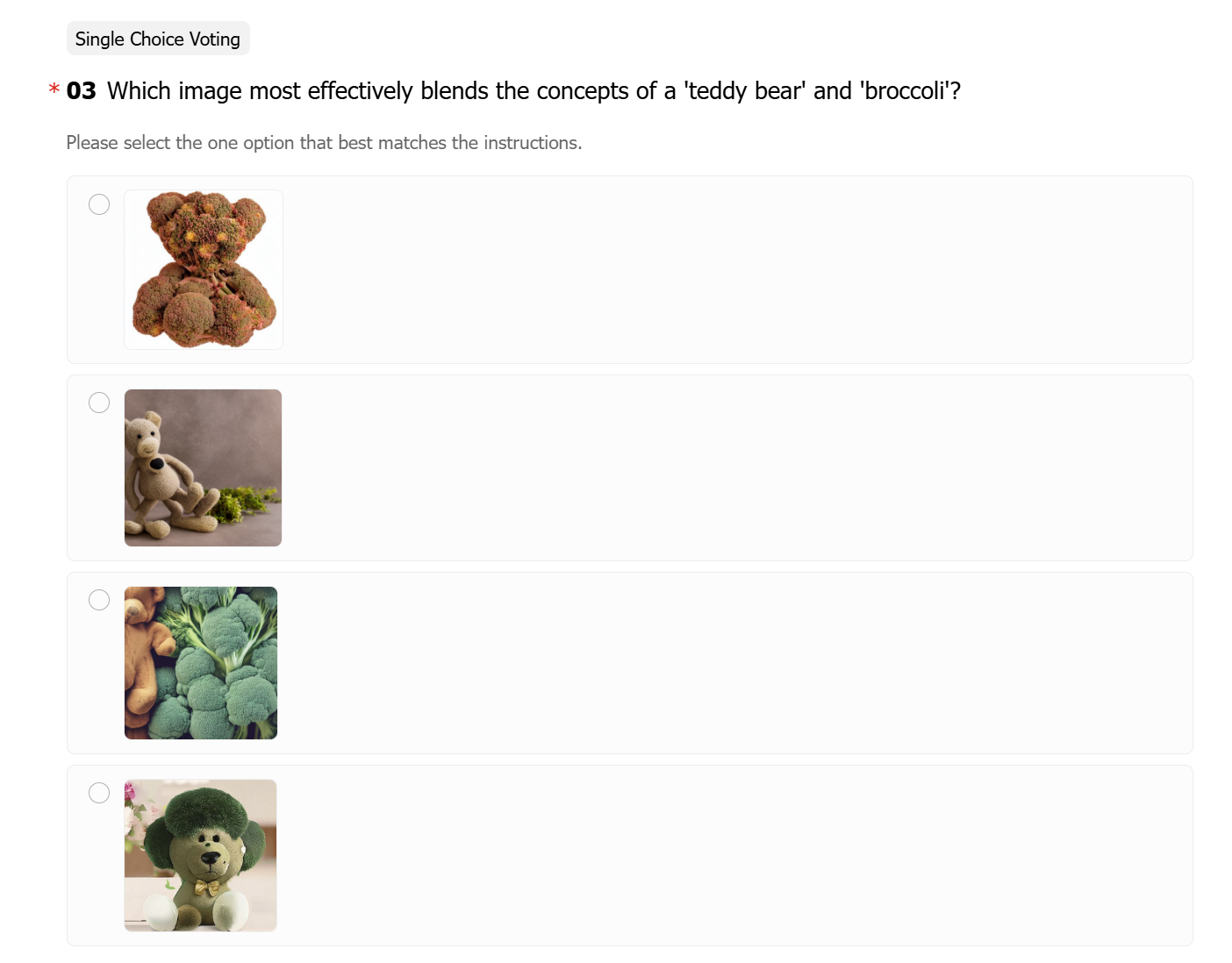}
\end{minipage}
\hfill
\begin{minipage}{0.48\textwidth}
    \centering
    \includegraphics[width=\textwidth, height=6cm]{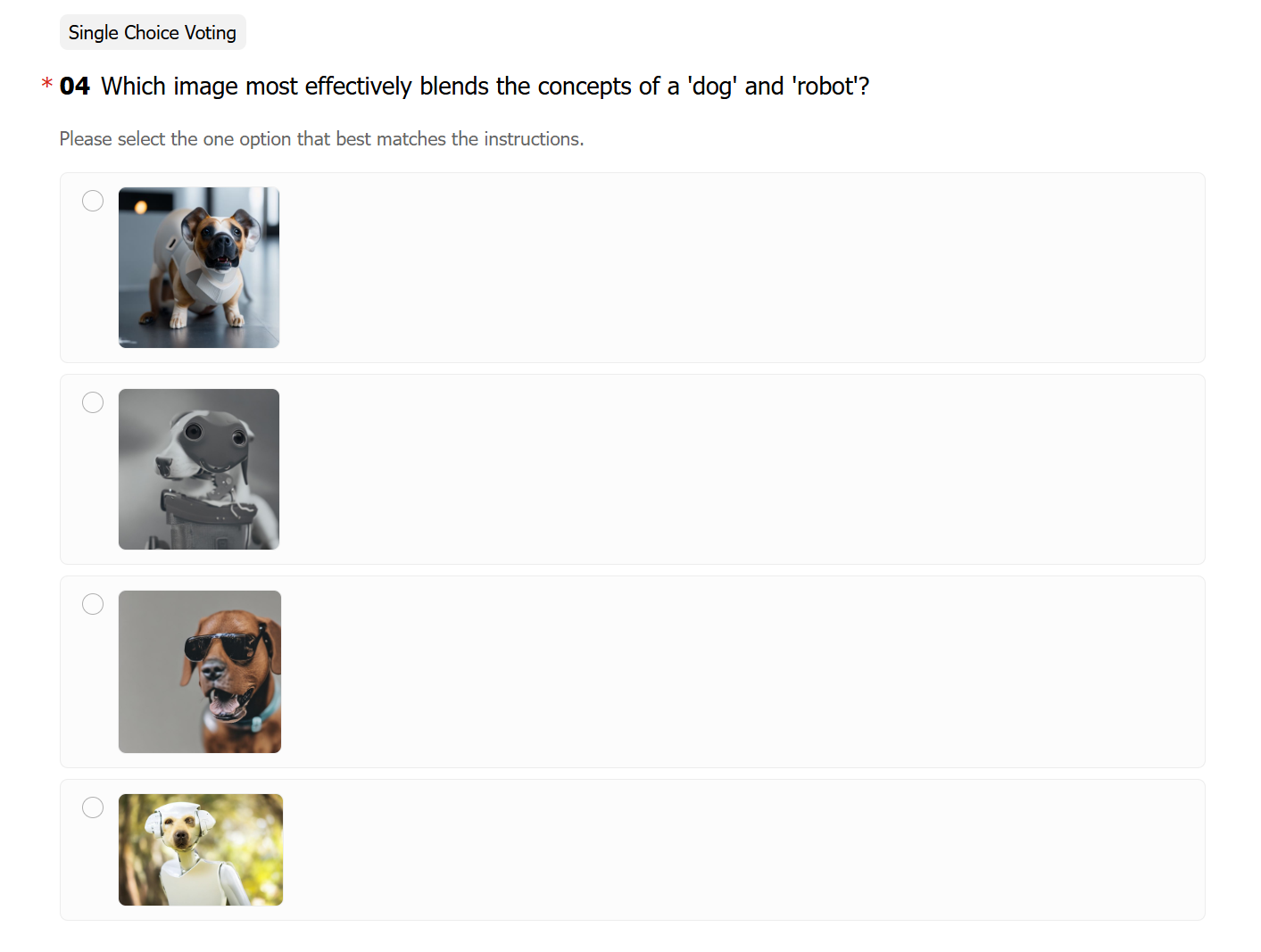}

\end{minipage}
\end{center}
\caption{Screenshots of user study.}
    \label{fig:screenshot}
\end{figure*}

The screenshots in \Cref{fig:screenshot} display our questionnaires used to assess the validation of different methods.

\section{More Visual Results}
\begin{figure*}[ht]
\centering
\resizebox{0.995\linewidth}{!}{
\setlength{\tabcolsep}{0mm}
\renewcommand{\arraystretch}{0.01}
\begin{tabular}{ccccc}
\includegraphics[trim=1cm 1cm 1cm 1cm,clip,width=0.2\linewidth]{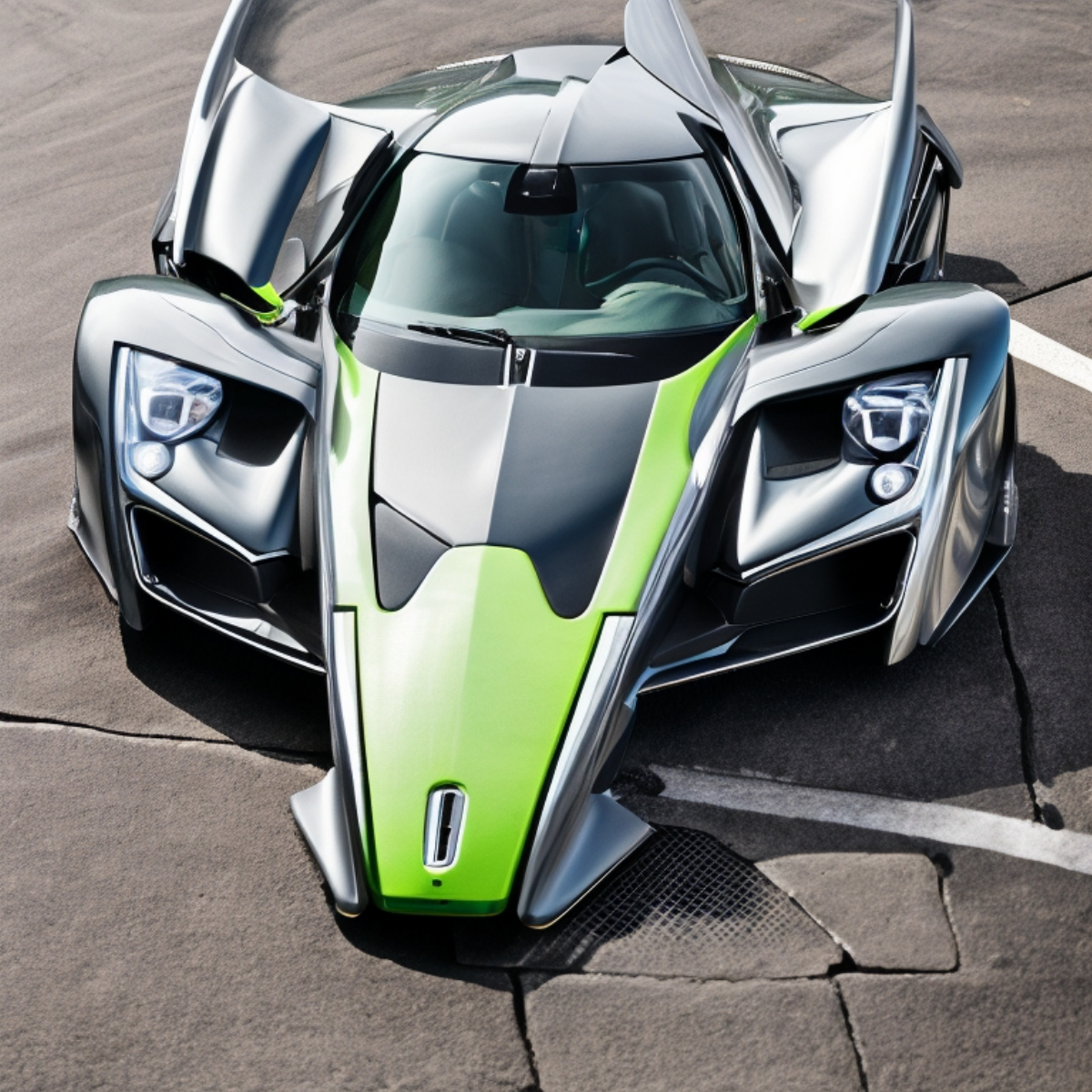} &
\includegraphics[trim=1cm 1cm 1cm 1cm,clip,width=0.2\linewidth]{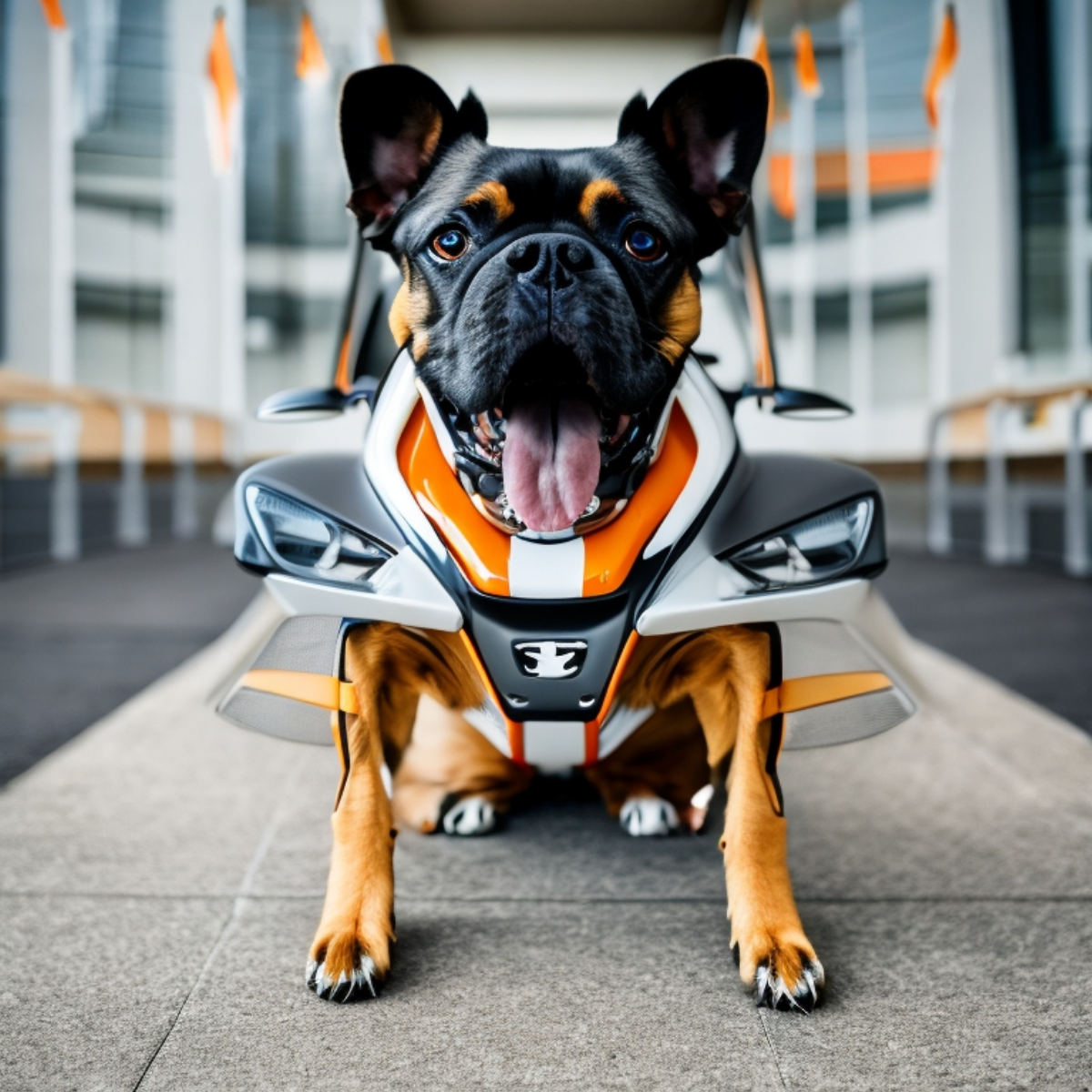} &
\includegraphics[trim=1cm 1cm 1cm 1cm,clip,width=0.2\linewidth]{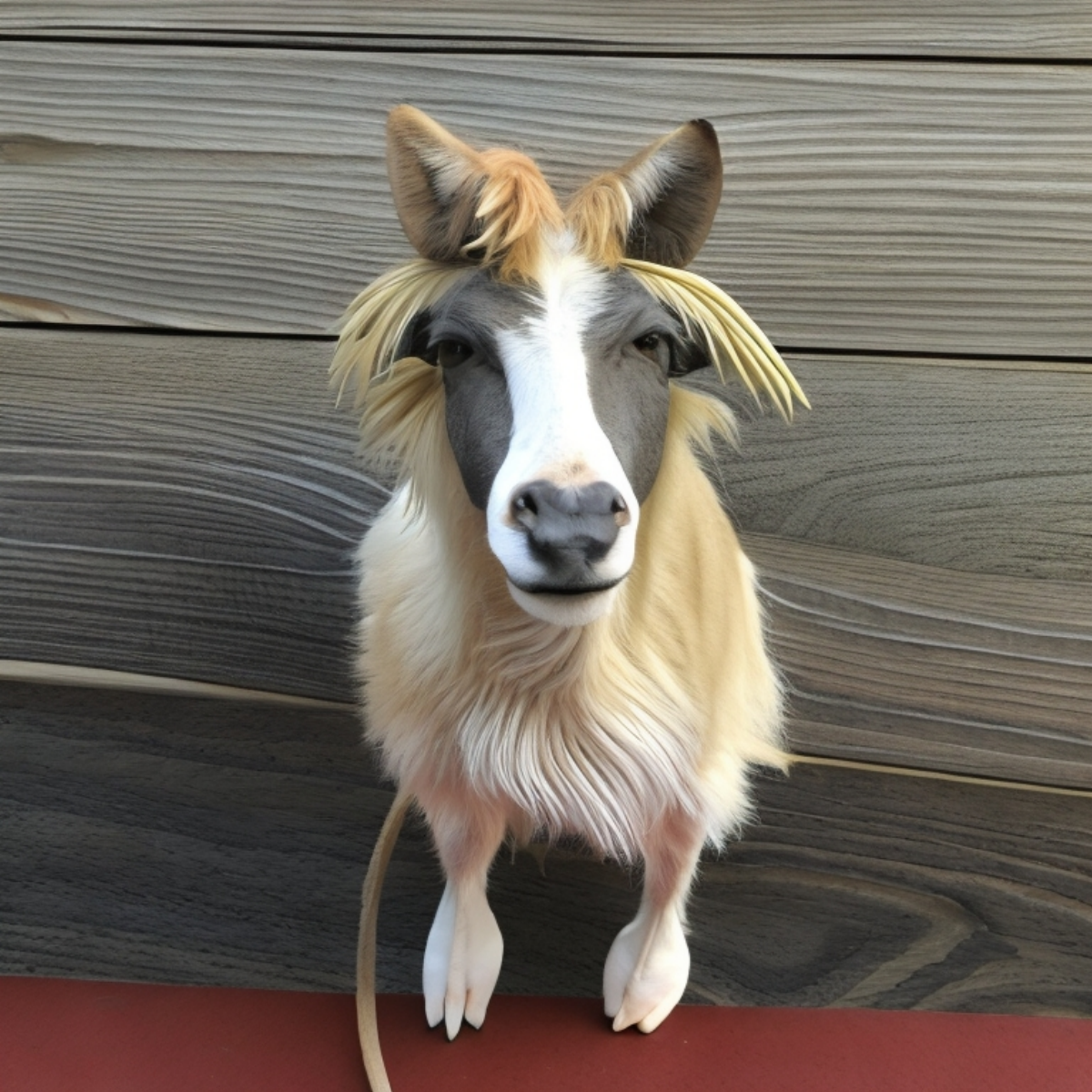} &
\includegraphics[trim=1cm 1cm 1cm 1cm,clip,width=0.2\linewidth]{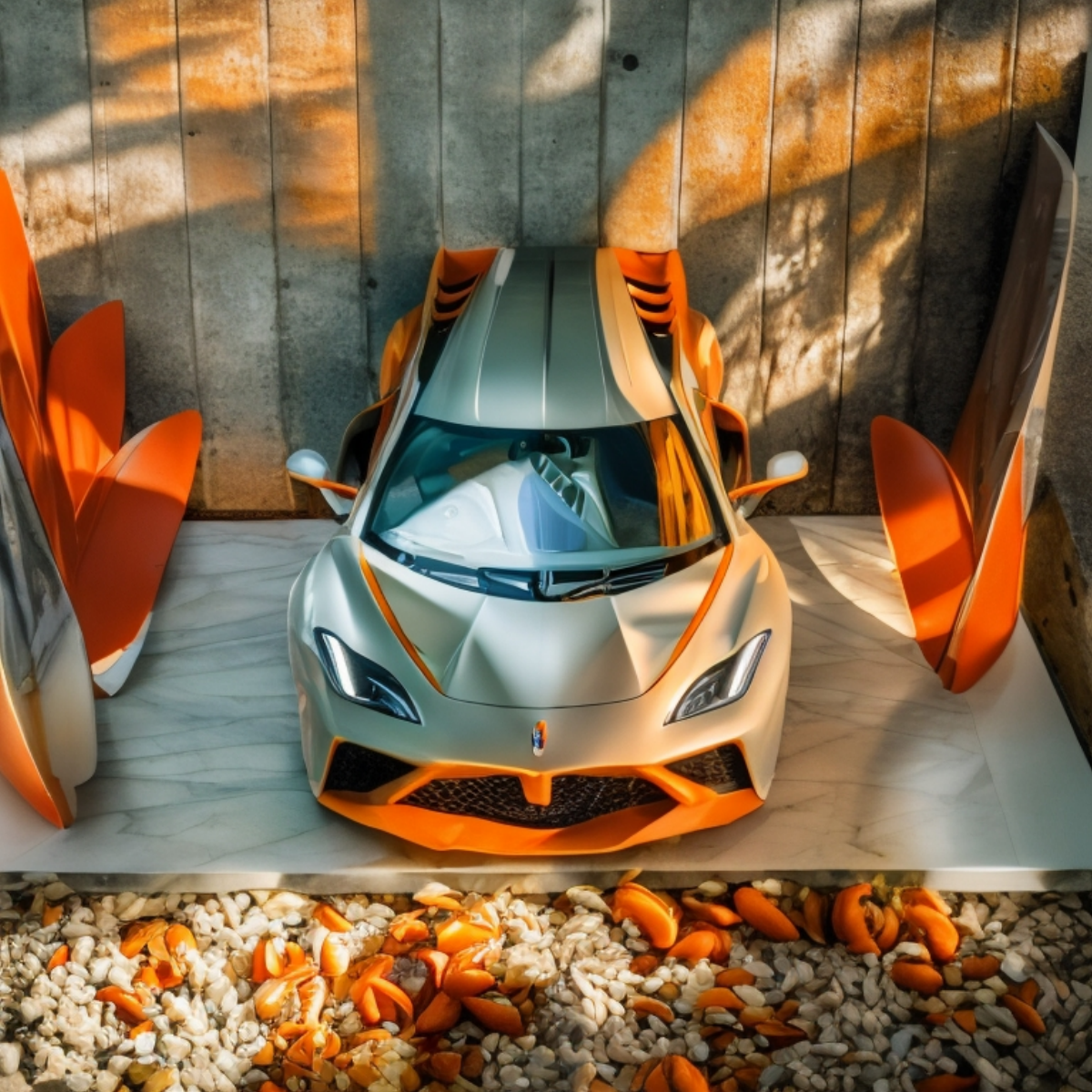} &
\includegraphics[trim=1cm 1cm 1cm 1cm,clip,width=0.2\linewidth]{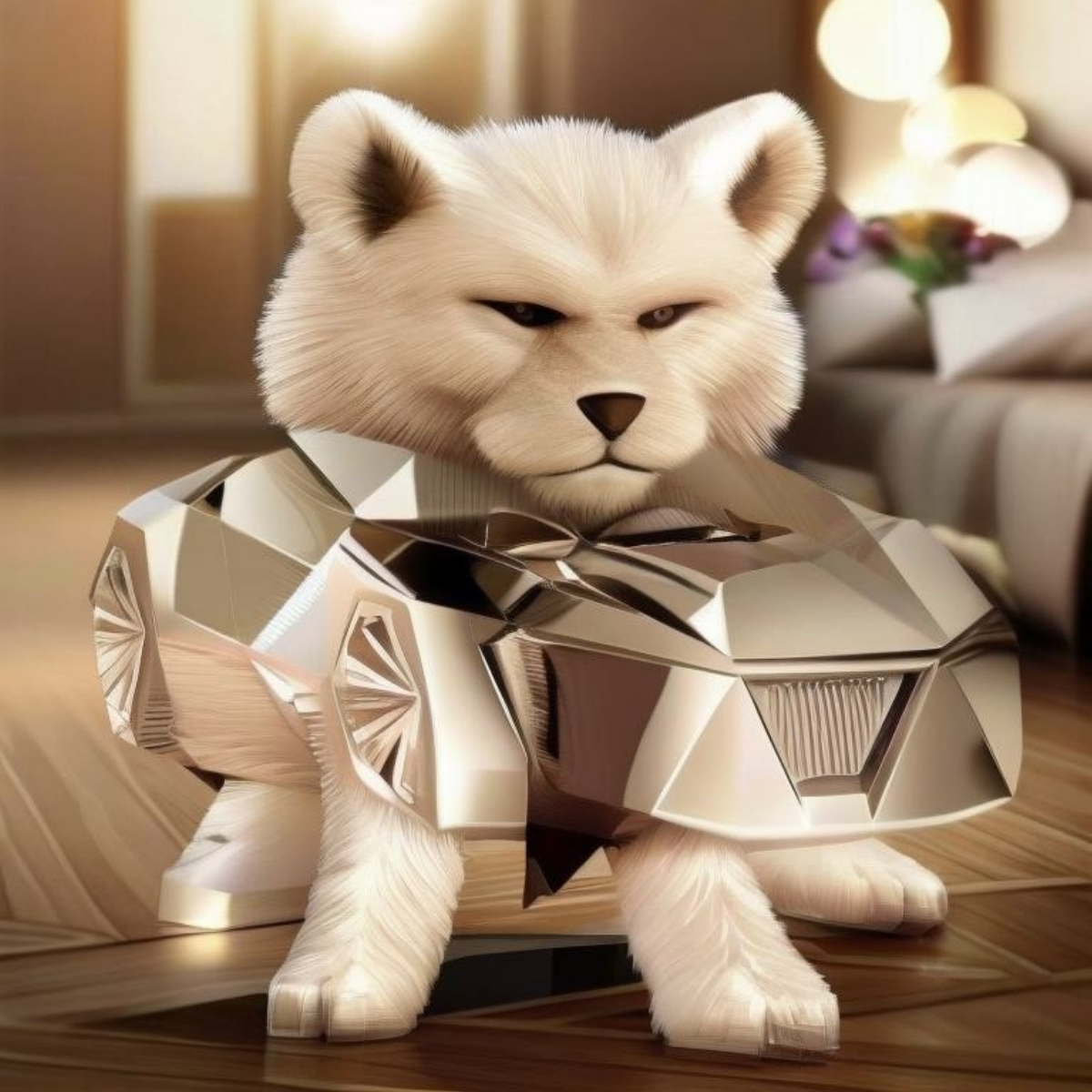} \\
\vspace{1mm} 
\small car-brocolli & \small car-dog & \small chicken-donkey & \small super car-orange& \small car-teddy    \\

\includegraphics[trim=1cm 1cm 1cm 1cm,clip,width=0.2\linewidth]{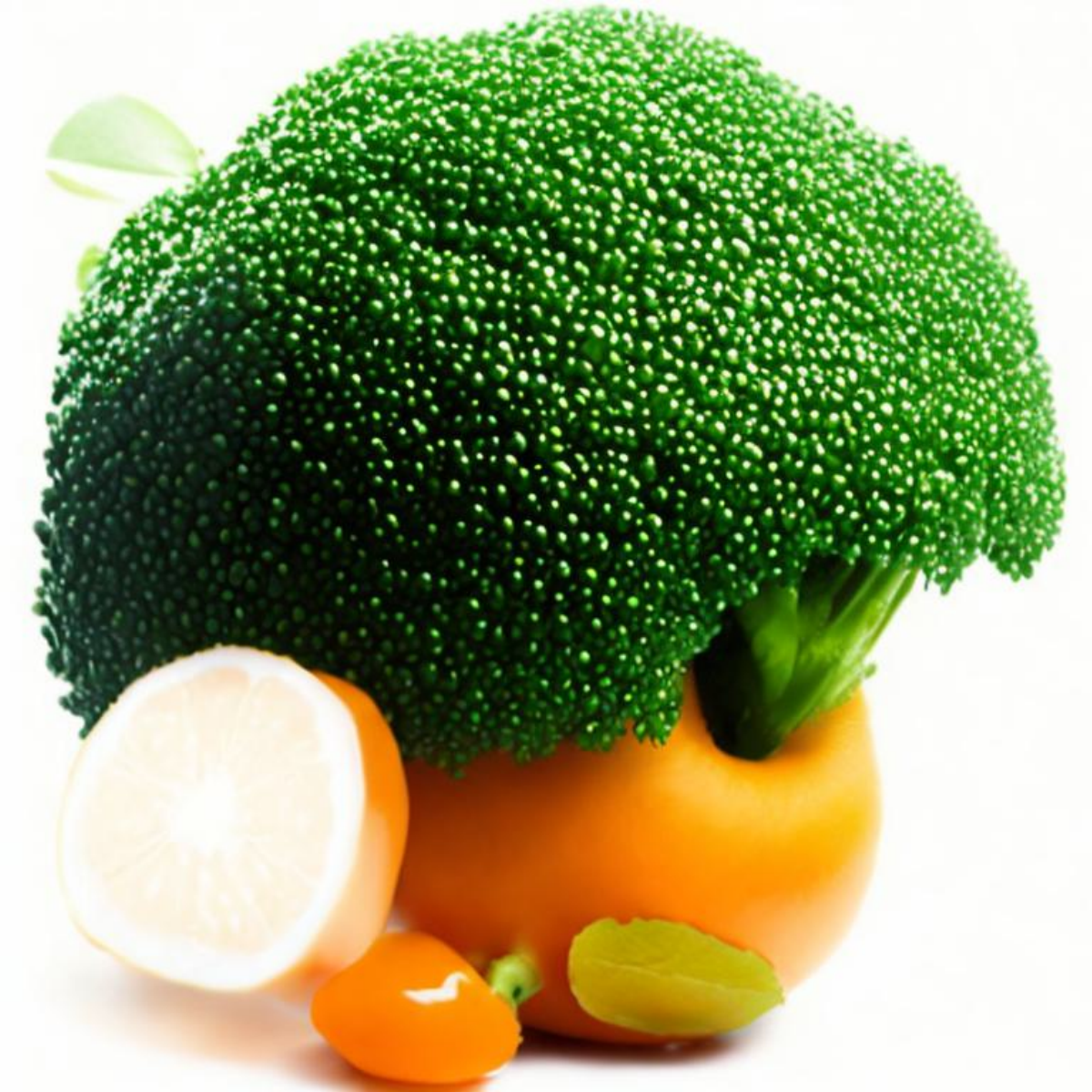} &
\includegraphics[trim=1cm 1cm 1cm 1cm,clip,width=0.2\linewidth]{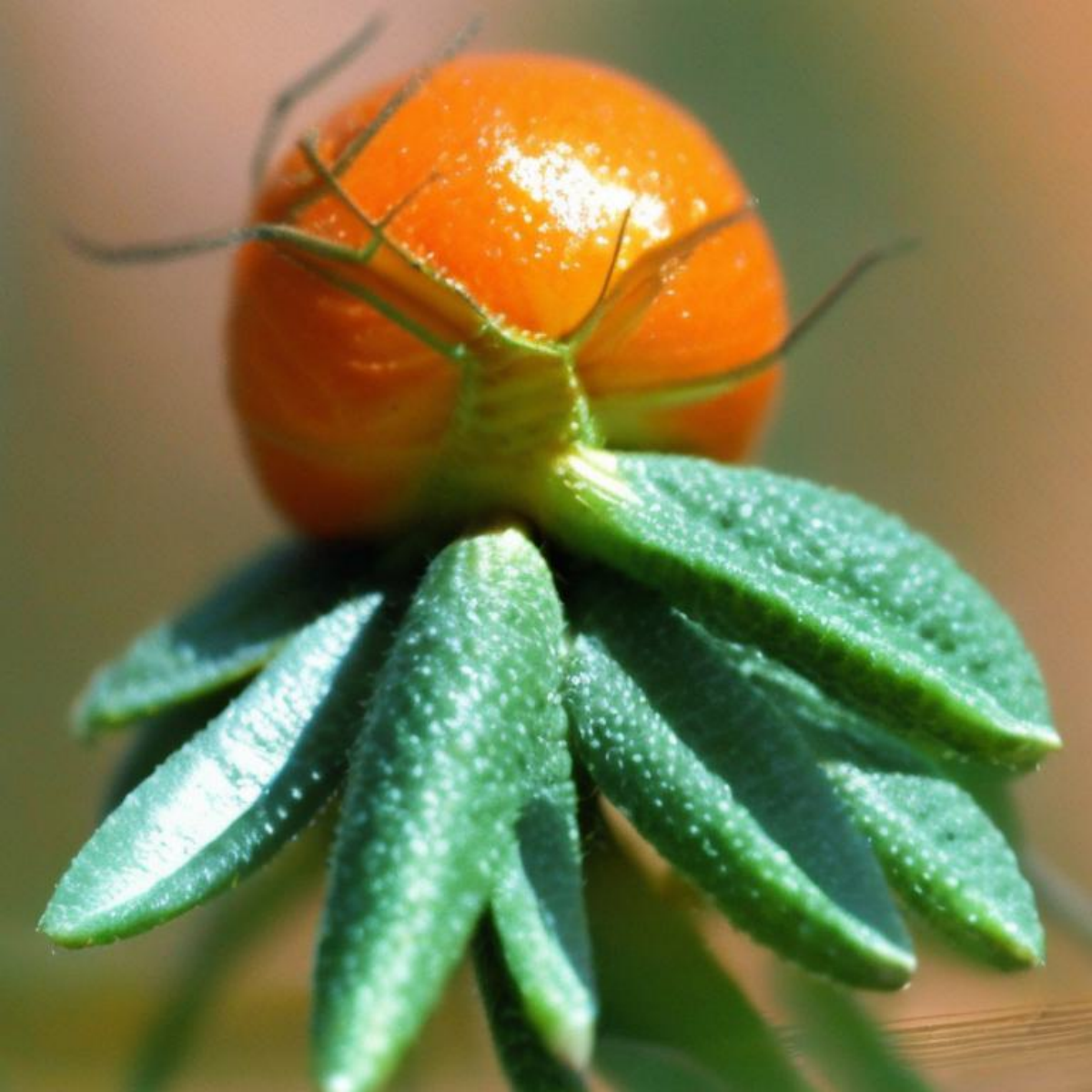} &
\includegraphics[trim=1cm 1cm 1cm 1cm,clip,width=0.2\linewidth]{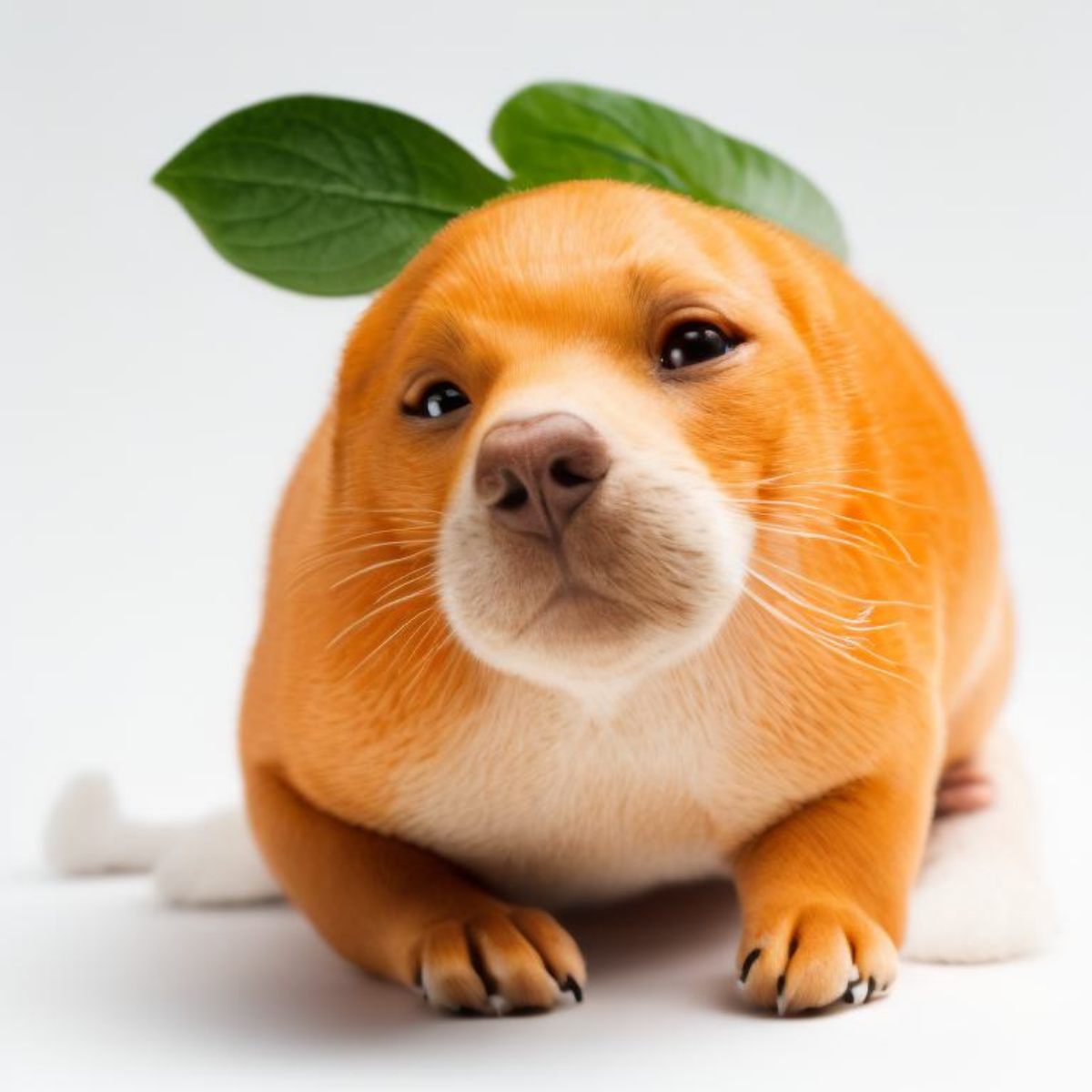} &
\includegraphics[trim=1cm 1cm 1cm 1cm,clip,width=0.2\linewidth]{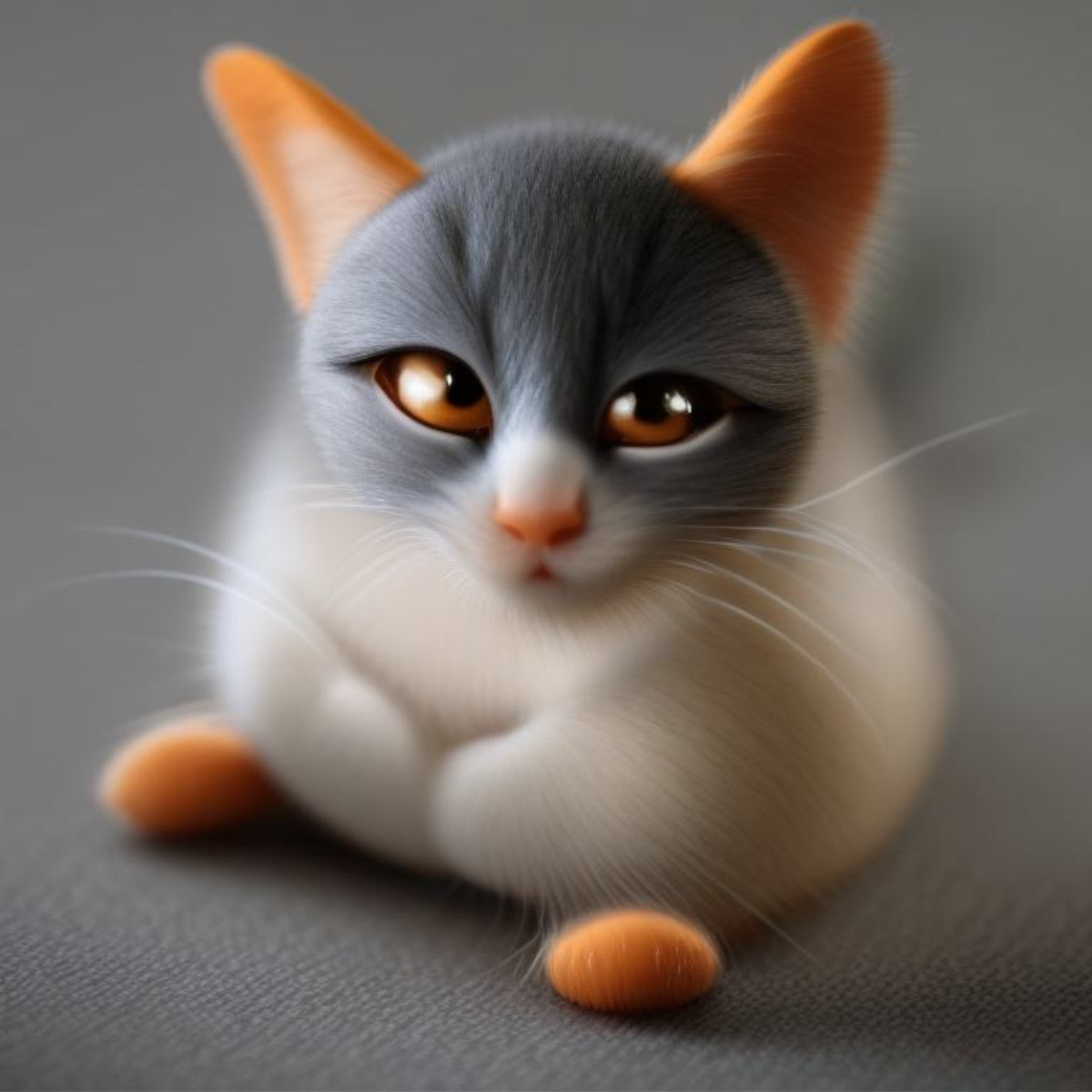} &
\includegraphics[trim=1cm 1cm 1cm 1cm,clip,width=0.2\linewidth]{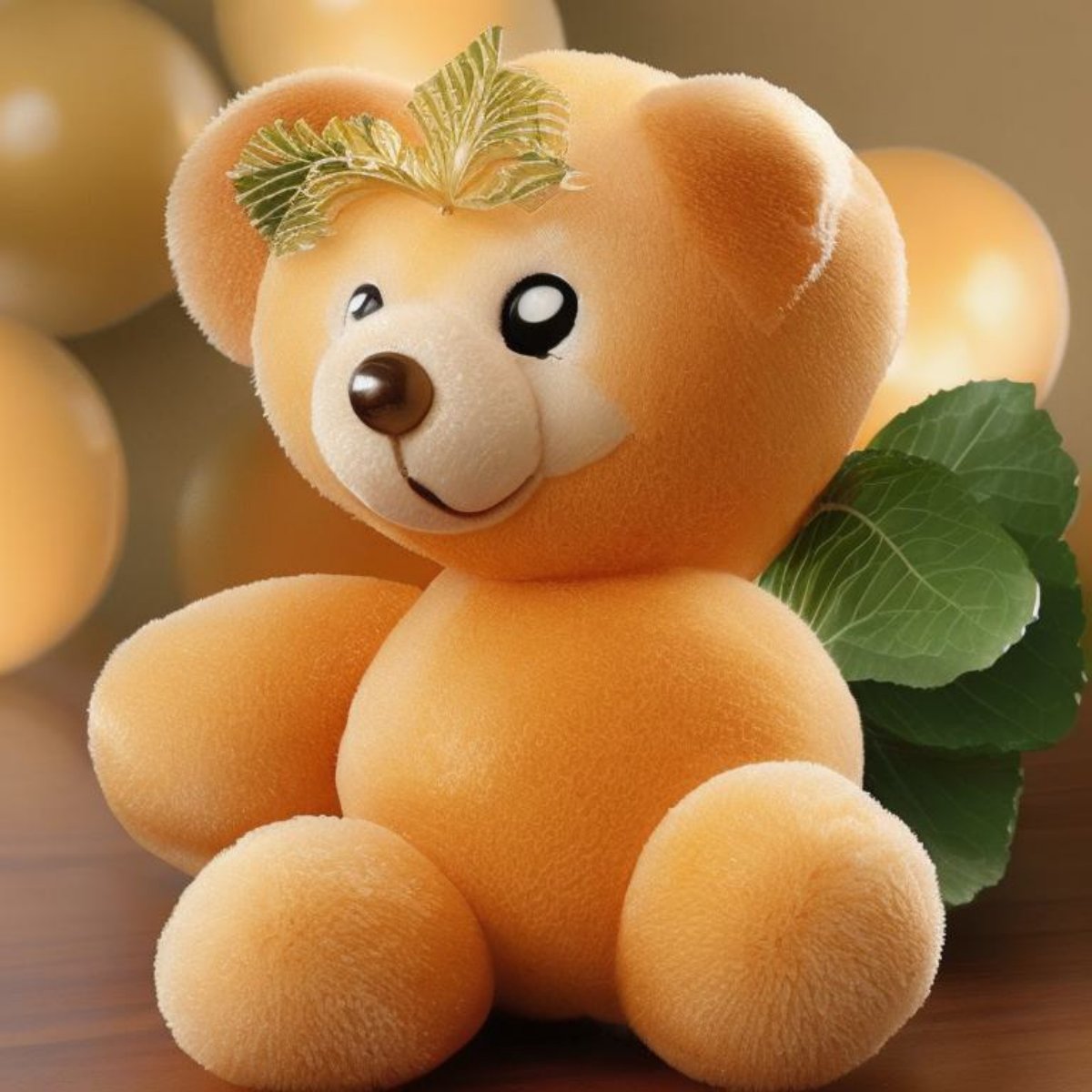} \\
\vspace{1mm} 
\small orange-brocolli & \small orange-cactus & \small orange-dog & \small orange-cat & \small orange-teddy  \\
\includegraphics[trim=1cm 1cm 1cm 1cm,clip,width=0.2\linewidth]{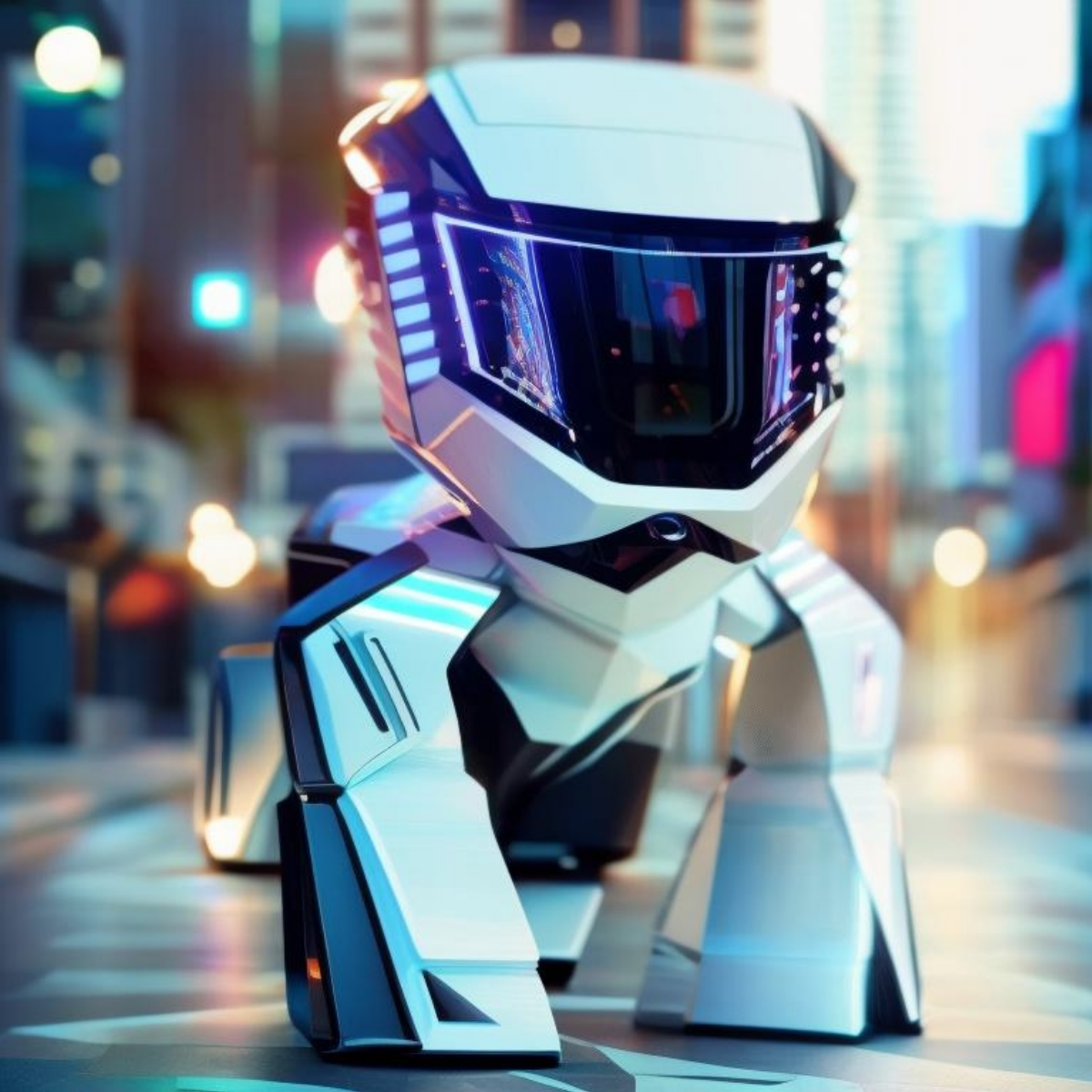} &
\includegraphics[trim=1cm 1cm 1cm 1cm,clip,width=0.2\linewidth]{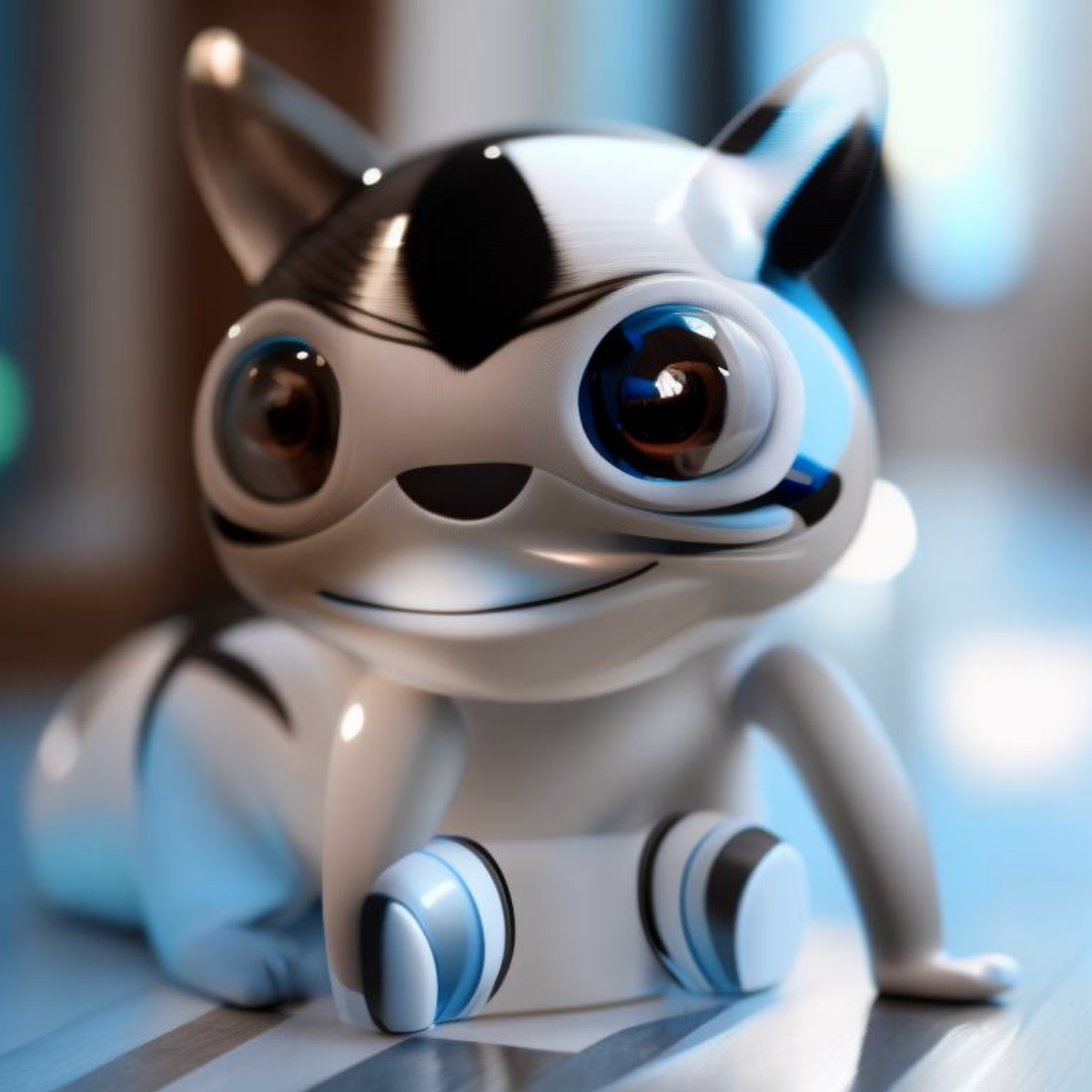} &
\includegraphics[trim=1cm 1cm 1cm 1cm,clip,width=0.2\linewidth]{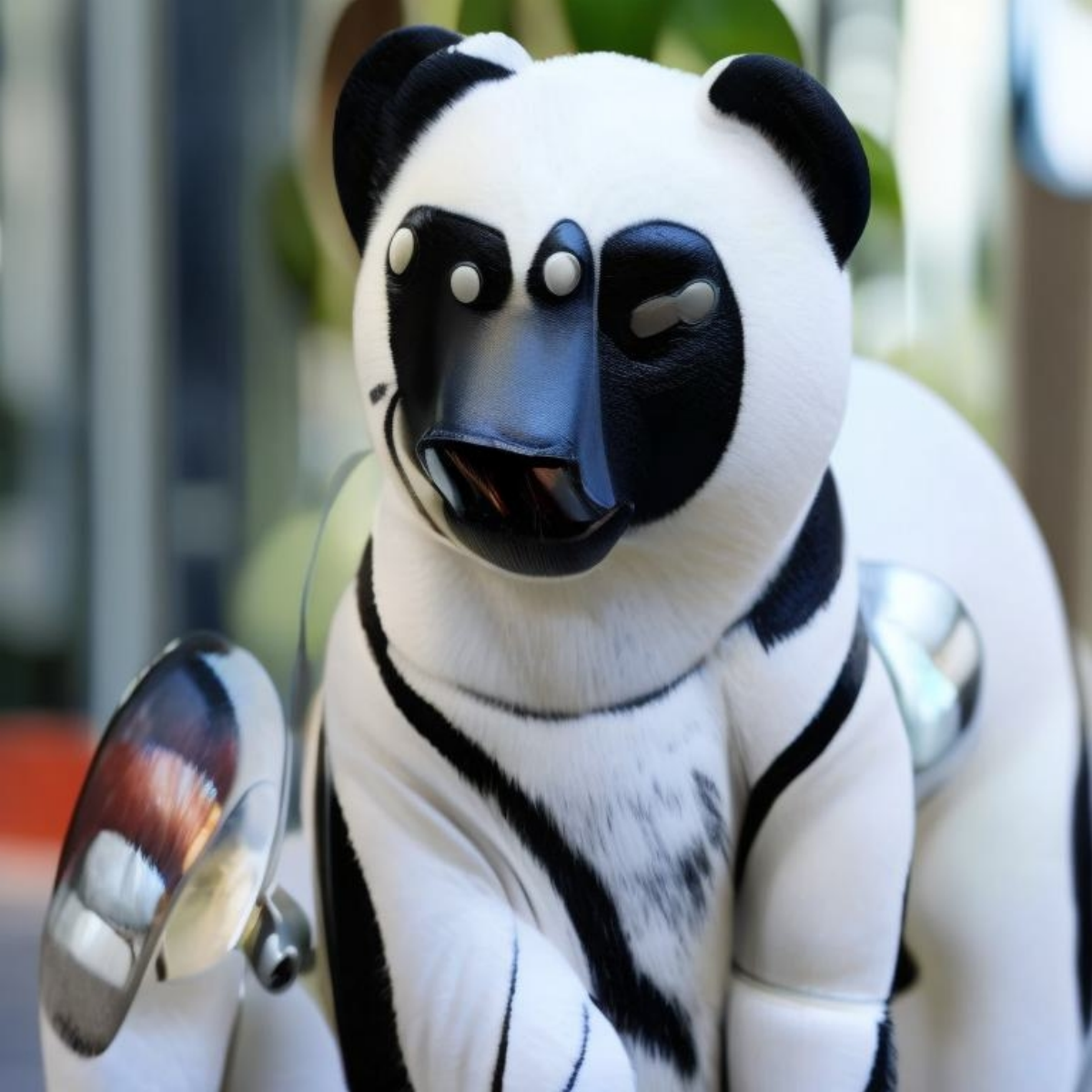} &
\includegraphics[trim=1cm 1cm 1cm 1cm,clip,width=0.2\linewidth]{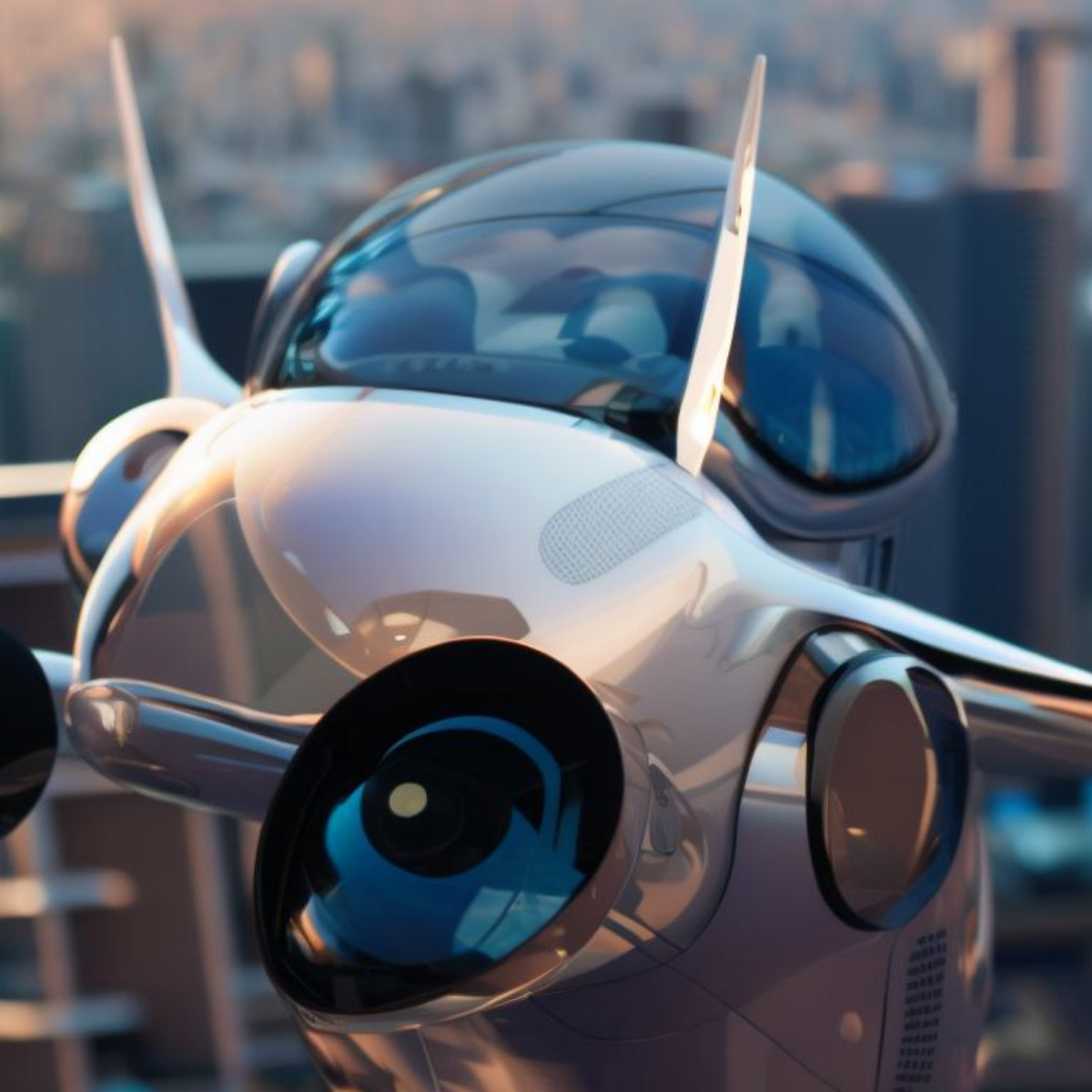} &
\includegraphics[trim=1cm 1cm 1cm 1cm,clip,width=0.2\linewidth]{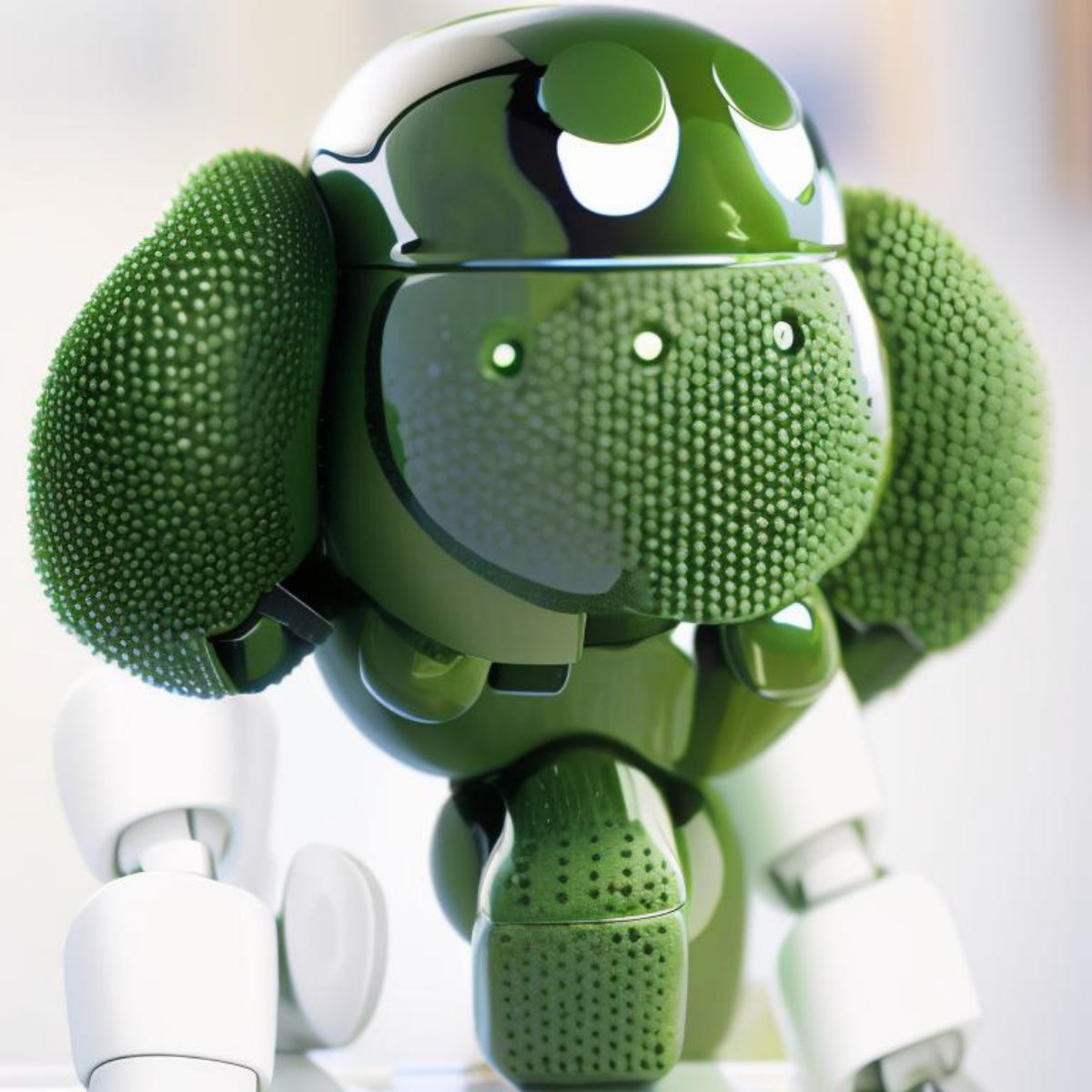} \\
\vspace{1mm} 
\small robot-car & \small robot-cat & \small robot-panda & \small robot-plane & \small robot-brocolli    \\
\includegraphics[trim=1cm 1cm 1cm 1cm,clip,width=0.2\linewidth]{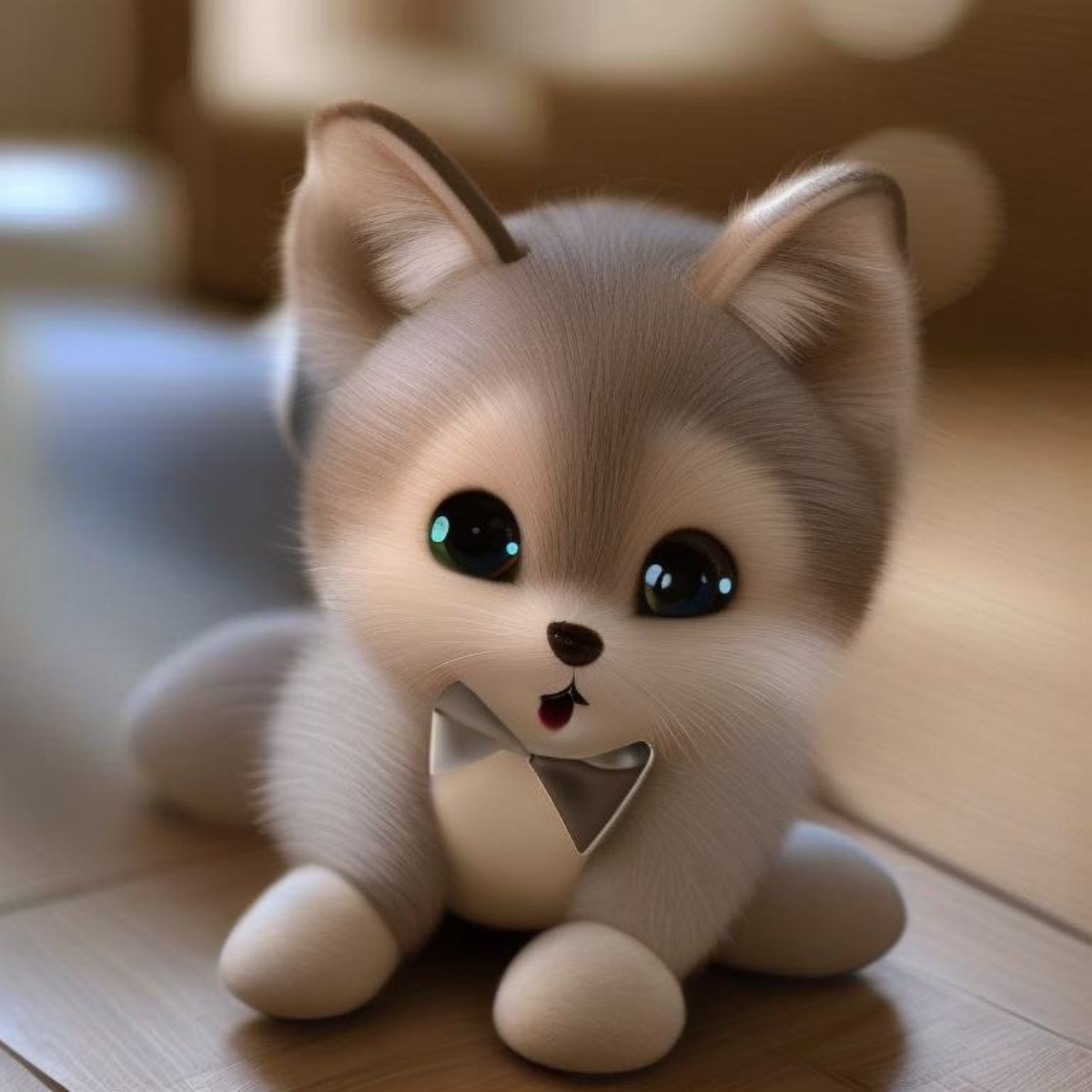} &
\includegraphics[trim=1cm 1cm 1cm 1cm,clip,width=0.2\linewidth]{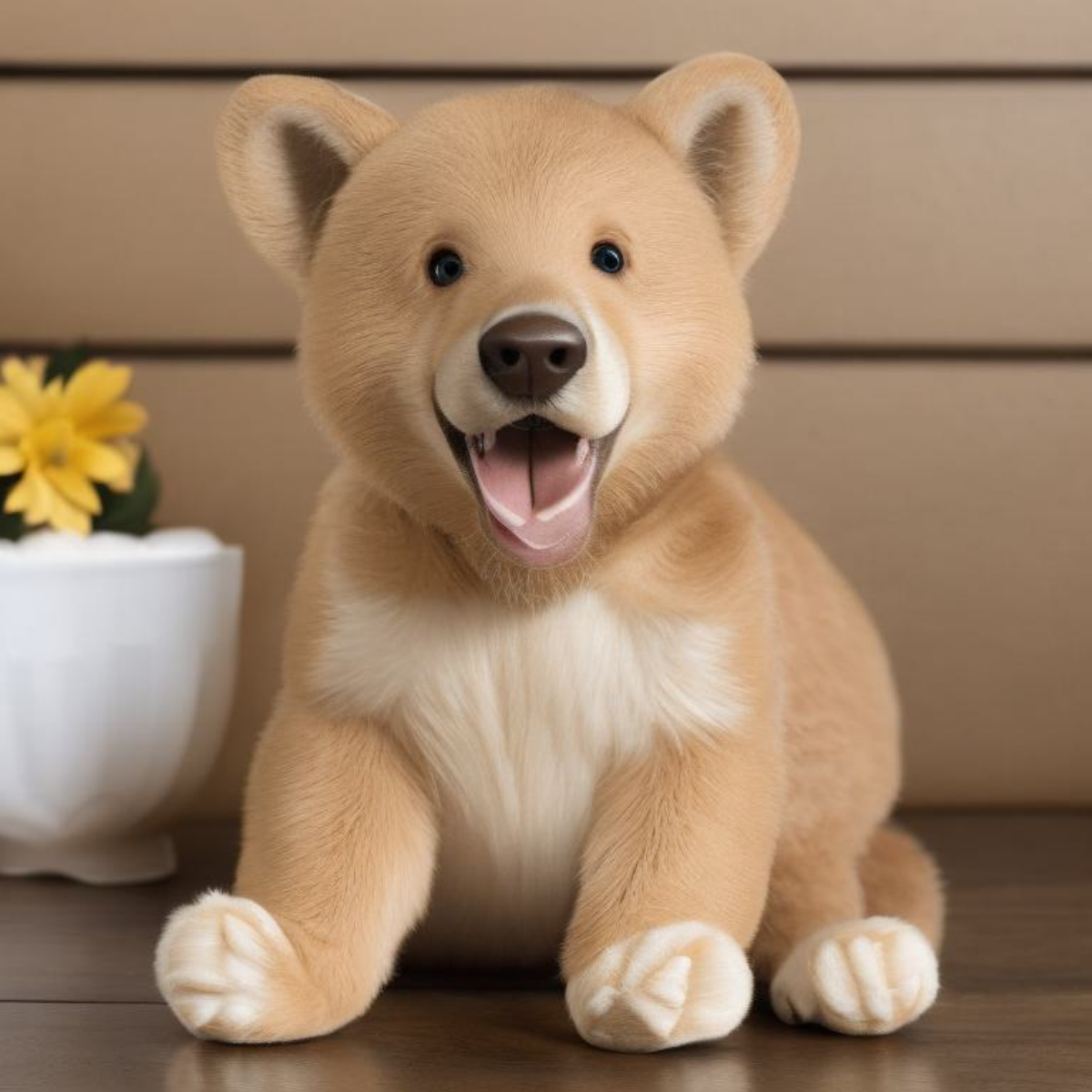} &
\includegraphics[trim=1cm 1cm 1cm 1cm,clip,width=0.2\linewidth]{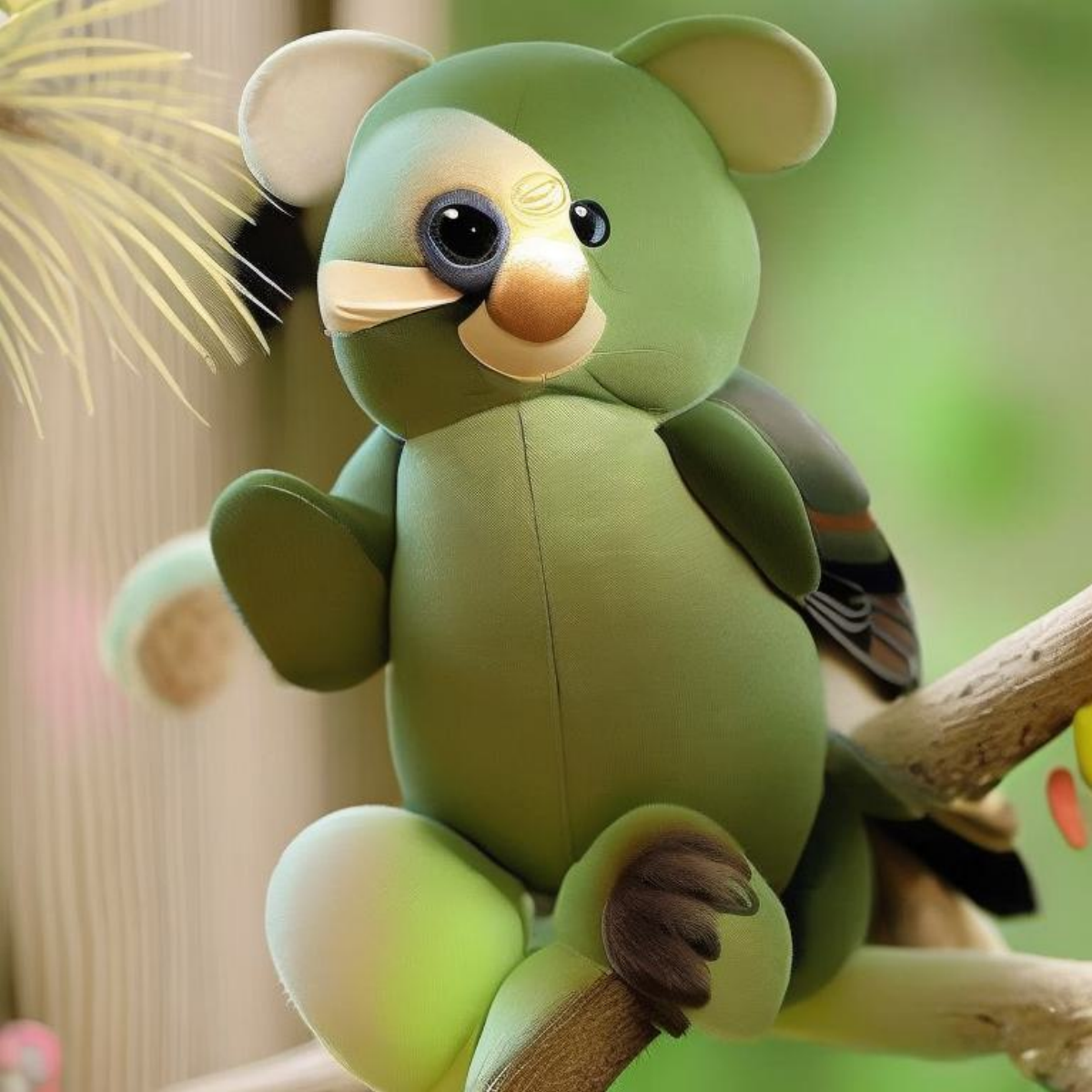} &
\includegraphics[trim=1cm 1cm 1cm 1cm,clip,width=0.2\linewidth]{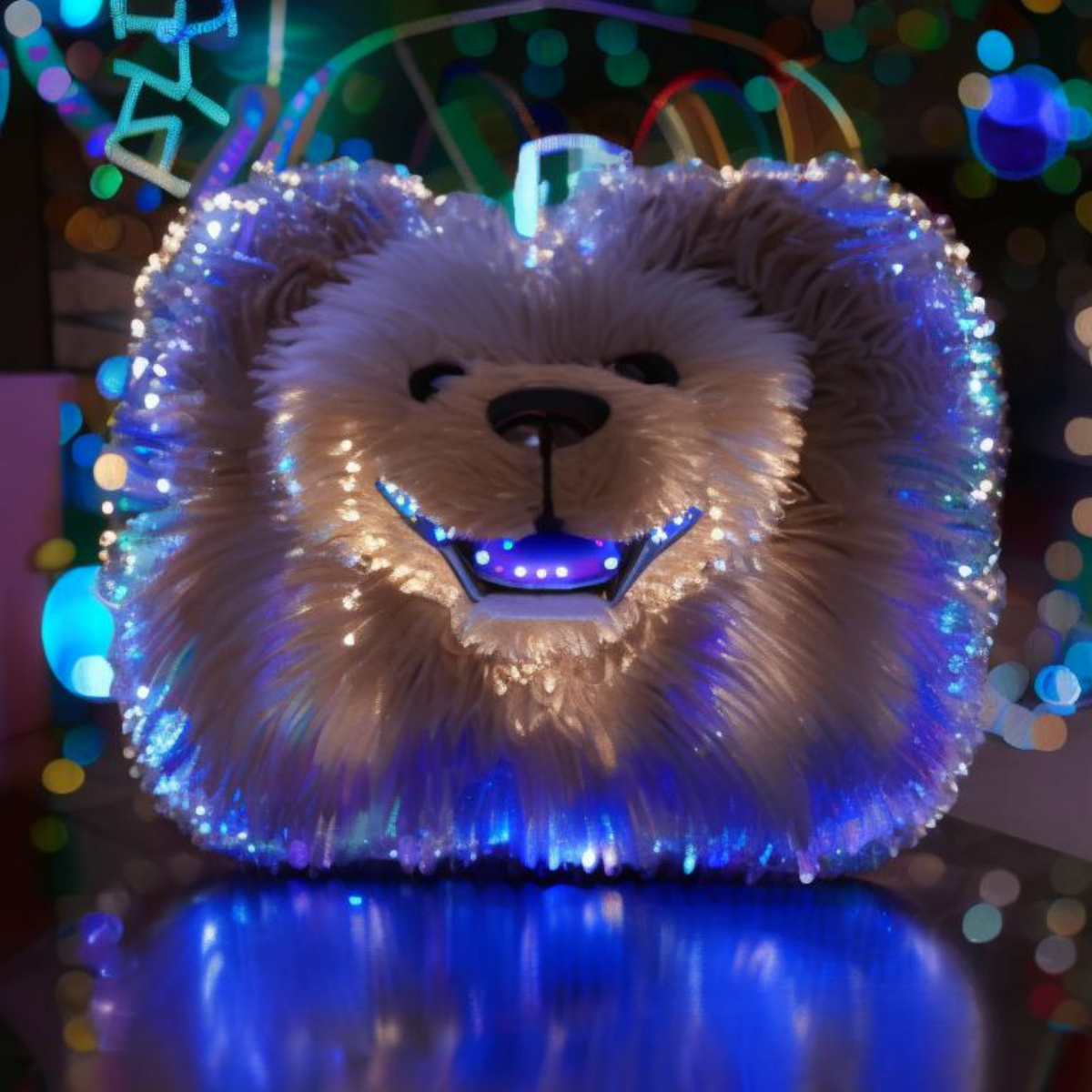} &
\includegraphics[trim=1cm 1cm 1cm 1cm,clip,width=0.2\linewidth]{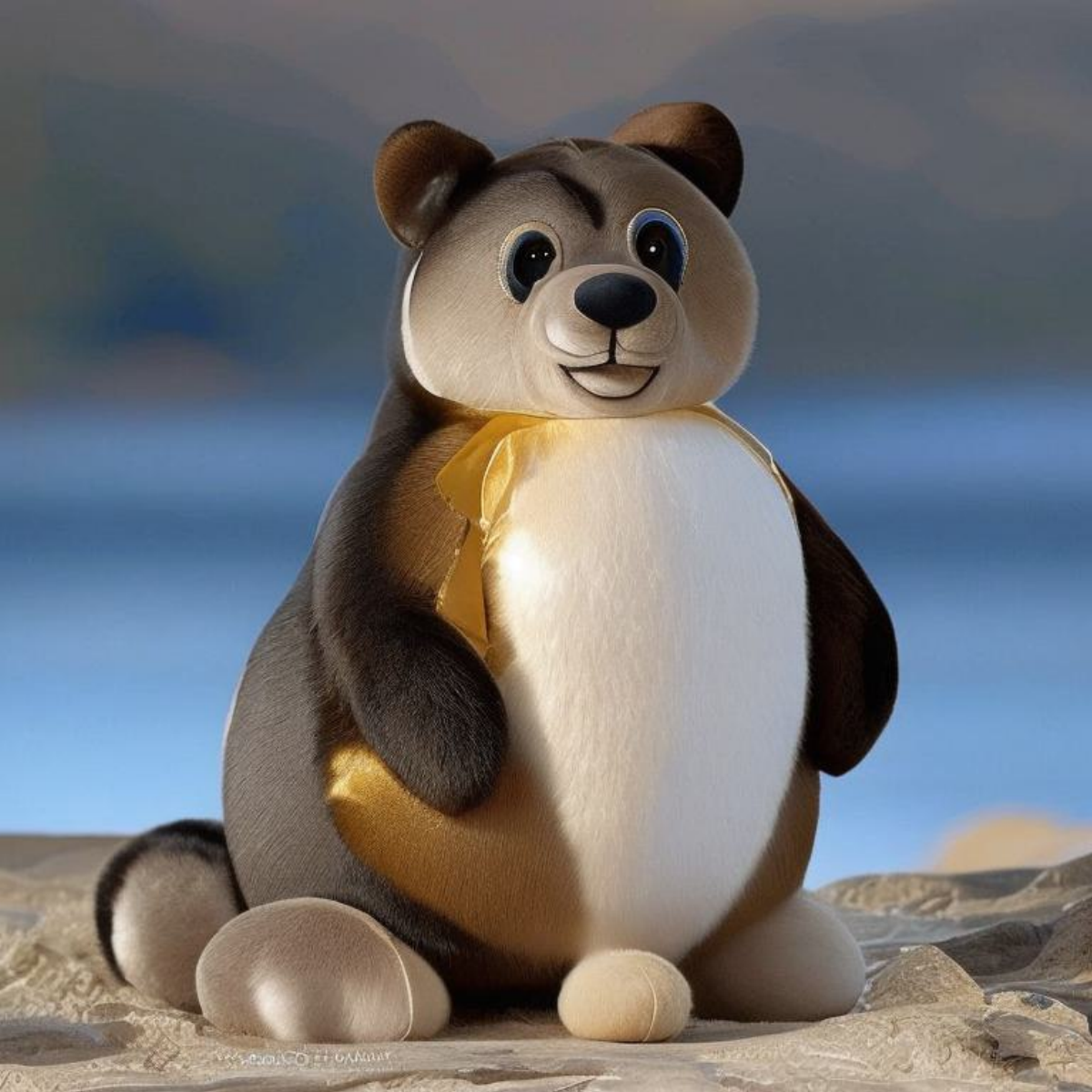} \\
\vspace{1mm} 
\small teddy-cat & \small teddy-dog & \small teddy-parrot & \small teddy-neon & \small teddy-penguin    \\

\includegraphics[trim=1cm 1cm 1cm 1cm,clip,width=0.2\linewidth]{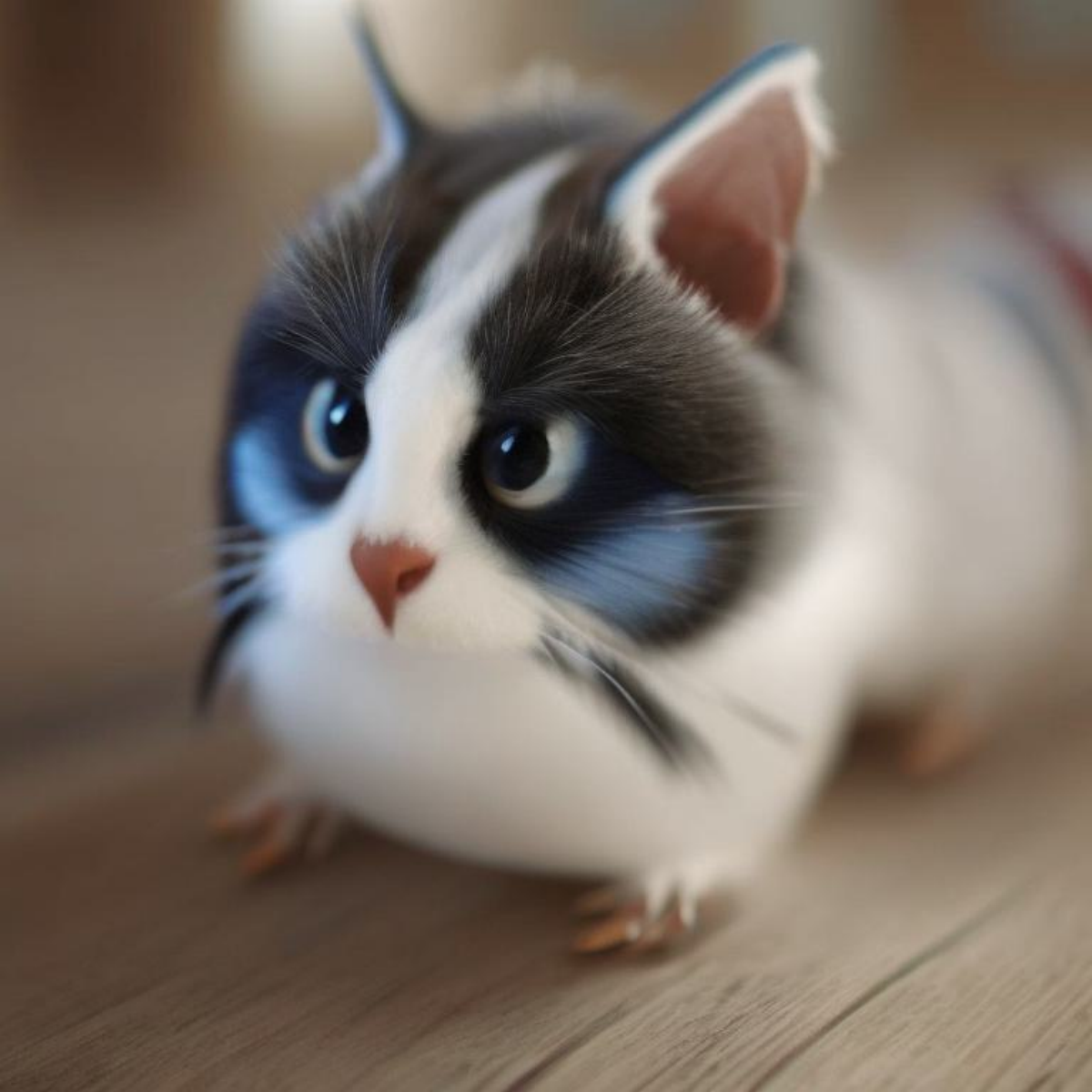} &
\includegraphics[trim=1cm 1cm 1cm 1cm,clip,width=0.2\linewidth]{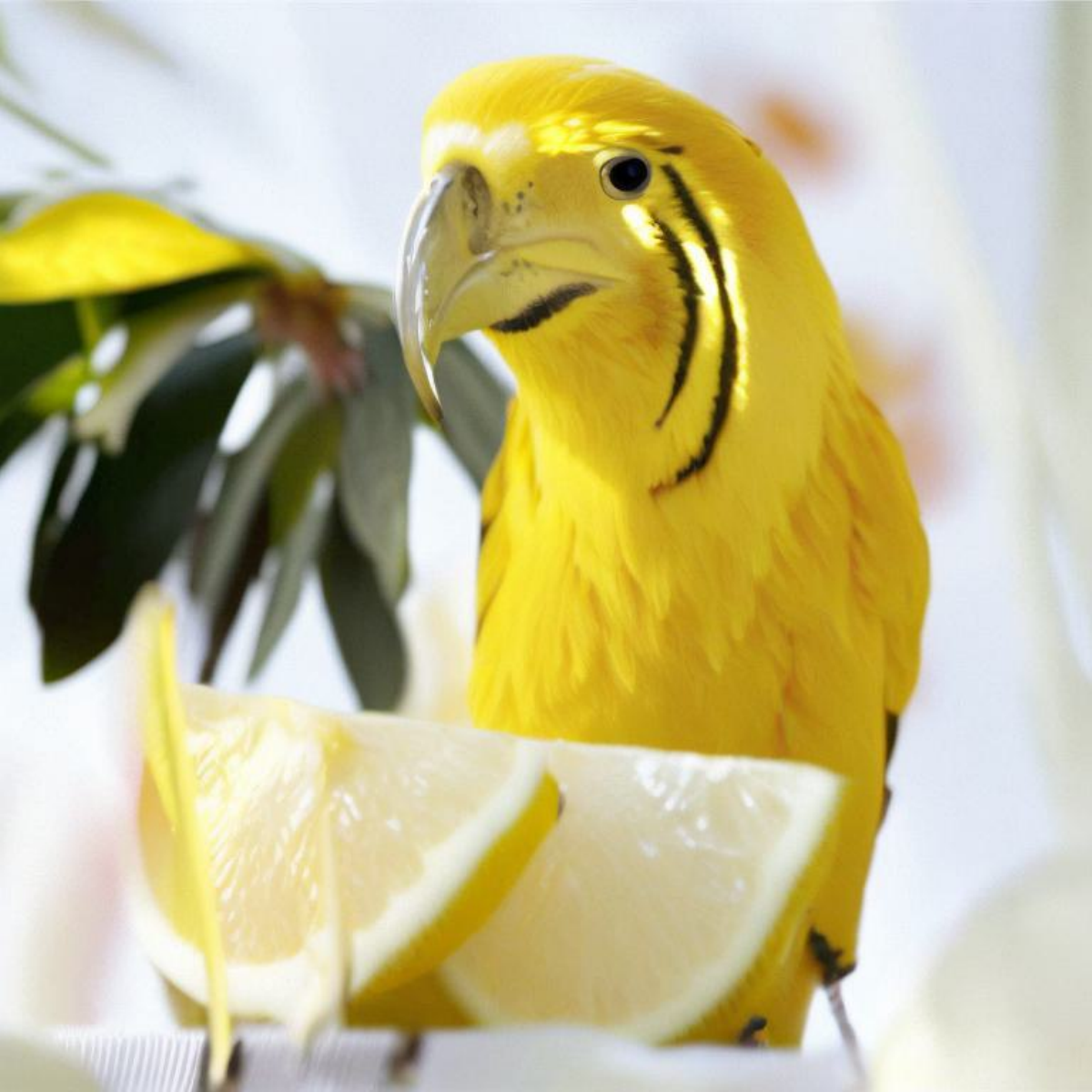} &
\includegraphics[trim=1cm 1cm 1cm 1cm,clip,width=0.2\linewidth]{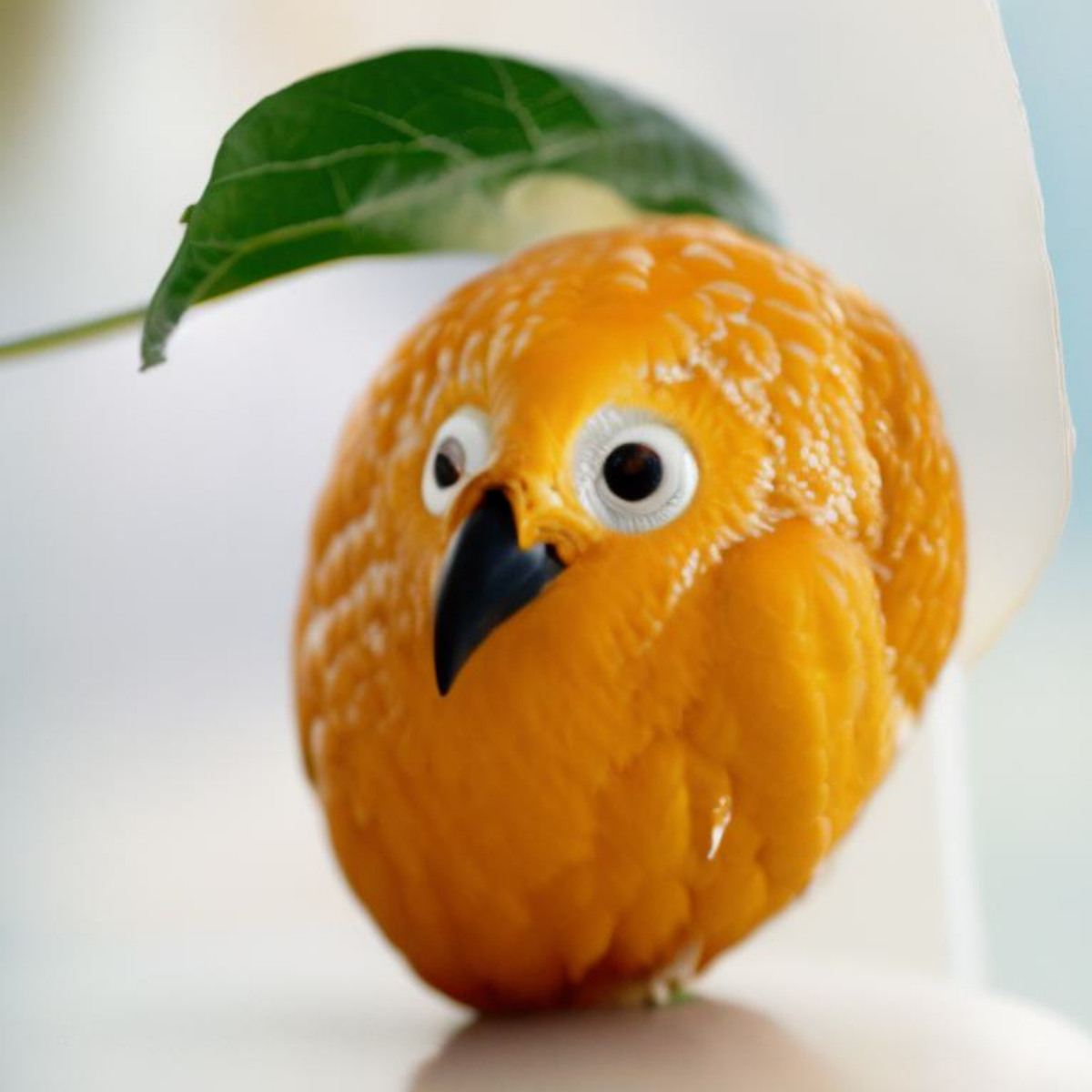} &
\includegraphics[trim=1cm 1cm 1cm 1cm,clip,width=0.2\linewidth]{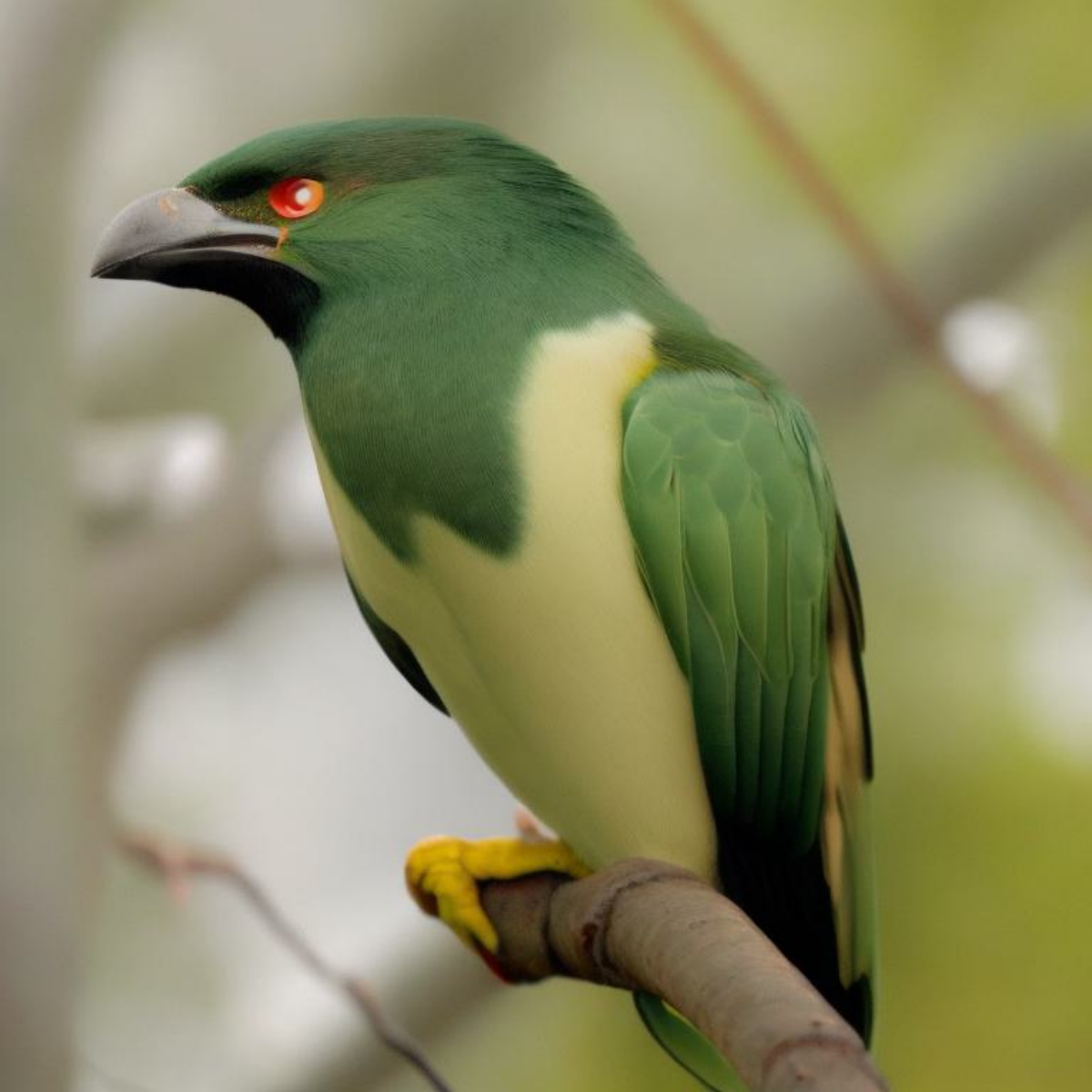} &
\includegraphics[trim=1cm 1cm 1cm 1cm,clip,width=0.2\linewidth]{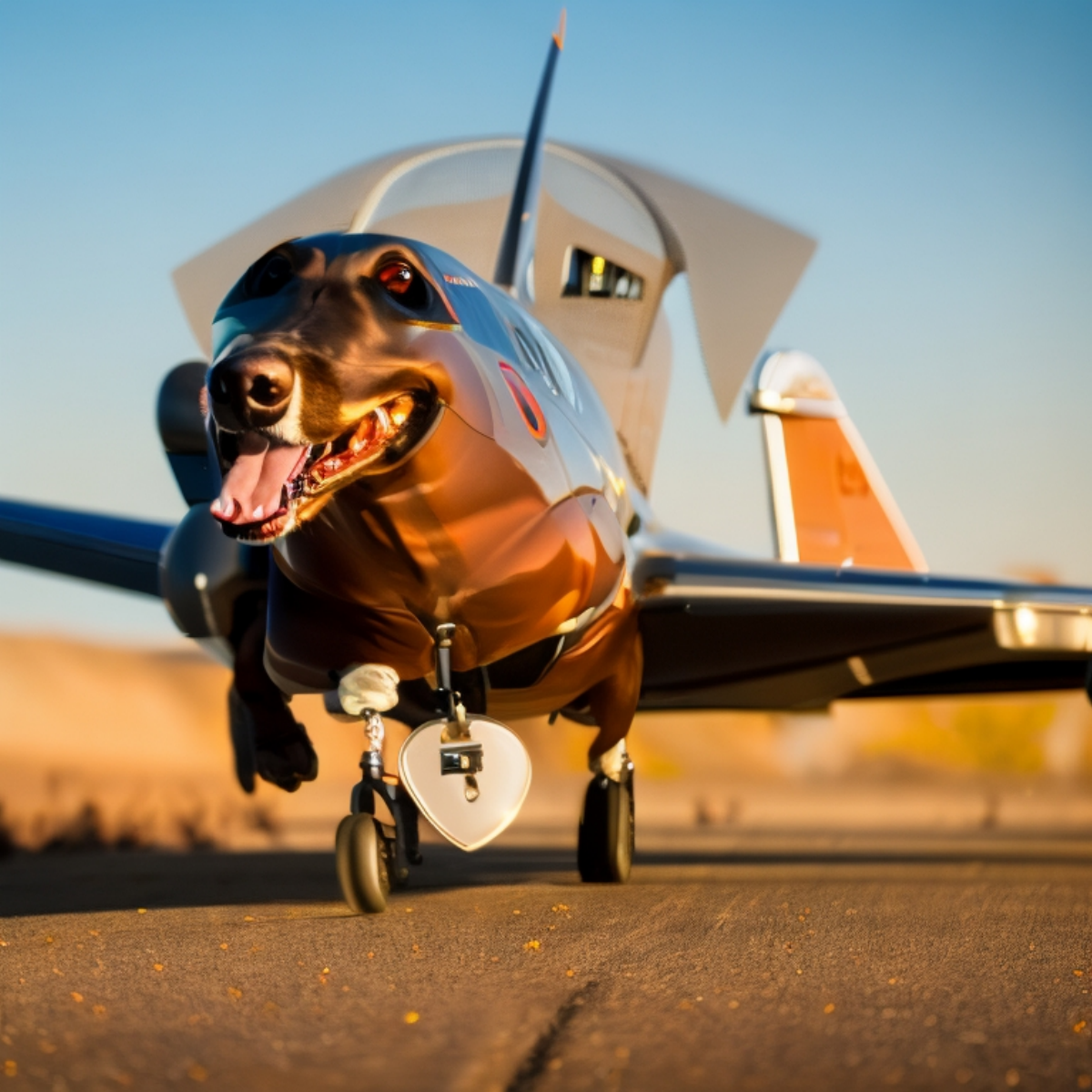} \\
\vspace{1mm} 
\small eagle-cat & \small eagle-lemon & \small eagle-orange & \small eagle-parrot & \small dog-plane  \\

\includegraphics[trim=1cm 1cm 1cm 1cm,clip,width=0.2\linewidth]{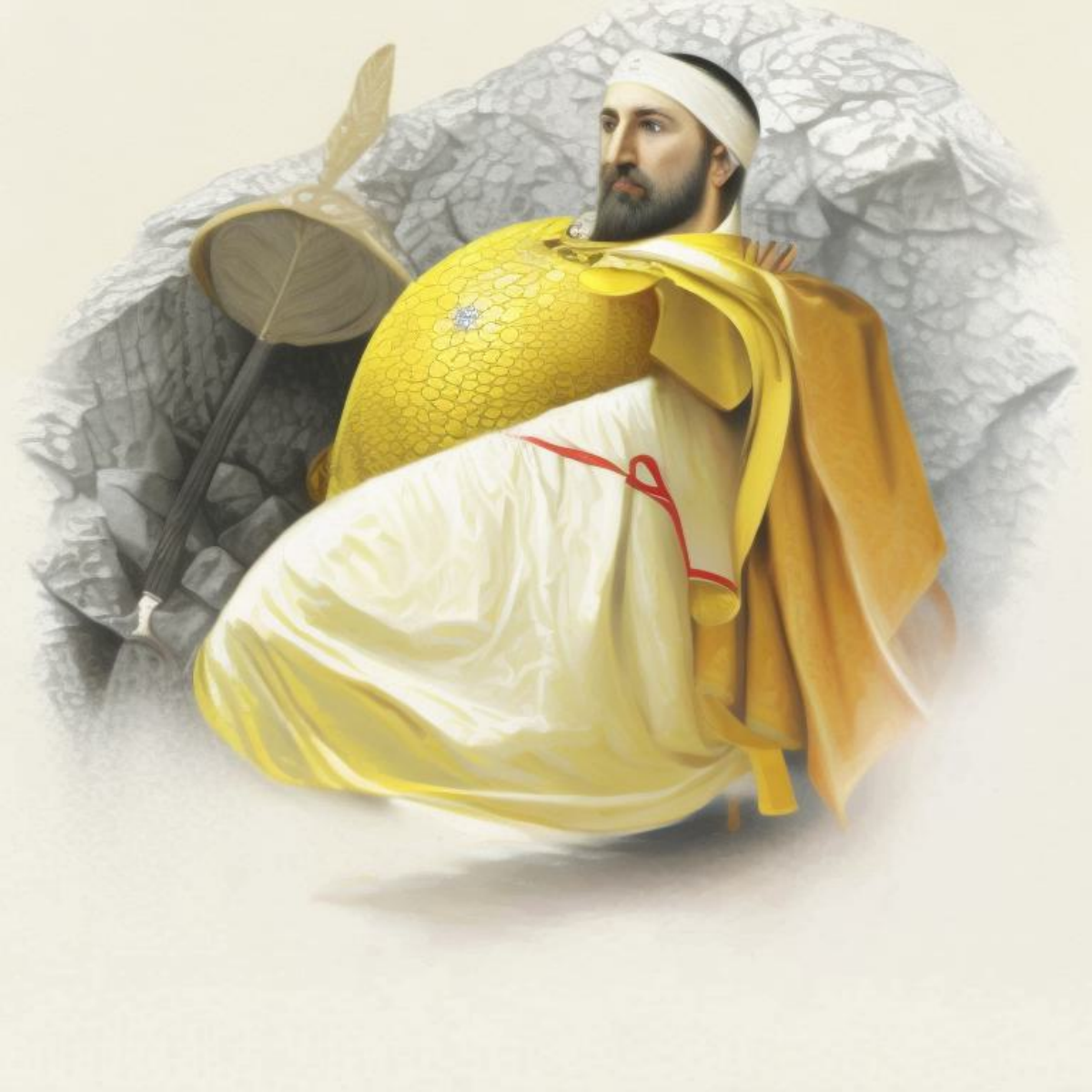} &
\includegraphics[trim=1cm 1cm 1cm 1cm,clip,width=0.2\linewidth]{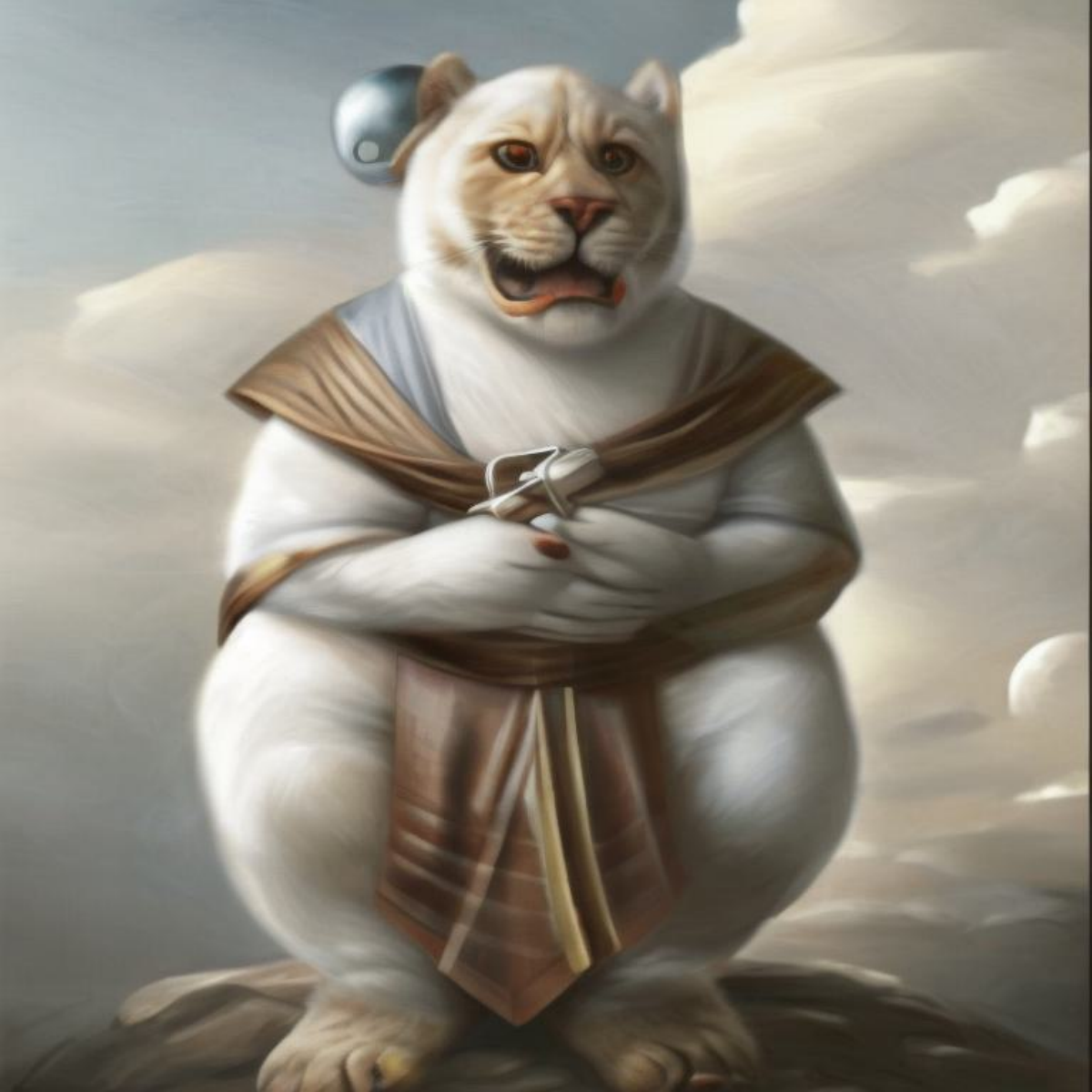} &
\includegraphics[trim=1cm 1cm 1cm 1cm,clip,width=0.2\linewidth]{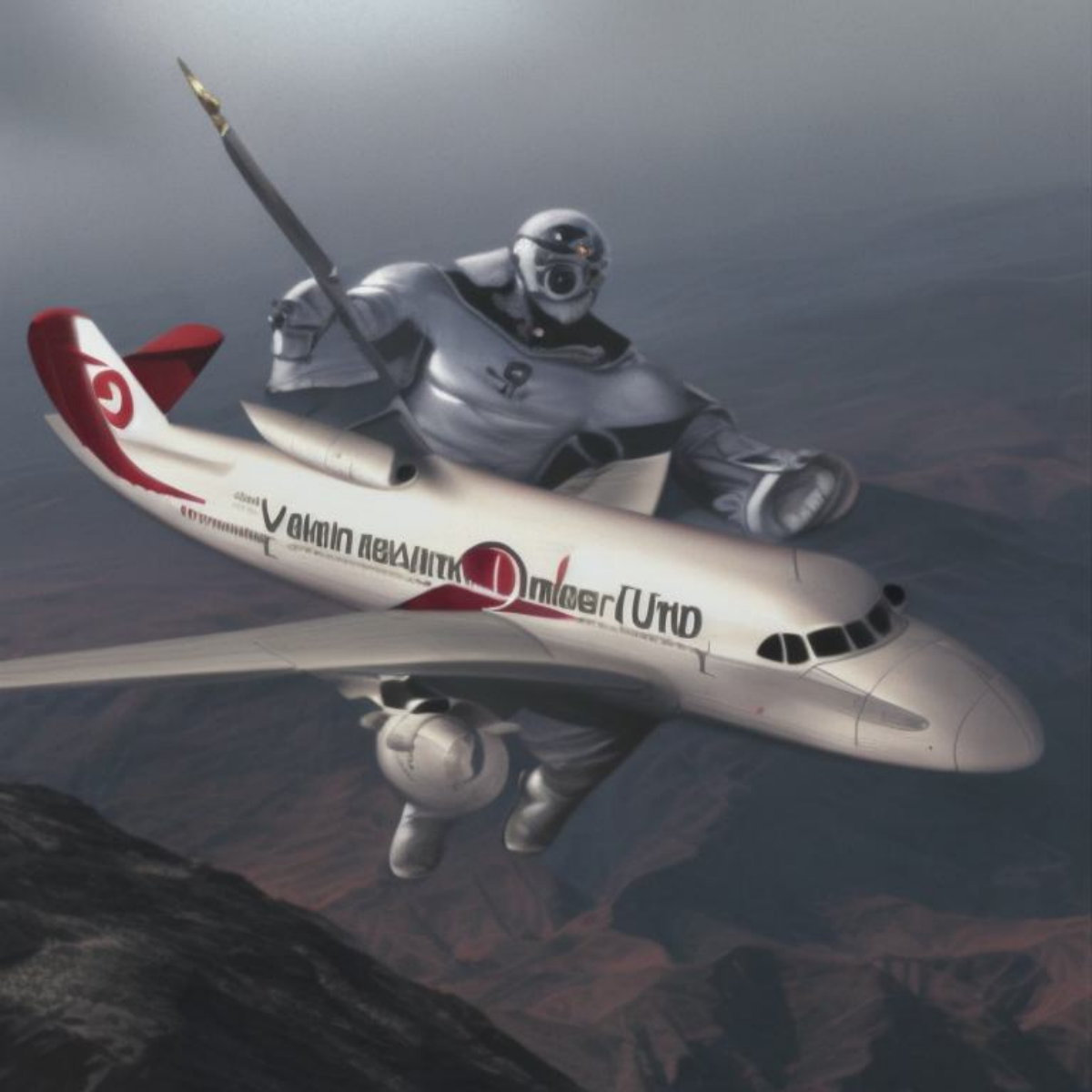} &
\includegraphics[trim=1cm 1cm 1cm 1cm,clip,width=0.2\linewidth]{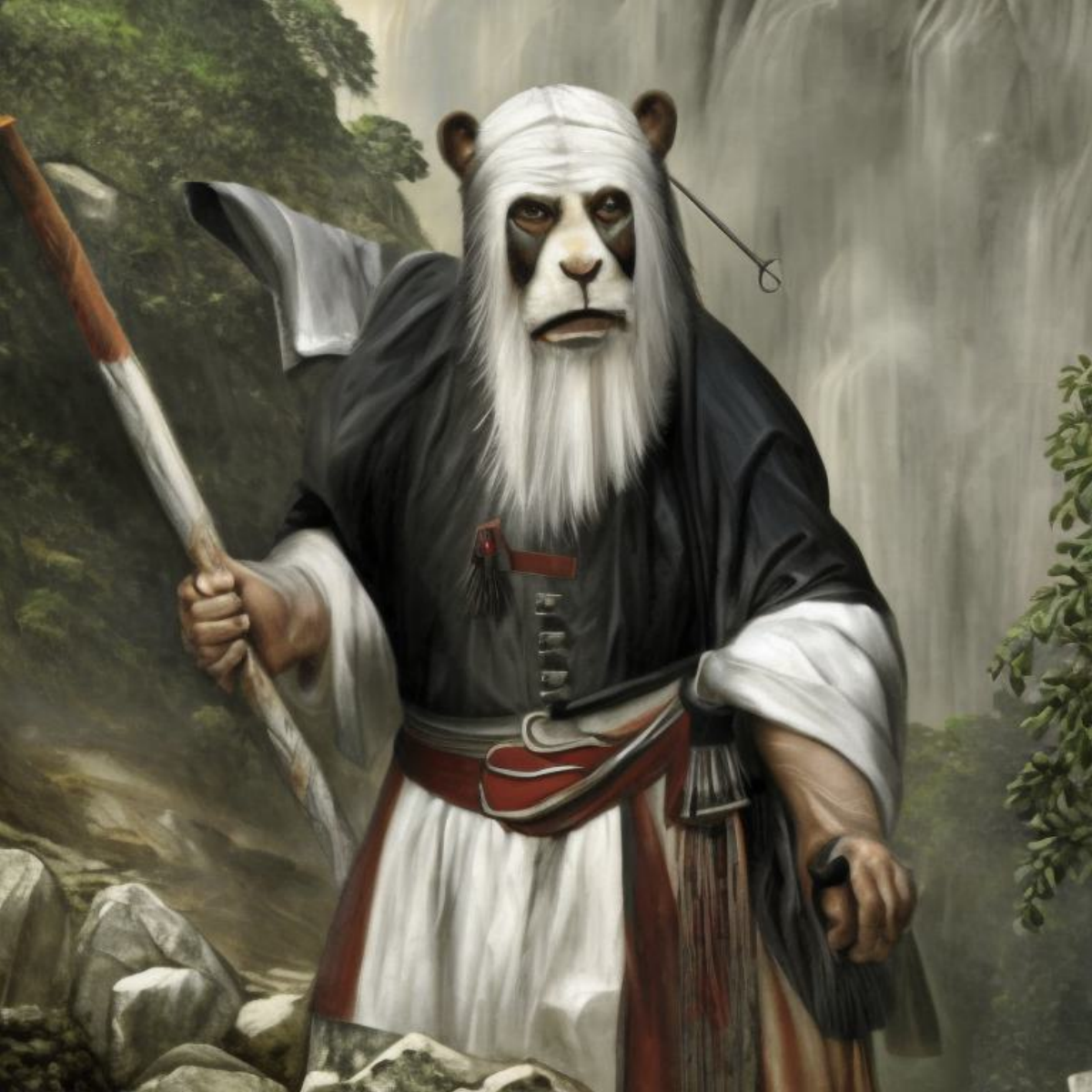} &
\includegraphics[trim=1cm 1cm 1cm 1cm,clip,width=0.2\linewidth]{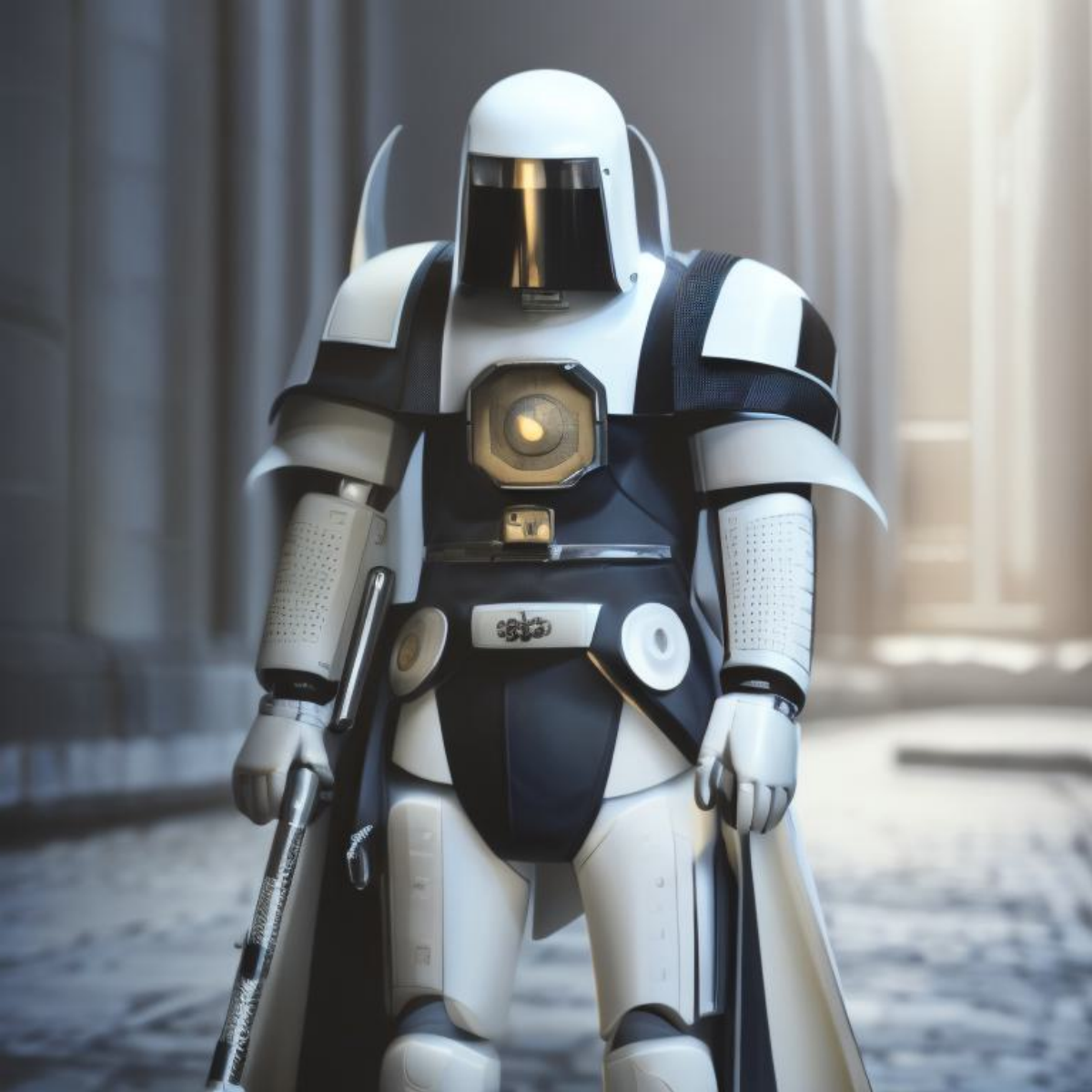} \\
\vspace{1mm} 
\small knight-lemon & \small knight-lion & \small knight-plane & \small knight-panda& \small knight-robot   \\

\end{tabular}}
\caption{More blending results. Given two concepts, our approach can blend them into a novel object that the model has never seen before, generating high-quality images.}
\label{more_blending_results}
\end{figure*}

\begin{figure*}[ht]
\centering
\resizebox{0.995\linewidth}{!}{
\setlength{\tabcolsep}{0mm}
\renewcommand{\arraystretch}{0.01}
\begin{tabular}{ccccc}

\includegraphics[trim=1cm 1cm 1cm 1cm,clip,width=0.2\linewidth]{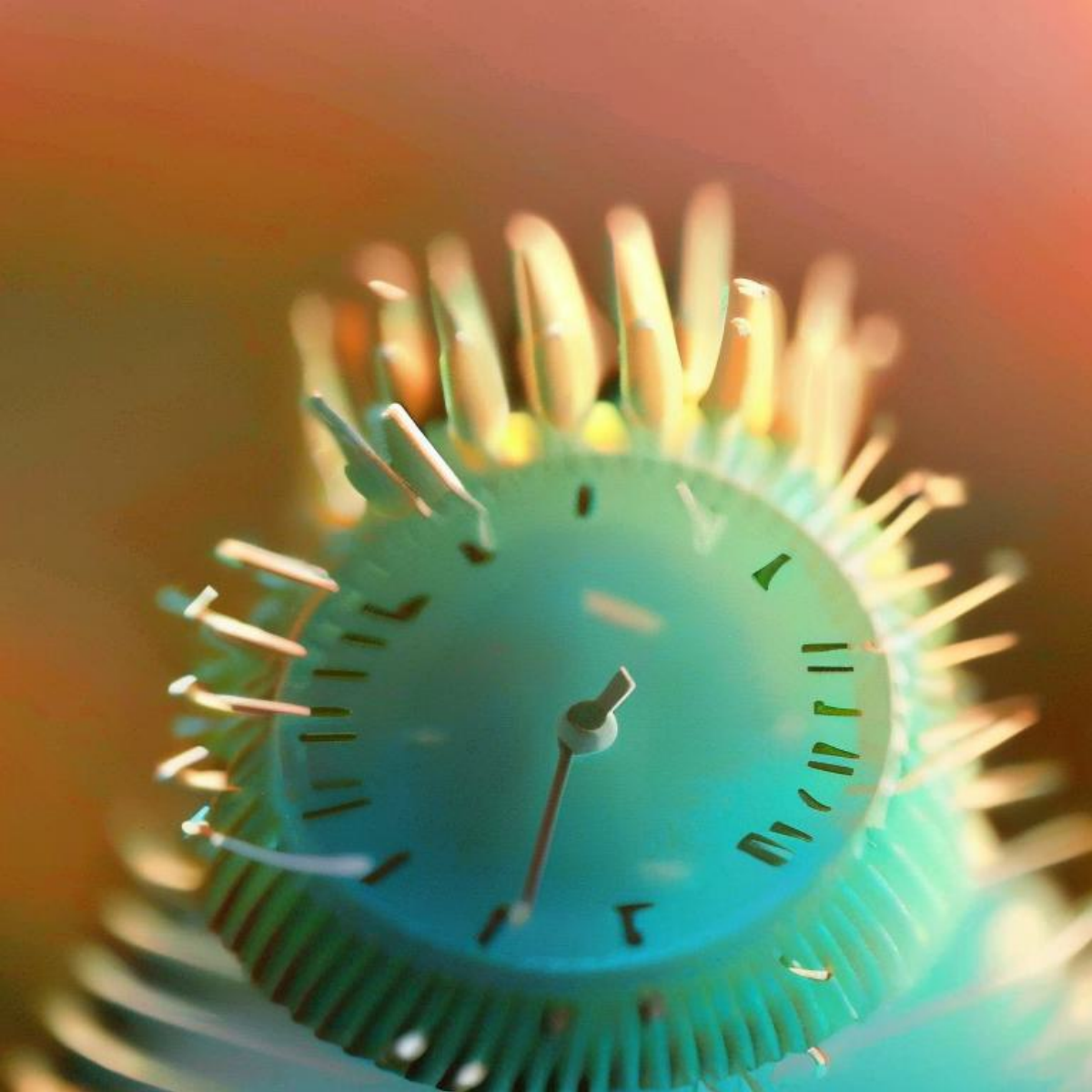} &
\includegraphics[trim=1cm 1cm 1cm 1cm,clip,width=0.2\linewidth]{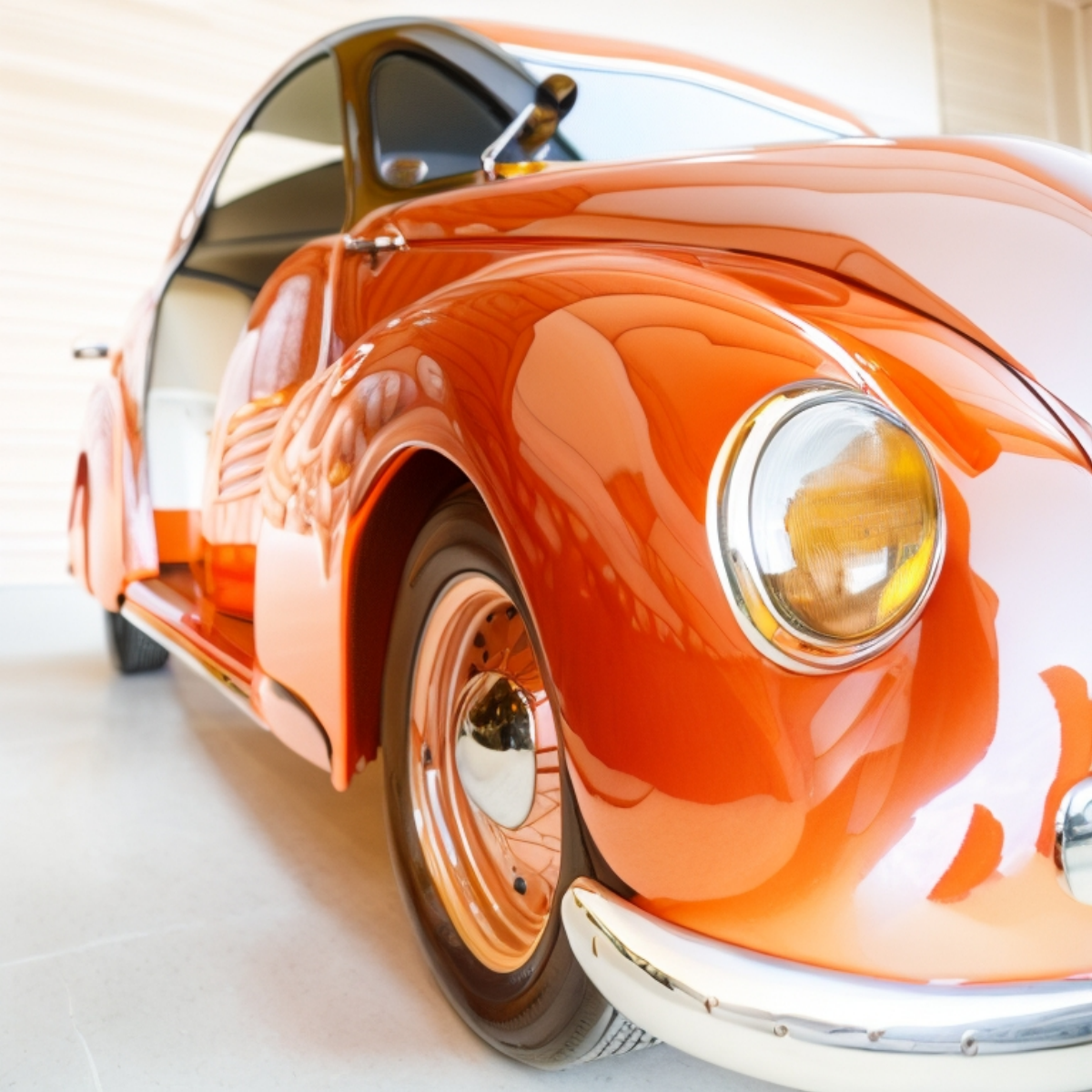} &
\includegraphics[trim=1cm 1cm 1cm 1cm,clip,width=0.2\linewidth]{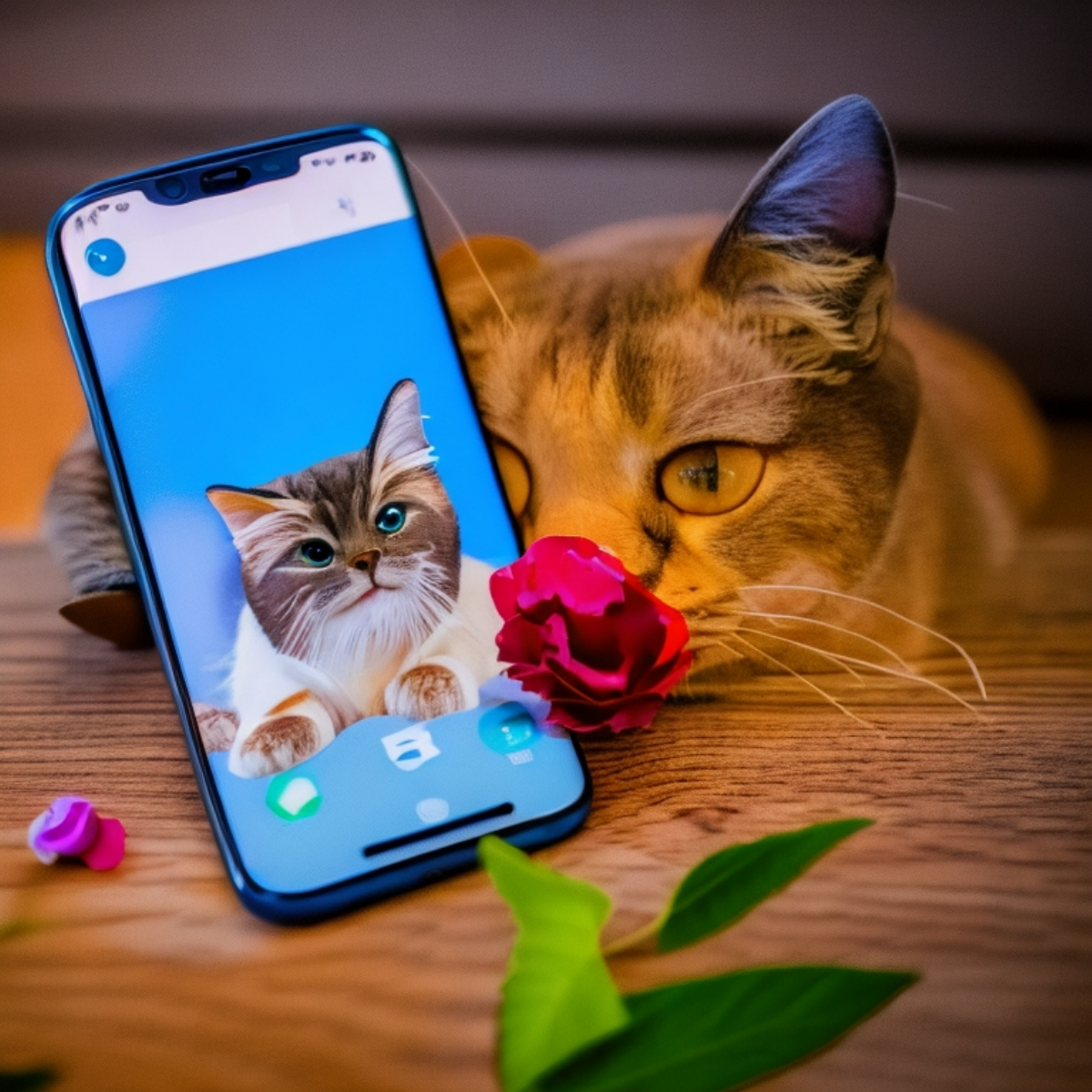} &
\includegraphics[trim=1cm 1cm 1cm 1cm,clip,width=0.2\linewidth]{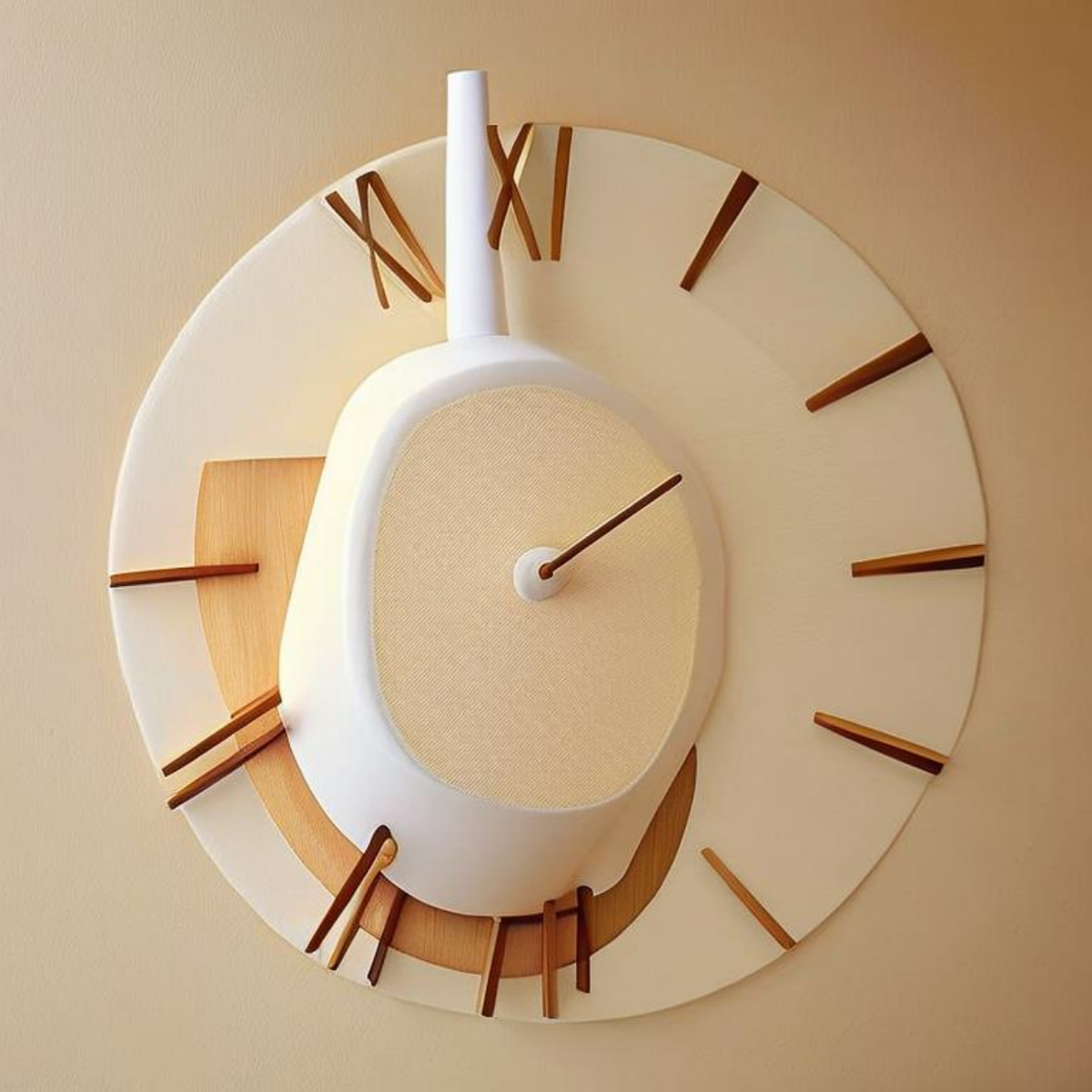} &
\includegraphics[trim=1cm 1cm 1cm 1cm,clip,width=0.2\linewidth]{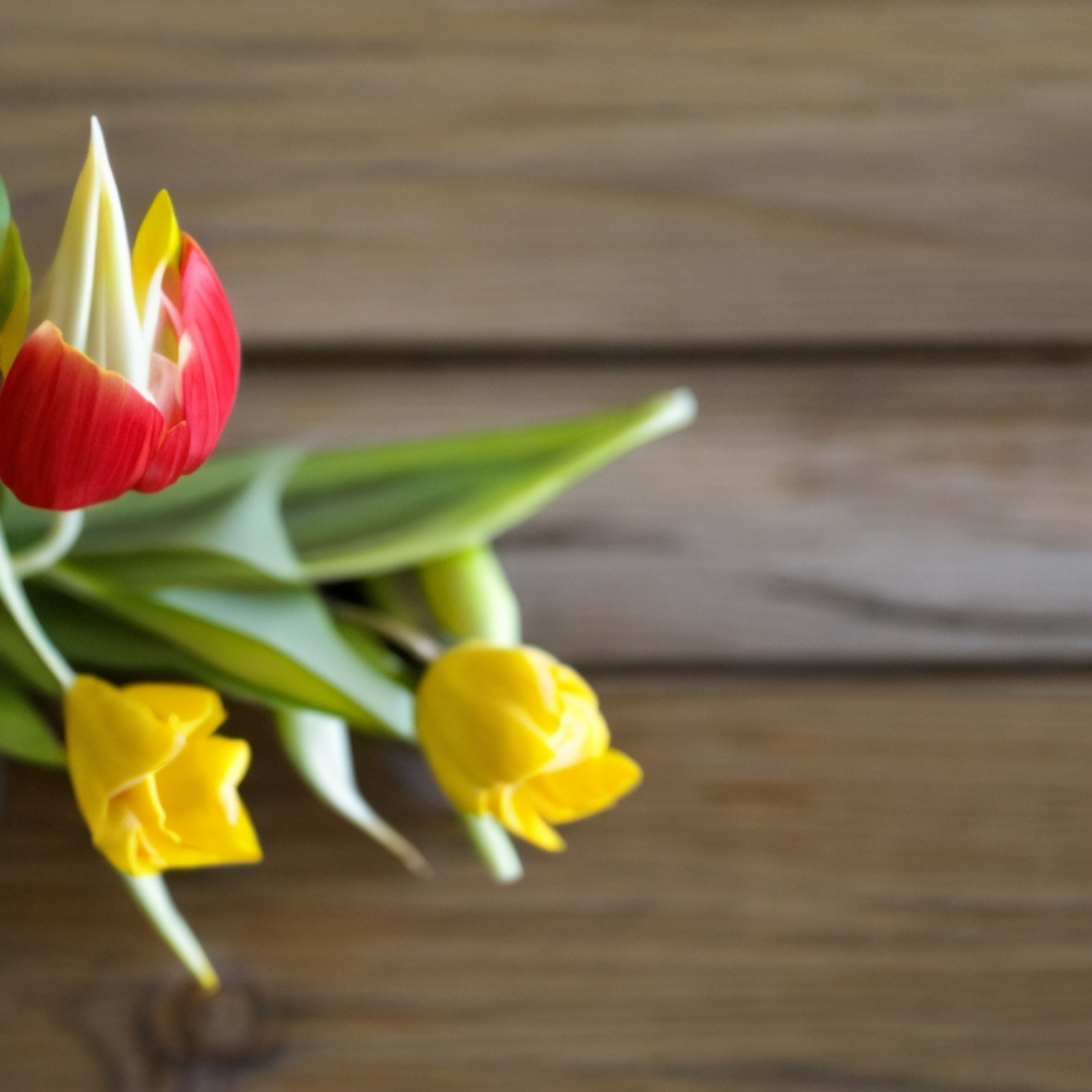} \\
\vspace{1mm} 
\small clock-cactus & \small orange-car & \small cat-phone & \small clock-ice cream & \small lily-lemon    \\
\includegraphics[trim=1cm 1cm 1cm 1cm,clip,width=0.2\linewidth]{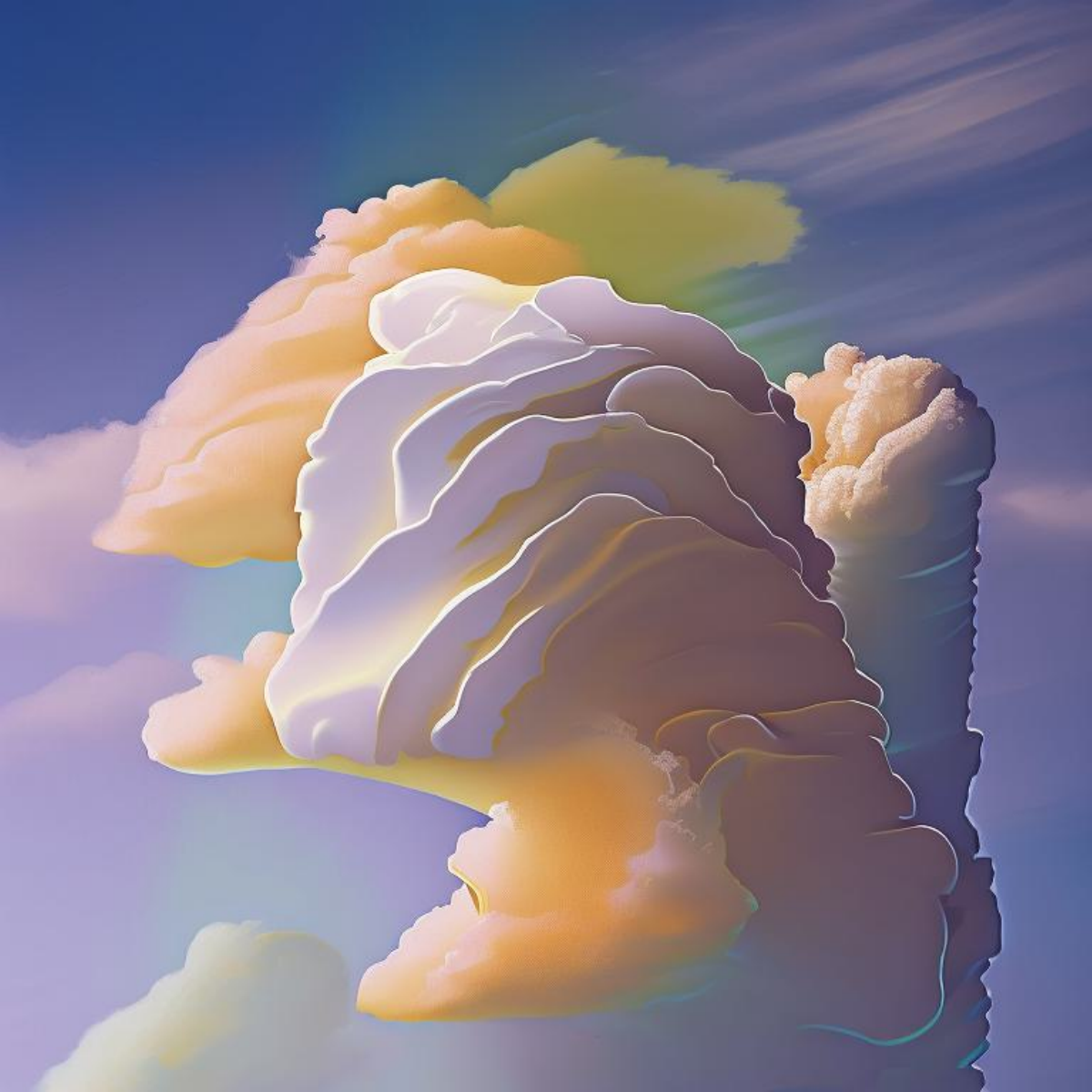} &
\includegraphics[trim=1cm 1cm 1cm 1cm,clip,width=0.2\linewidth]{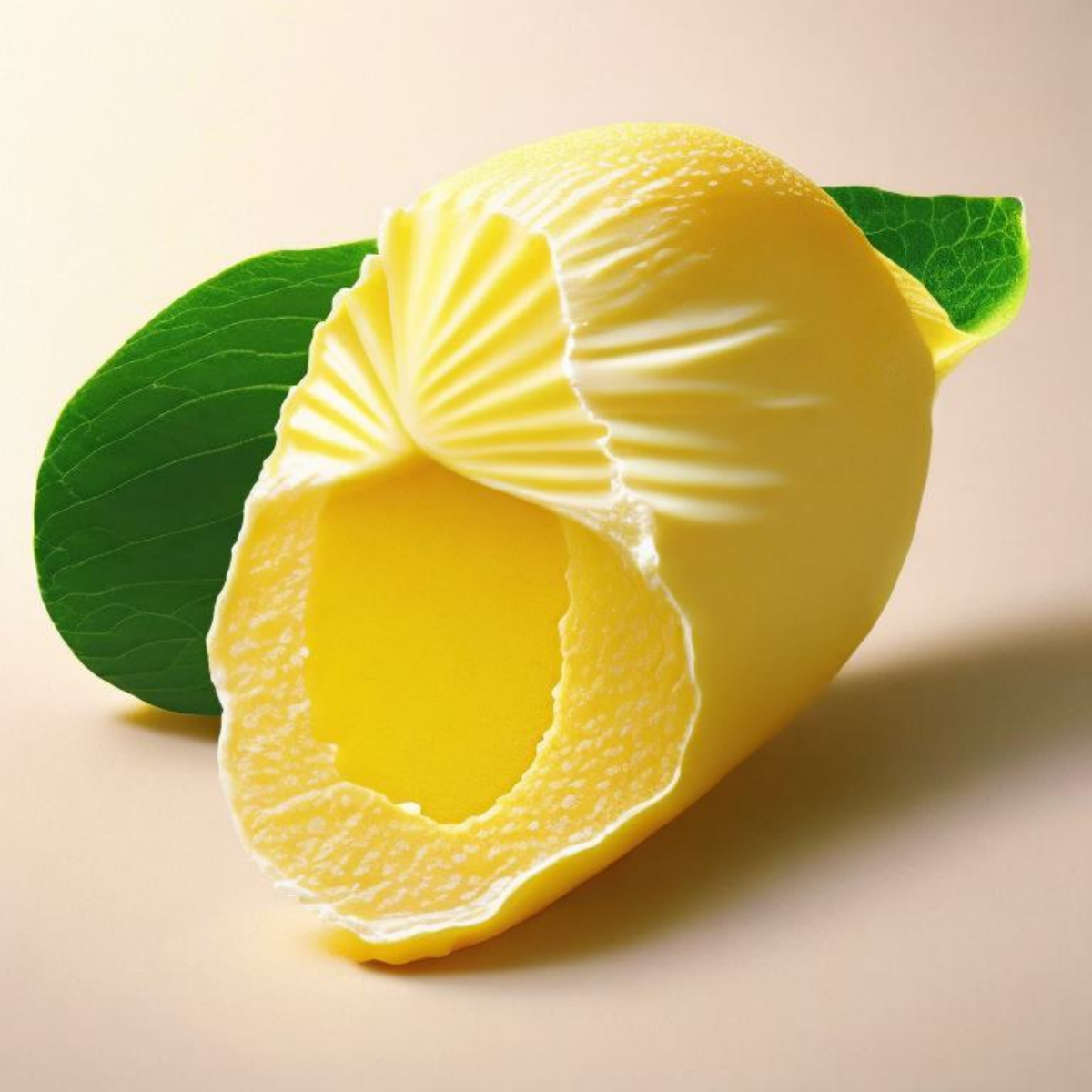} &
\includegraphics[trim=1cm 1cm 1cm 1cm,clip,width=0.2\linewidth]{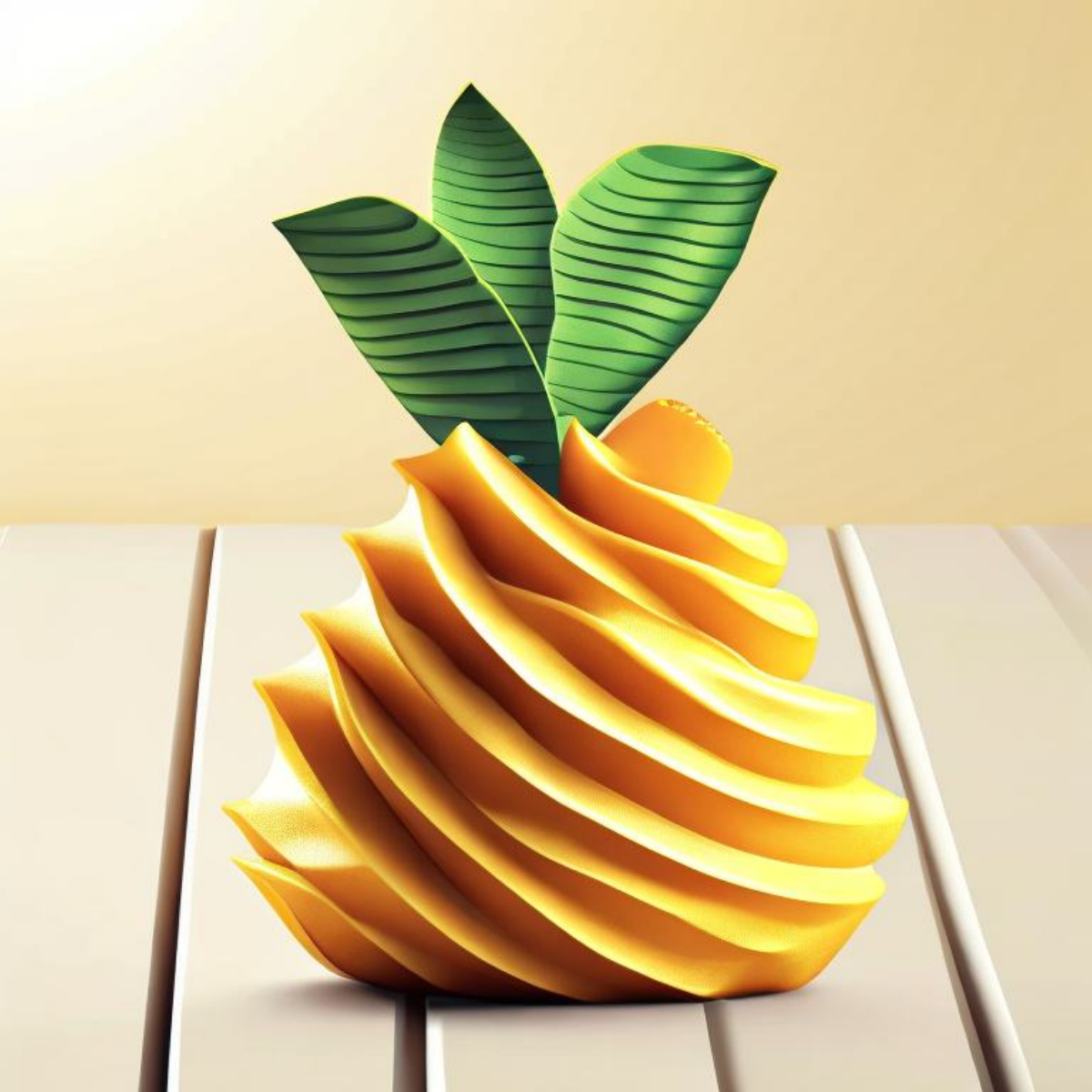} &
\includegraphics[trim=1cm 1cm 1cm 1cm,clip,width=0.2\linewidth]{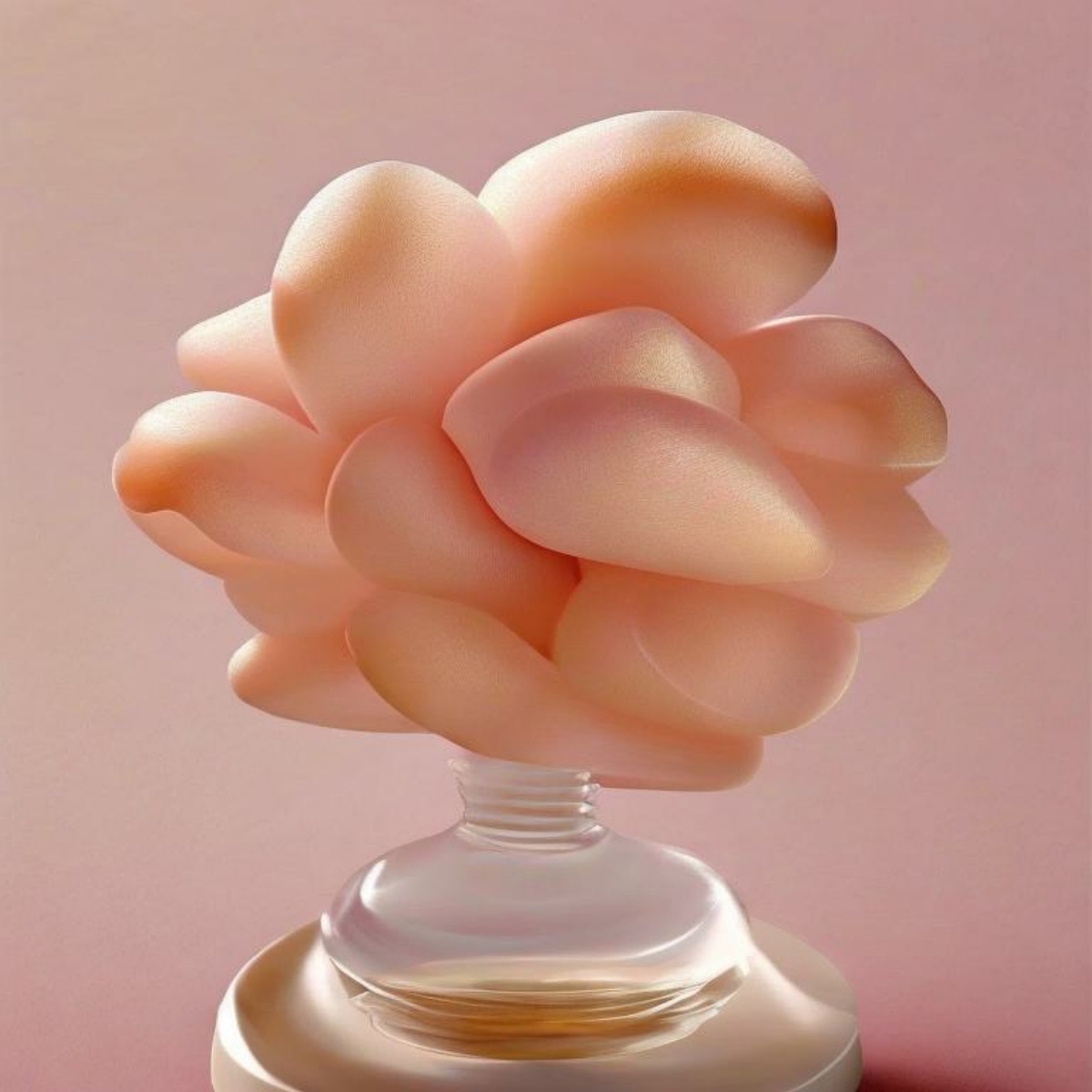} &
\includegraphics[trim=1cm 1cm 1cm 1cm,clip,width=0.2\linewidth]{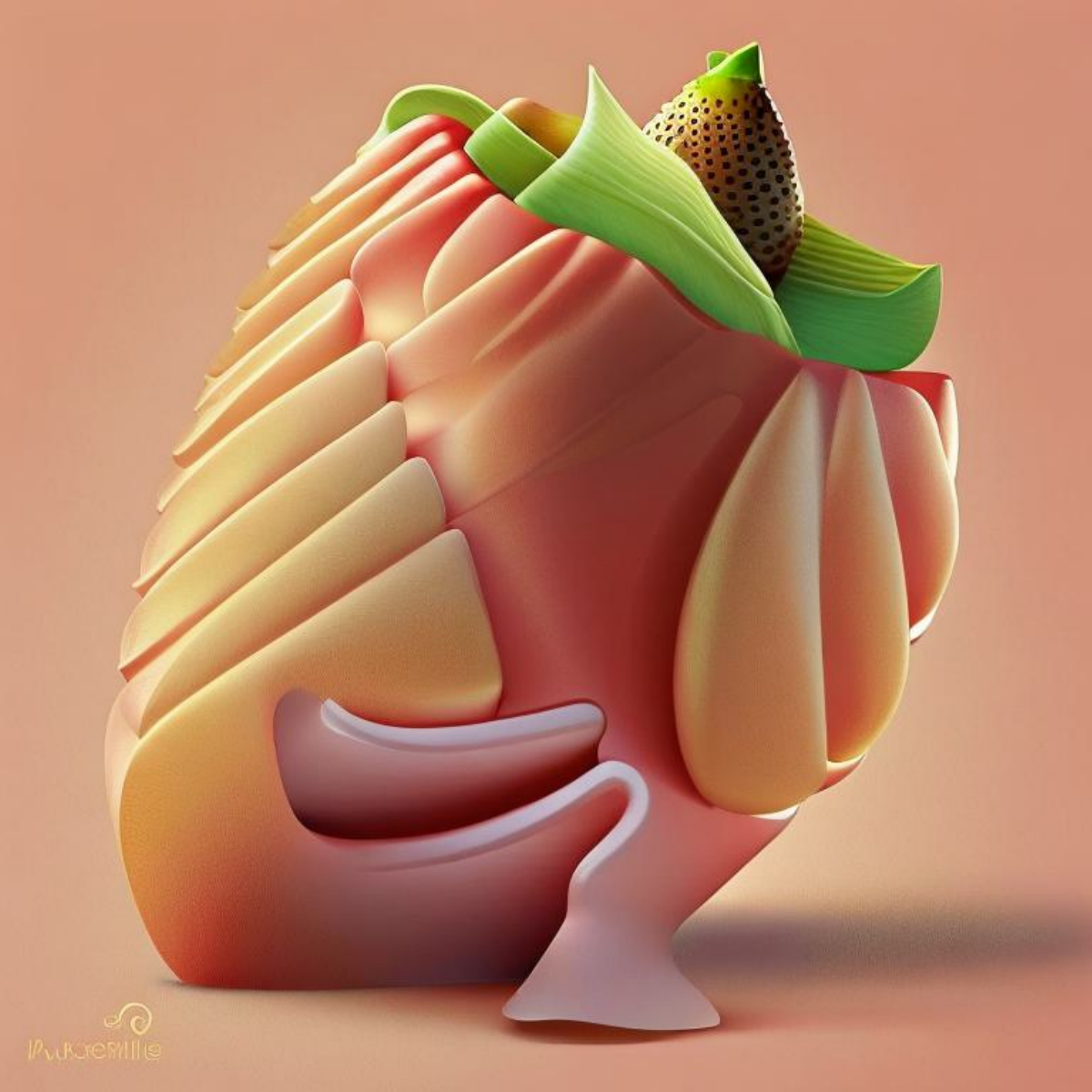} \\
\vspace{1mm} 
\small ice cream-cat & \small ice cream-lemon & \small ice cream-pineapple & \small ice cream-sakura & \small ice cream-strawberry    \\

\includegraphics[trim=1cm 1cm 1cm 1cm,clip,width=0.2\linewidth]{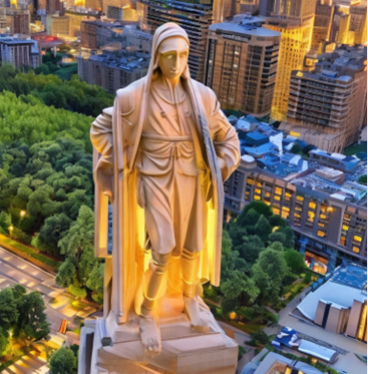} &
\includegraphics[trim=1cm 1cm 1cm 1cm,clip,width=0.2\linewidth]{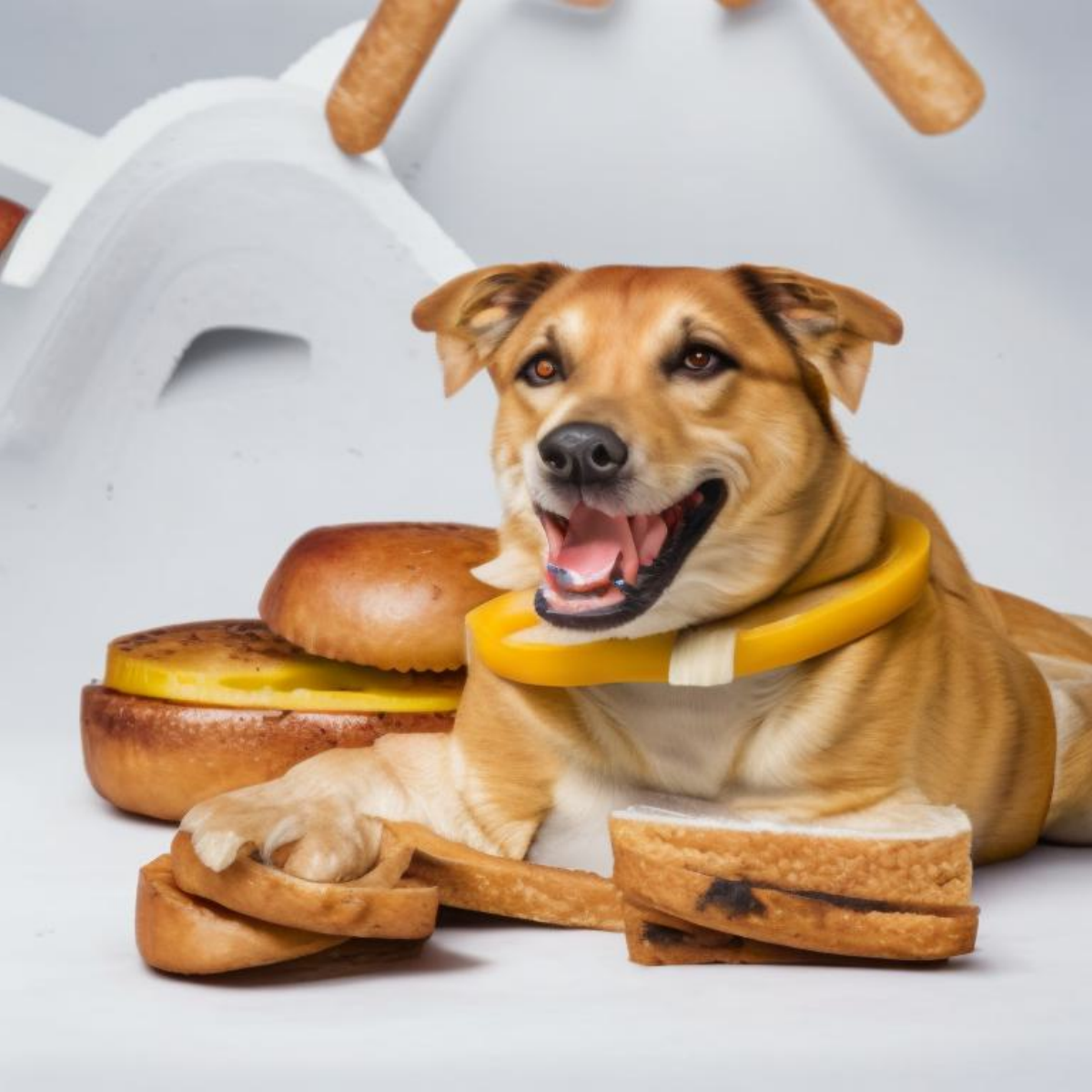} &
\includegraphics[trim=1cm 1cm 1cm 1cm,clip,width=0.2\linewidth]{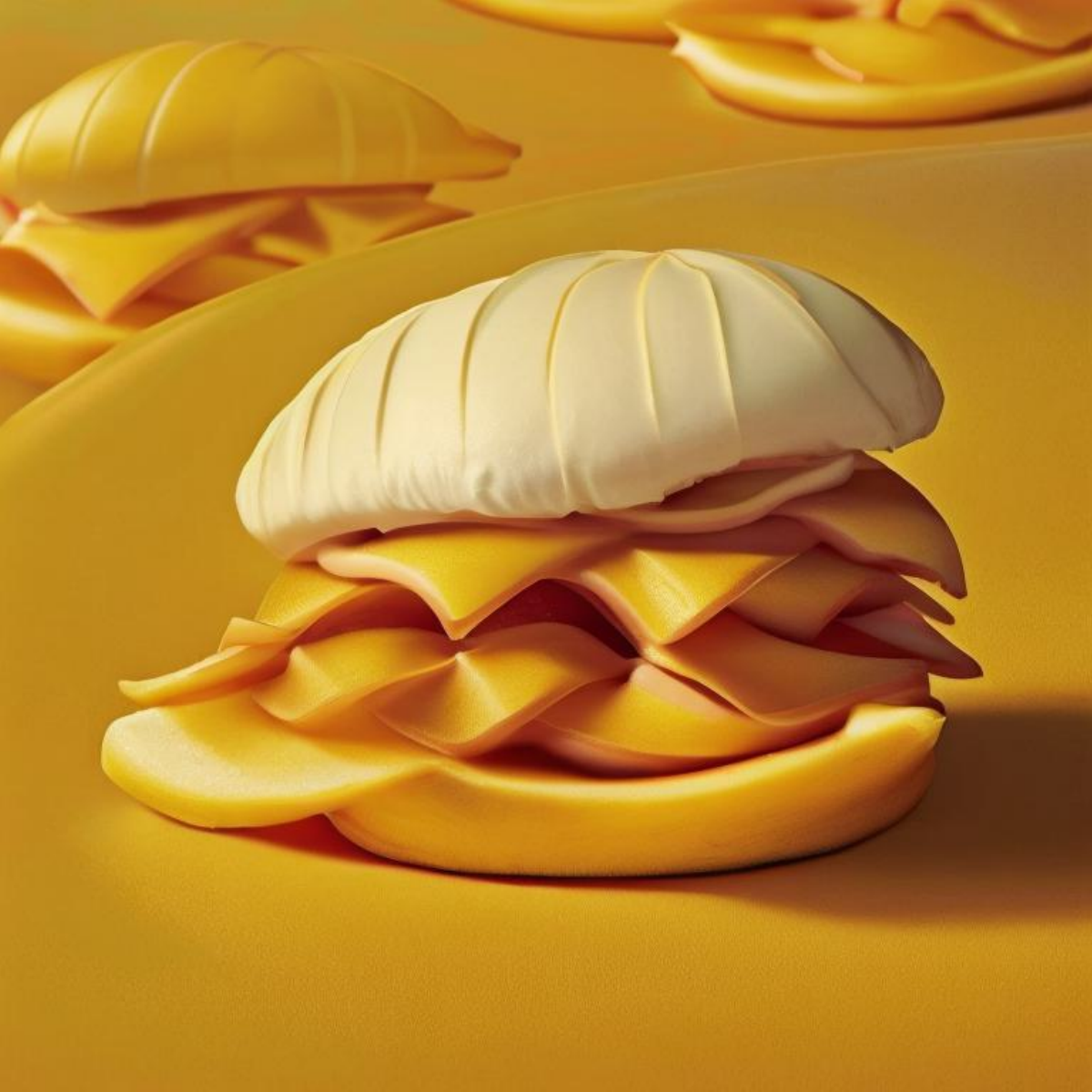} &
\includegraphics[trim=1cm 1cm 1cm 1cm,clip,width=0.2\linewidth]{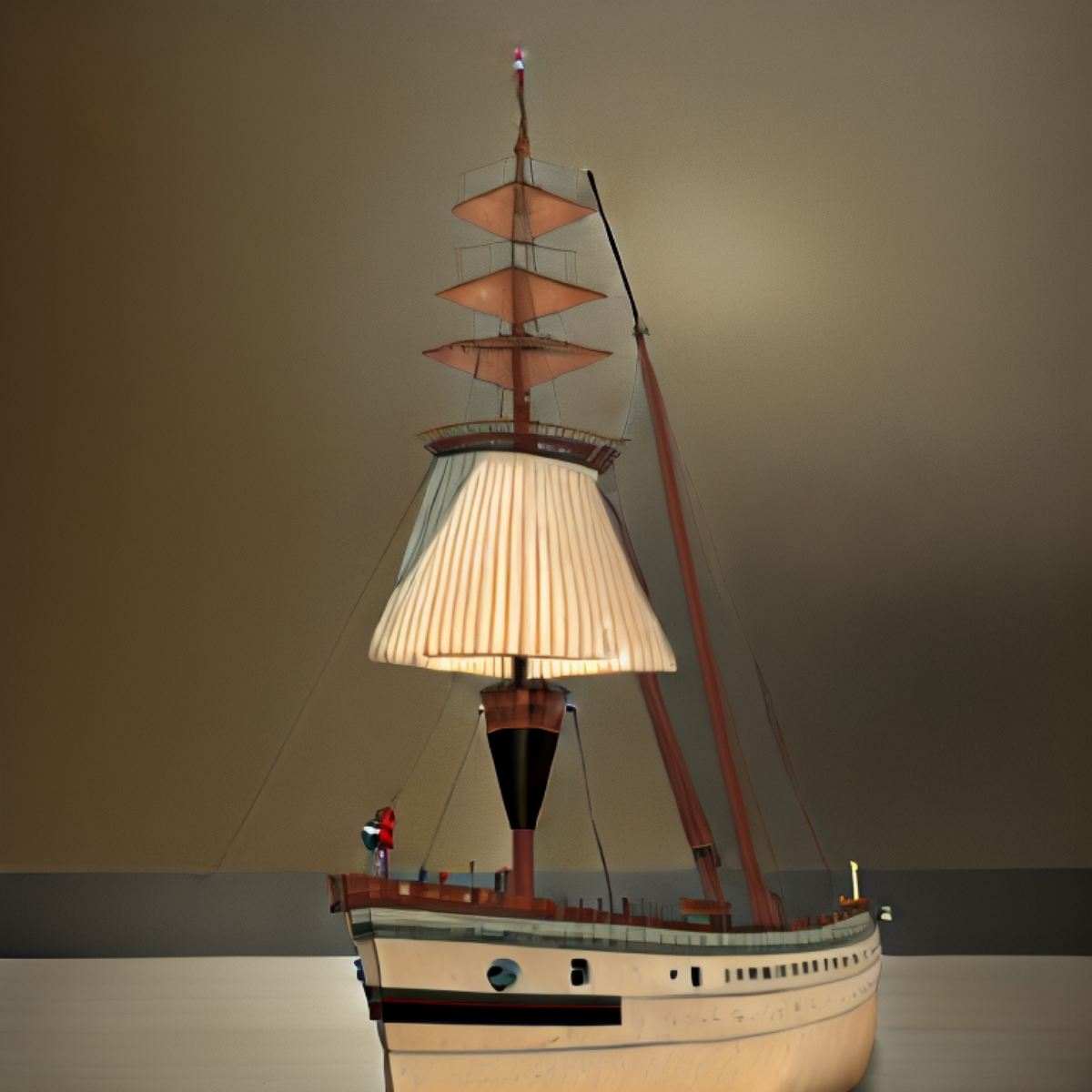} &
\includegraphics[trim=1cm 1cm 1cm 1cm,clip,width=0.2\linewidth]{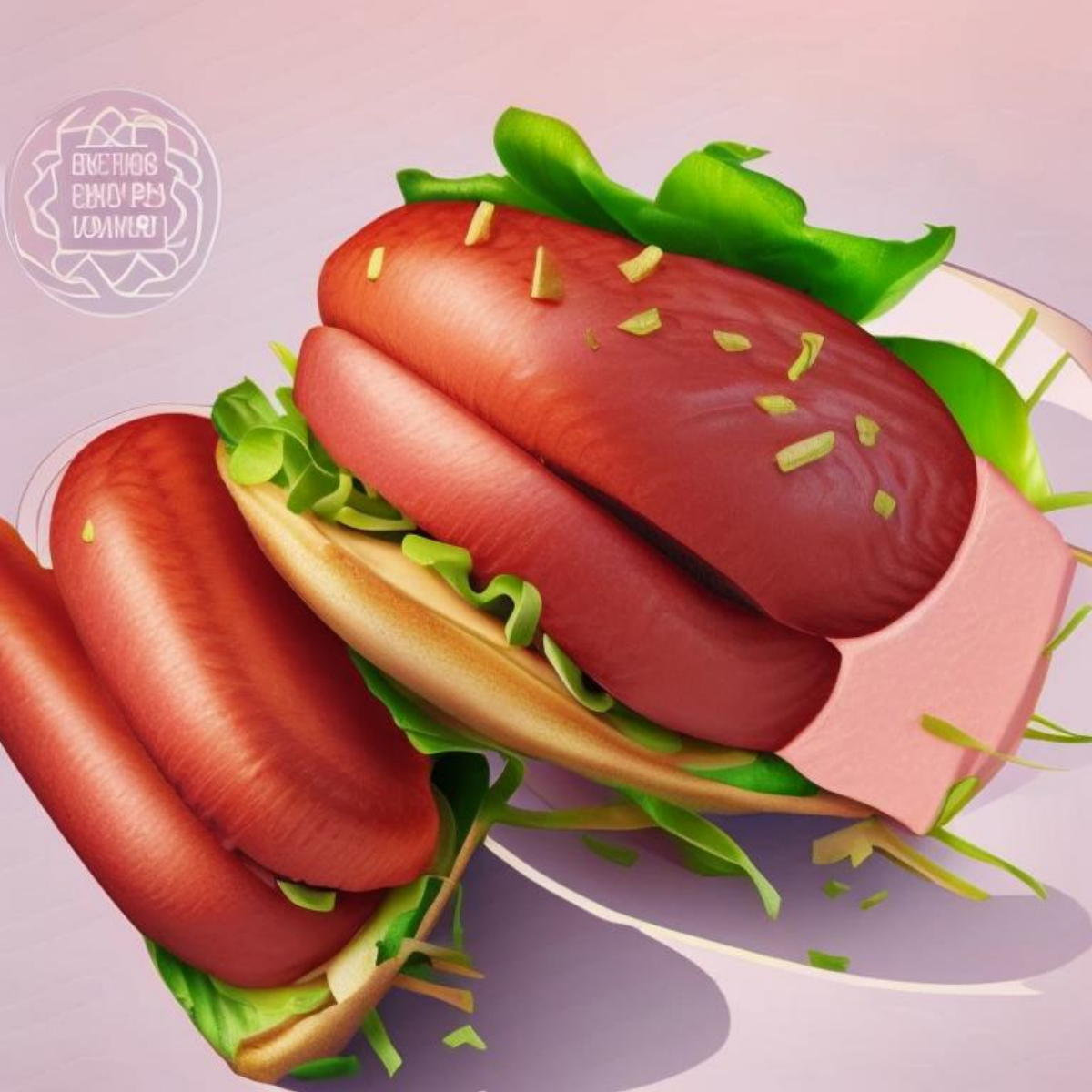} \\
\vspace{1mm} 
\small city-monument & \small burger-dog & \small burger-ice cream & \small ship-lamp & \small burger-strawberry  \\

\includegraphics[trim=1cm 1cm 1cm 1cm,clip,width=0.2\linewidth]{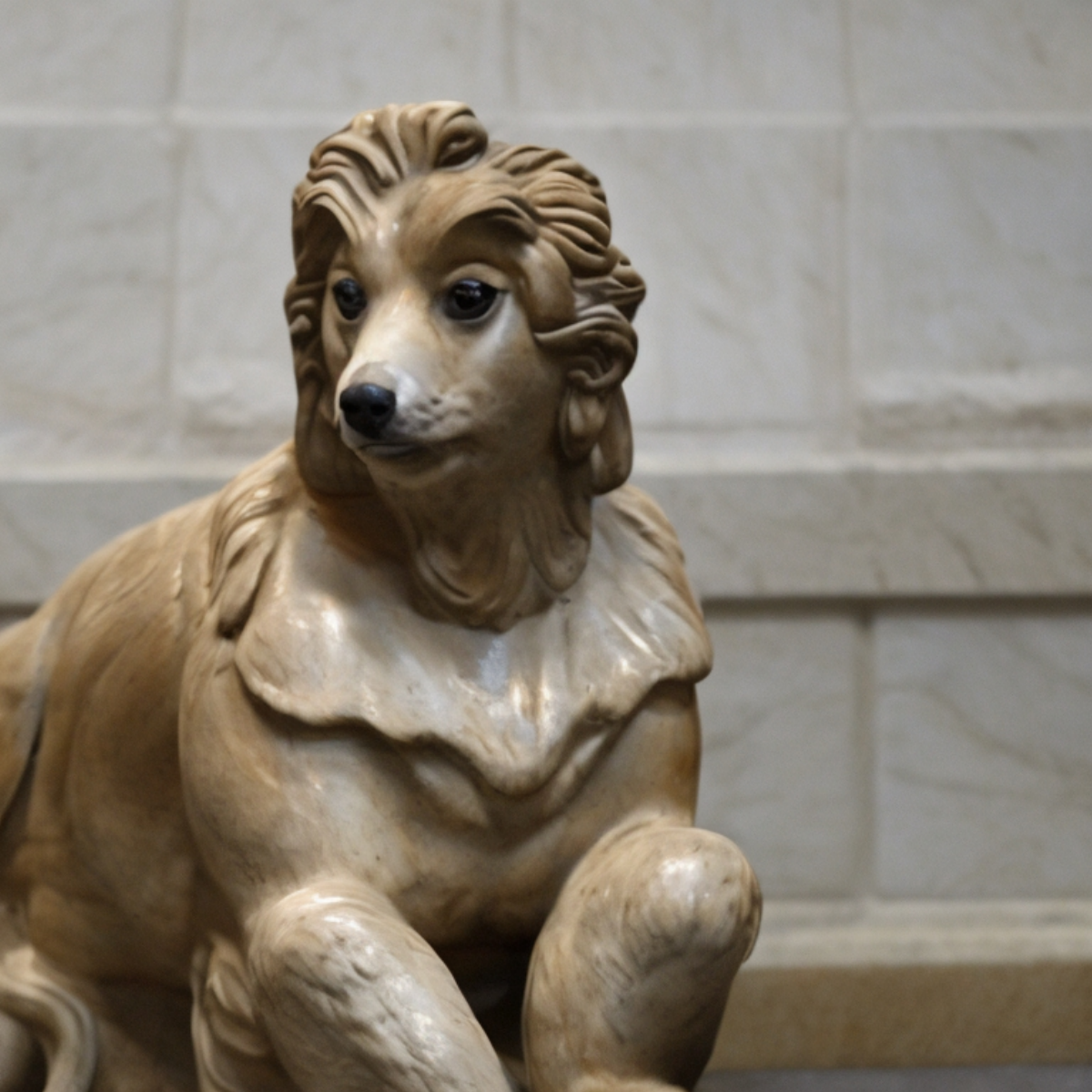} &
\includegraphics[trim=1cm 1cm 1cm 1cm,clip,width=0.2\linewidth]{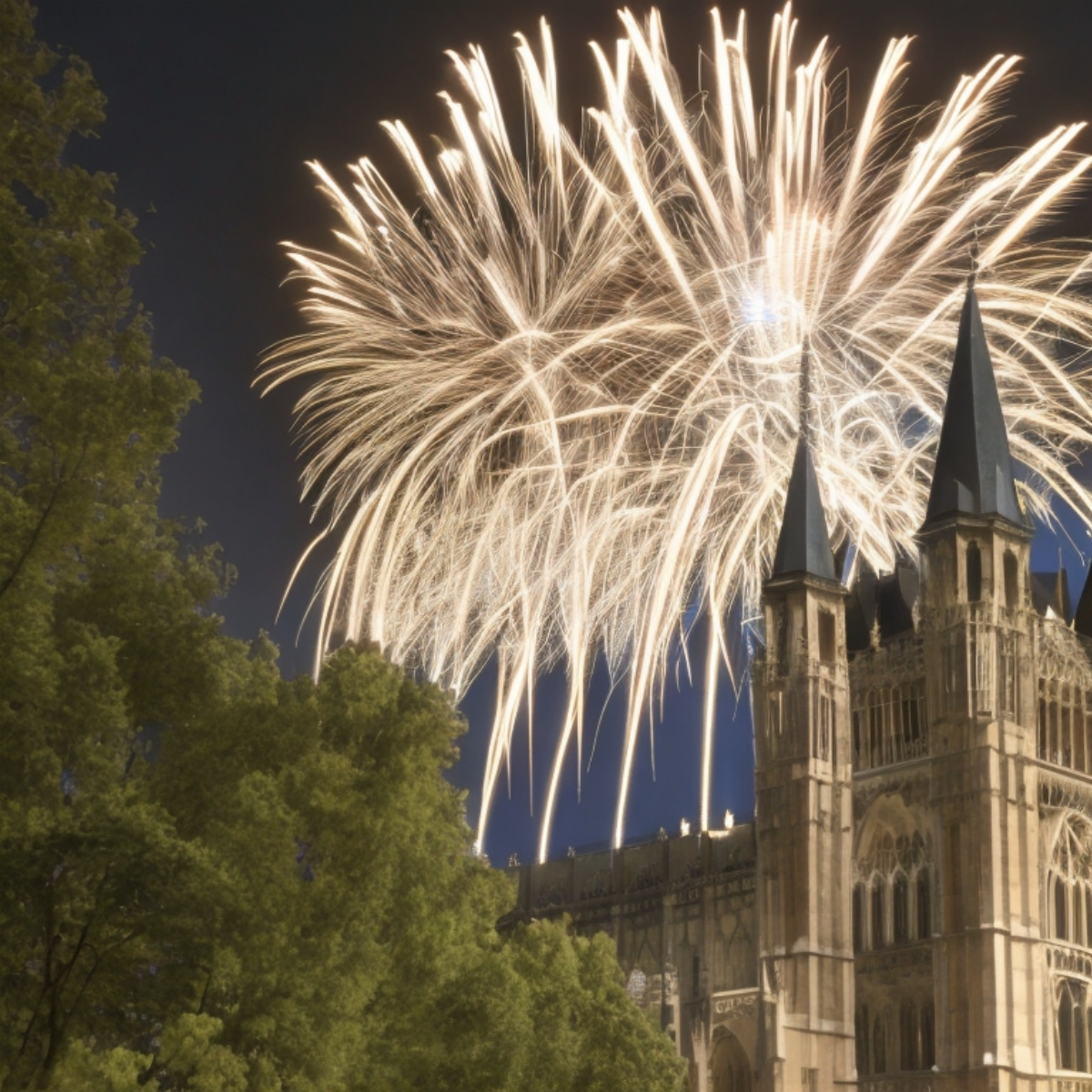} &
\includegraphics[trim=1cm 1cm 1cm 1cm,clip,width=0.2\linewidth]{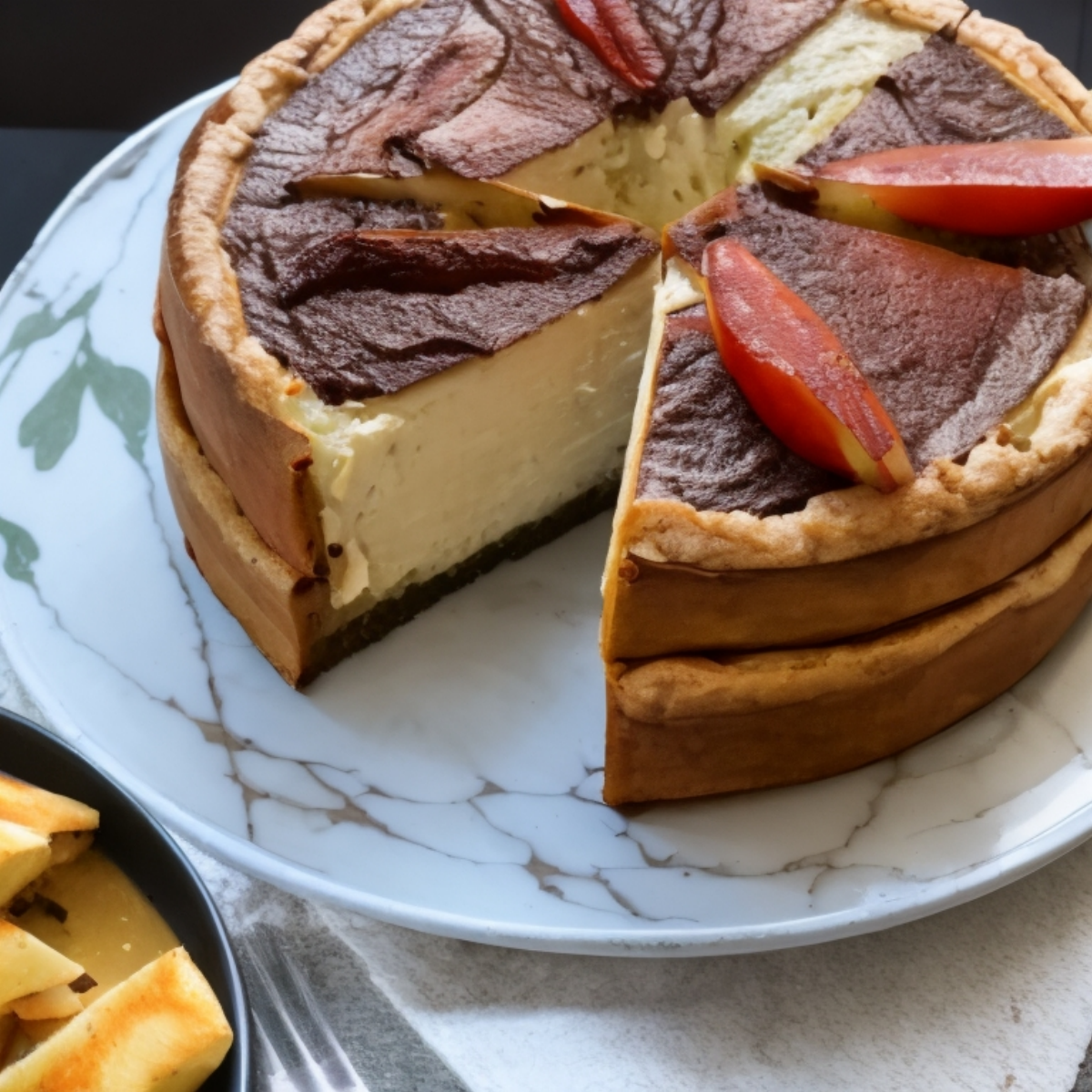} &
\includegraphics[trim=1cm 1cm 1cm 1cm,clip,width=0.2\linewidth]{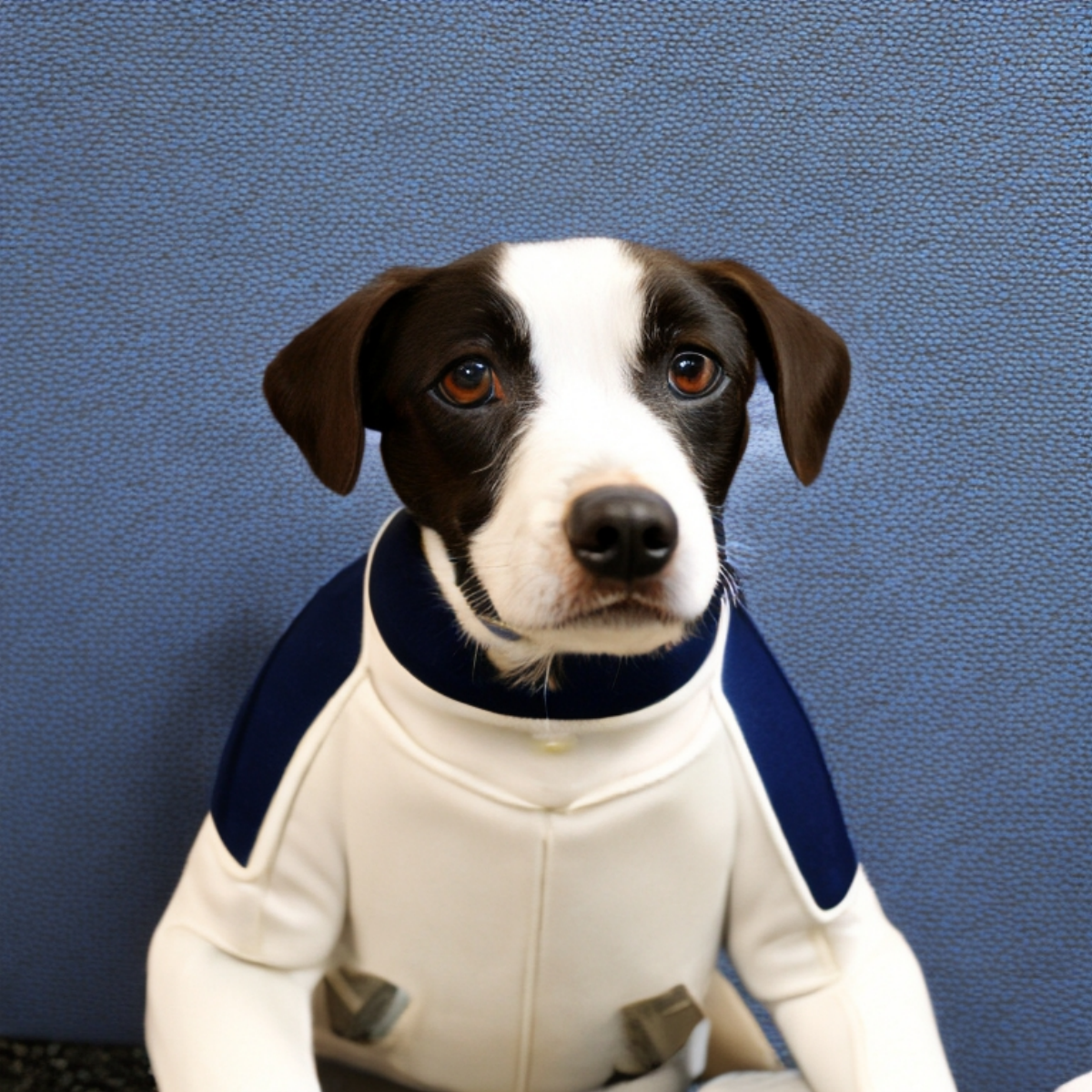} &
\includegraphics[trim=1cm 1cm 1cm 1cm,clip,width=0.2\linewidth]{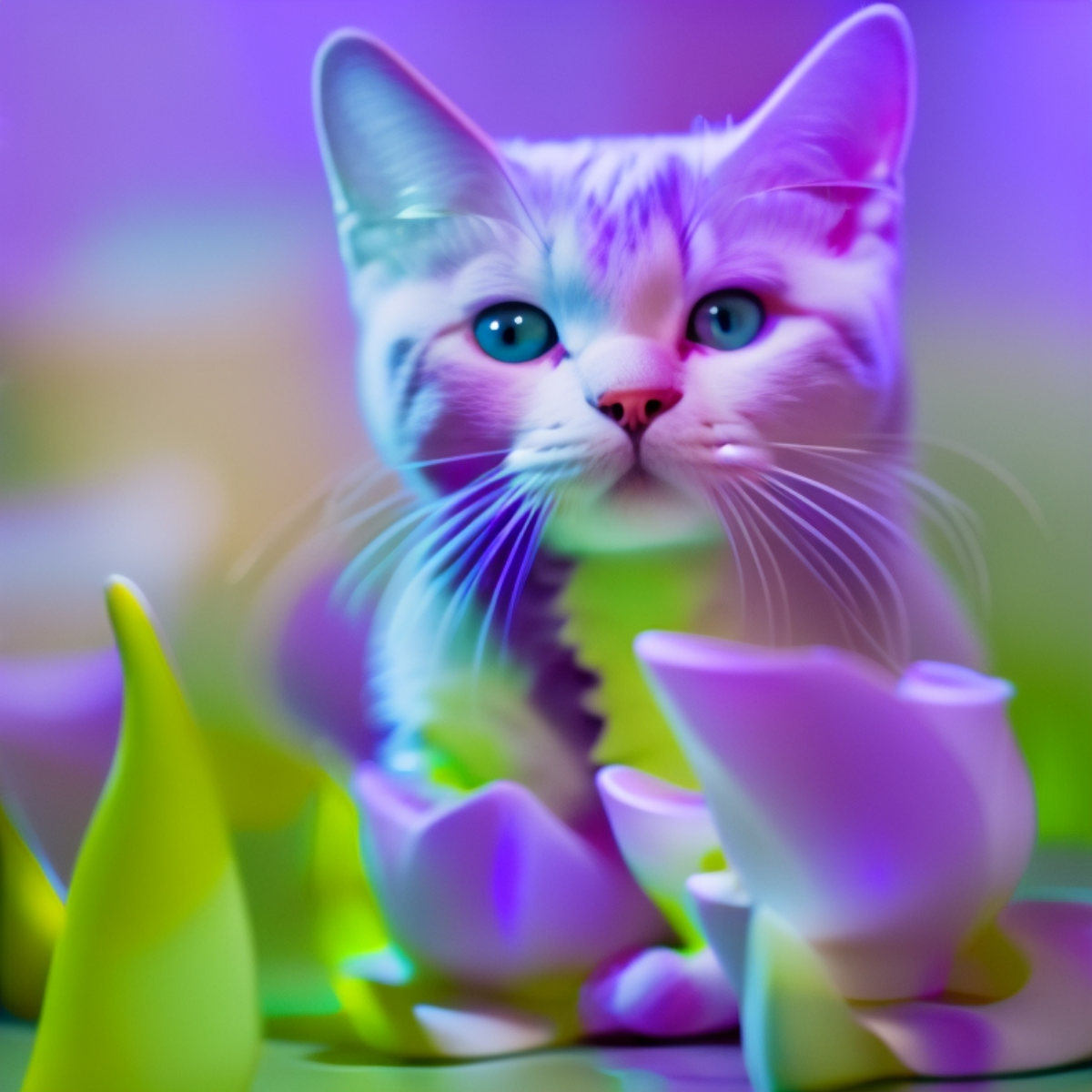} \\
\vspace{1mm} 
\small dog-Roman statue & \small cathedral-fireworks& \small pizza-tiramisu cake & \small dog-astronaut& \small cat-lavender  \\

\includegraphics[trim=1cm 1cm 1cm 1cm,clip,width=0.2\linewidth]{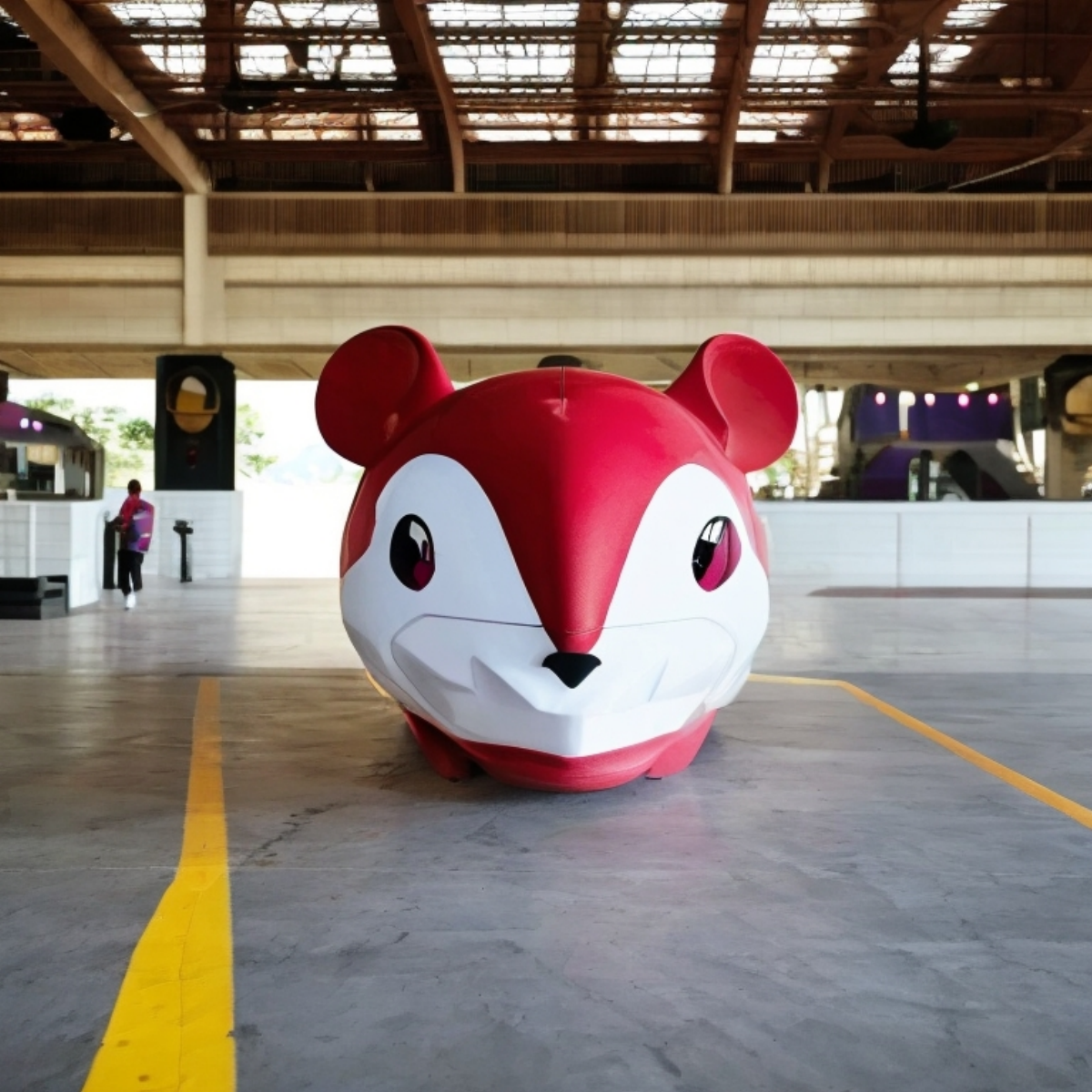} &
\includegraphics[trim=1cm 1cm 1cm 1cm,clip,width=0.2\linewidth]{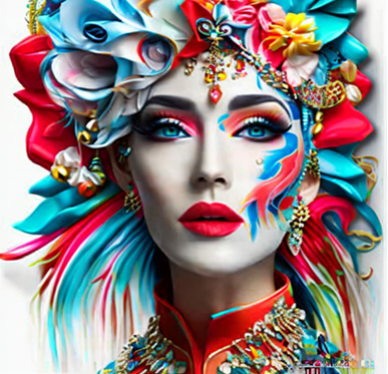} &
\includegraphics[trim=1cm 1cm 1cm 1cm,clip,width=0.2\linewidth]{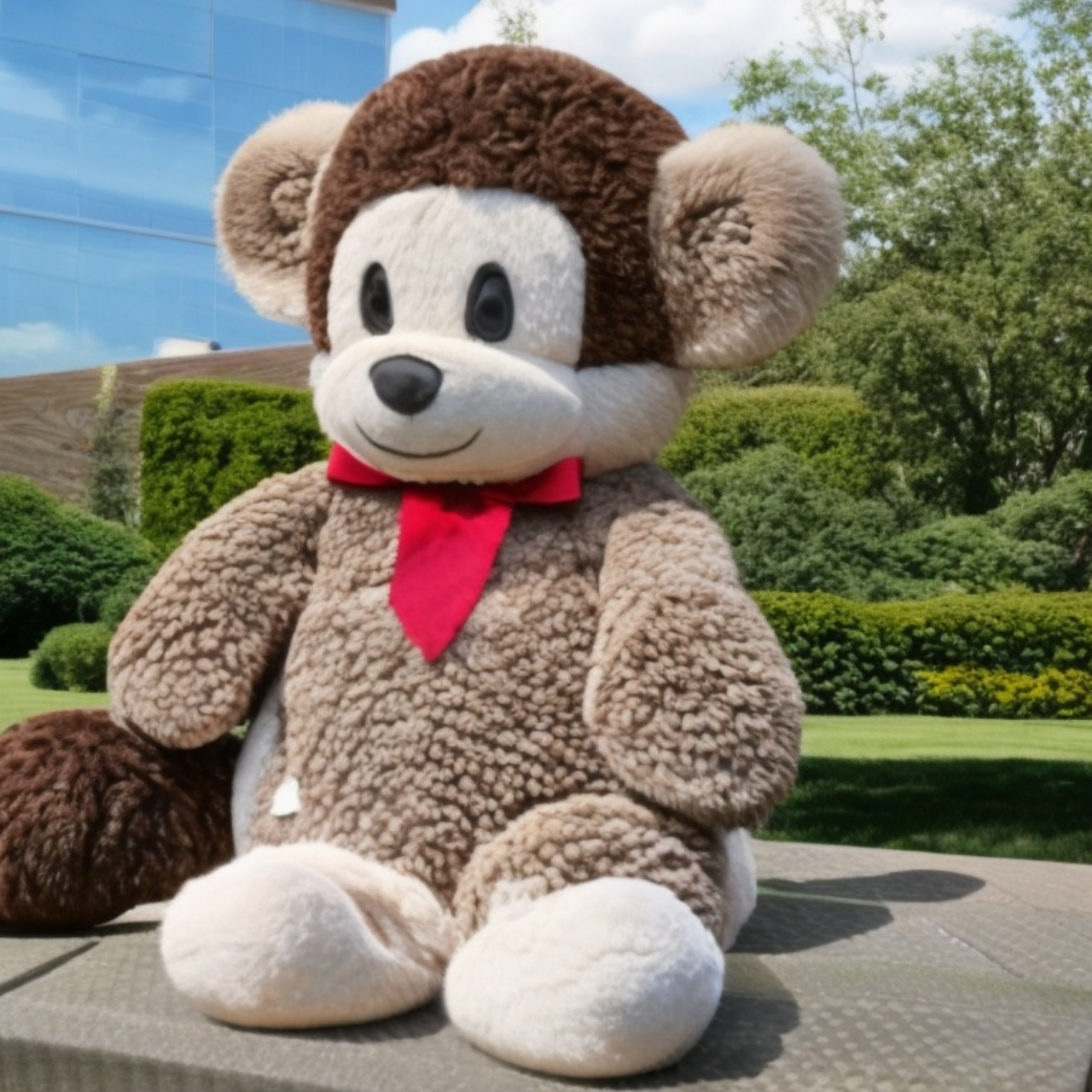} &
\includegraphics[trim=1cm 1cm 1cm 1cm,clip,width=0.2\linewidth]{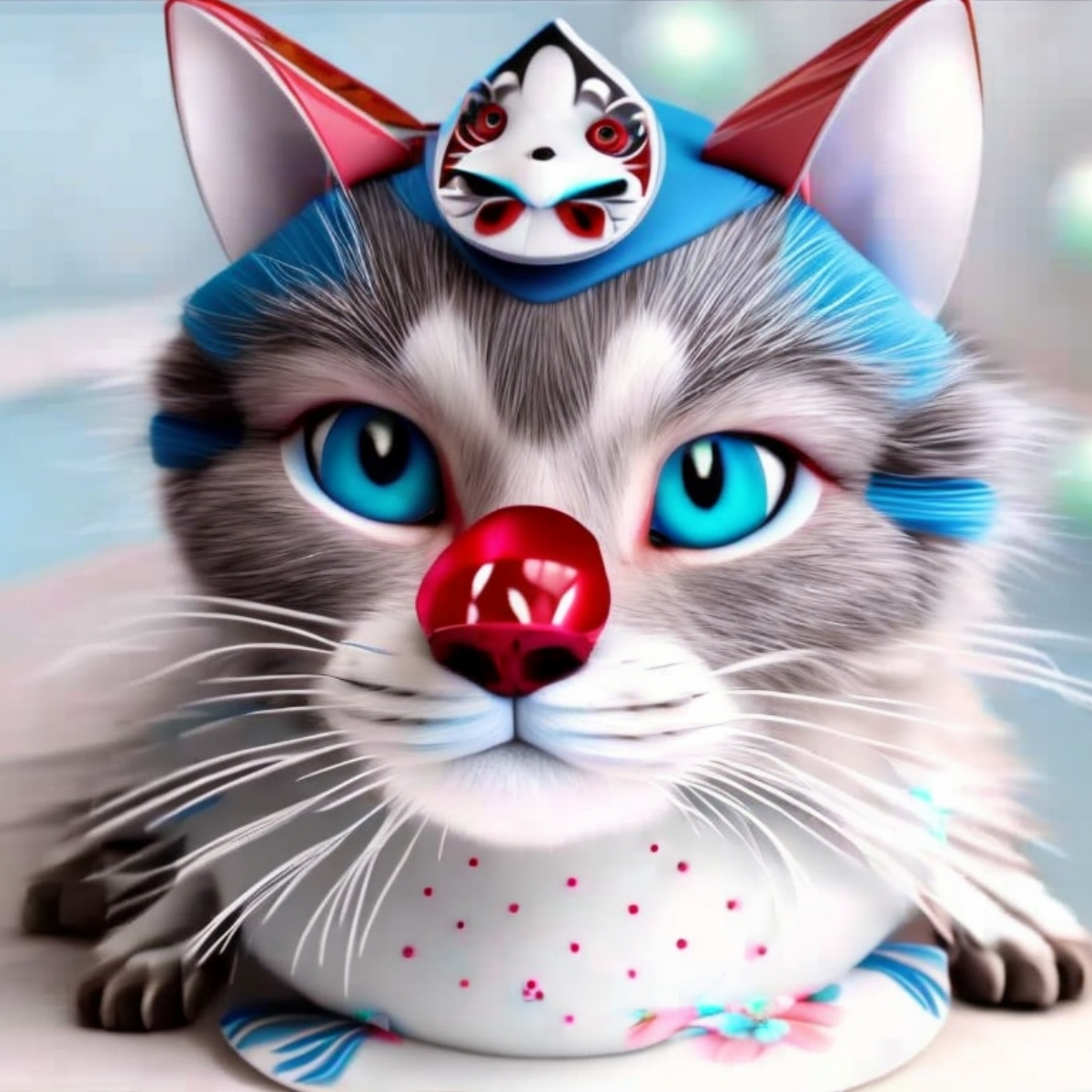} &
\includegraphics[trim=1cm 1cm 1cm 1cm,clip,width=0.2\linewidth]{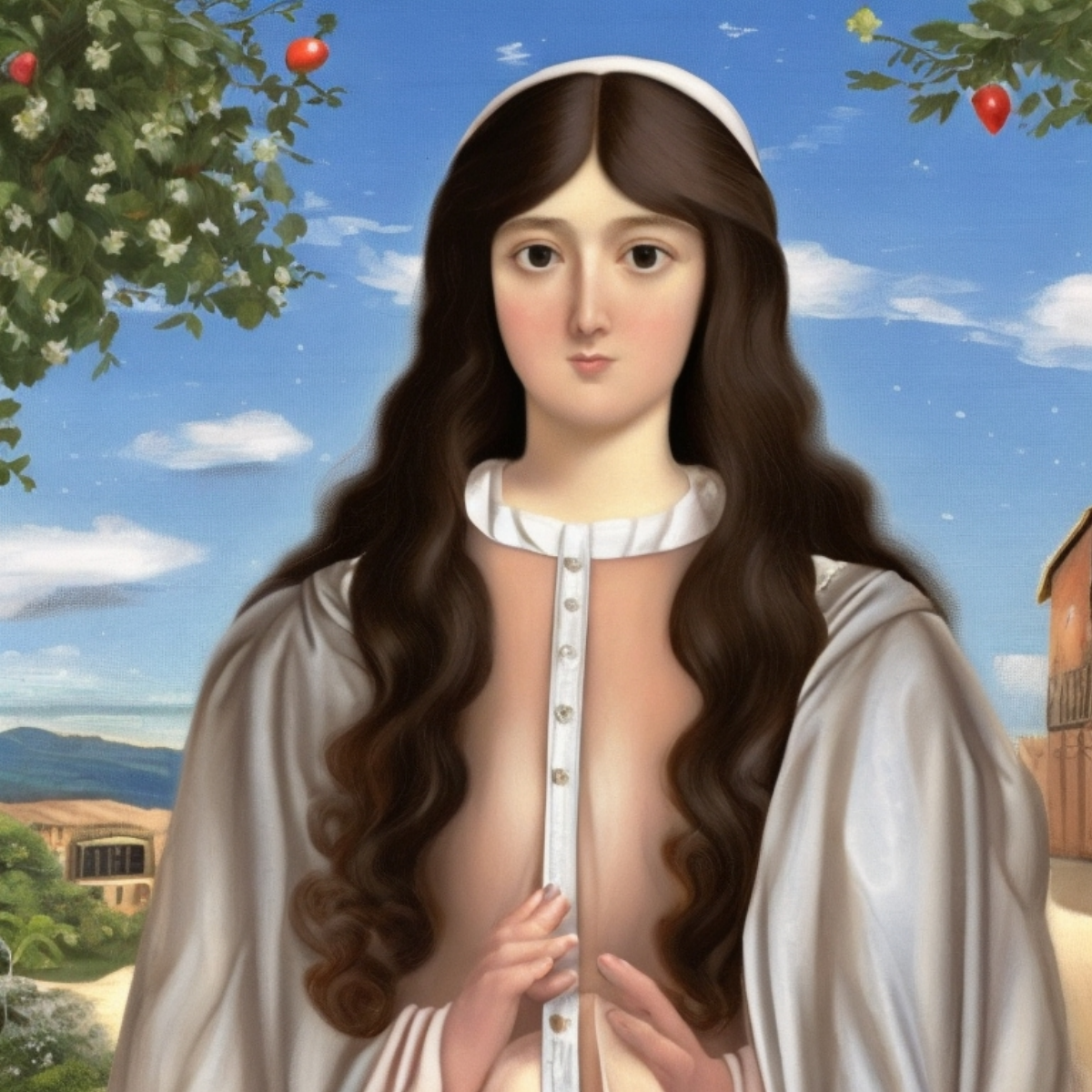} \\
\vspace{1mm} 
\small red ferrari-hello kitty & \small woman-opera player & \small teddy bear-girl & \small car-clown & \small anime-Renaissance    \\

\includegraphics[trim=1cm 1cm 1cm 1cm,clip,width=0.2\linewidth]{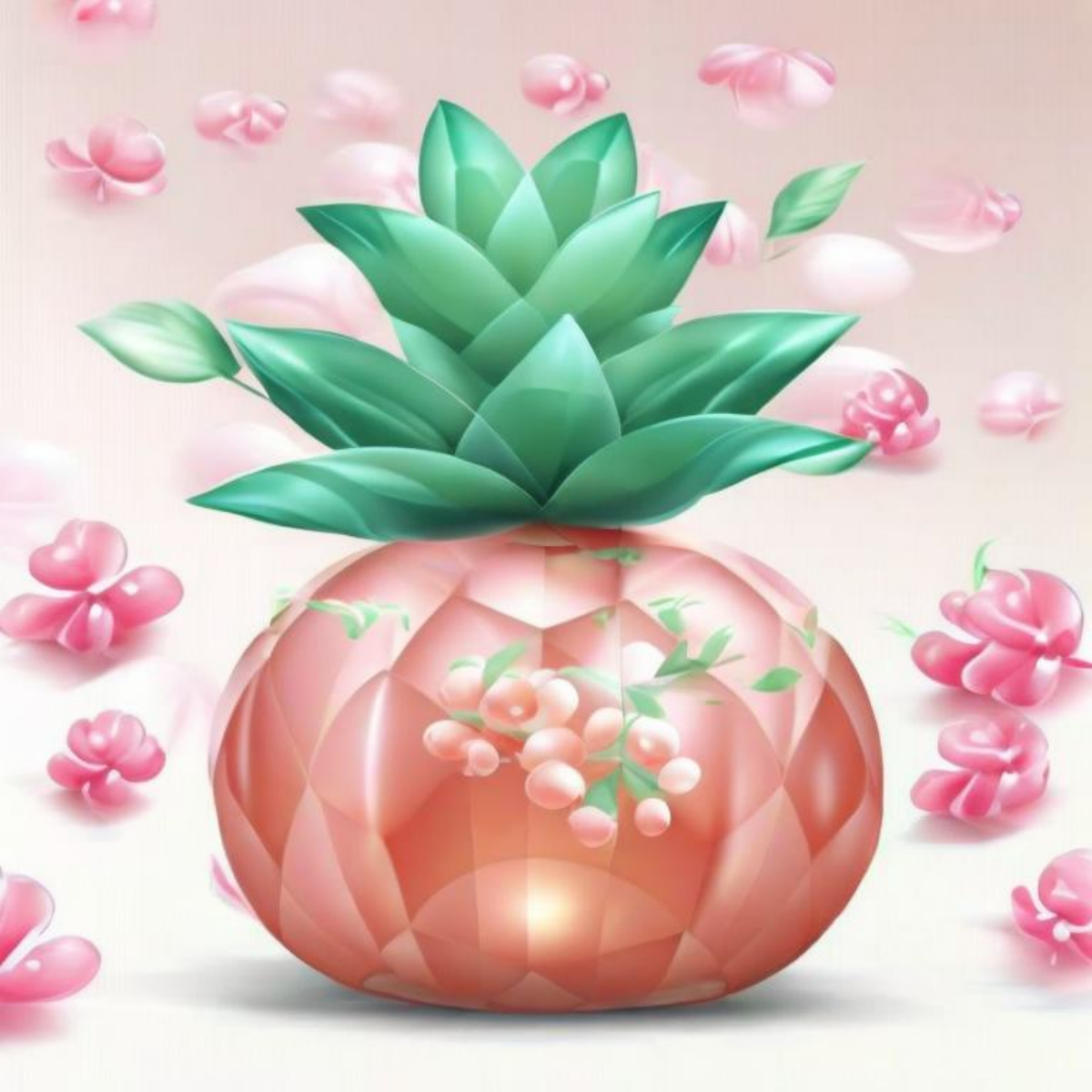} &
\includegraphics[trim=1cm 1cm 1cm 1cm,clip,width=0.2\linewidth]{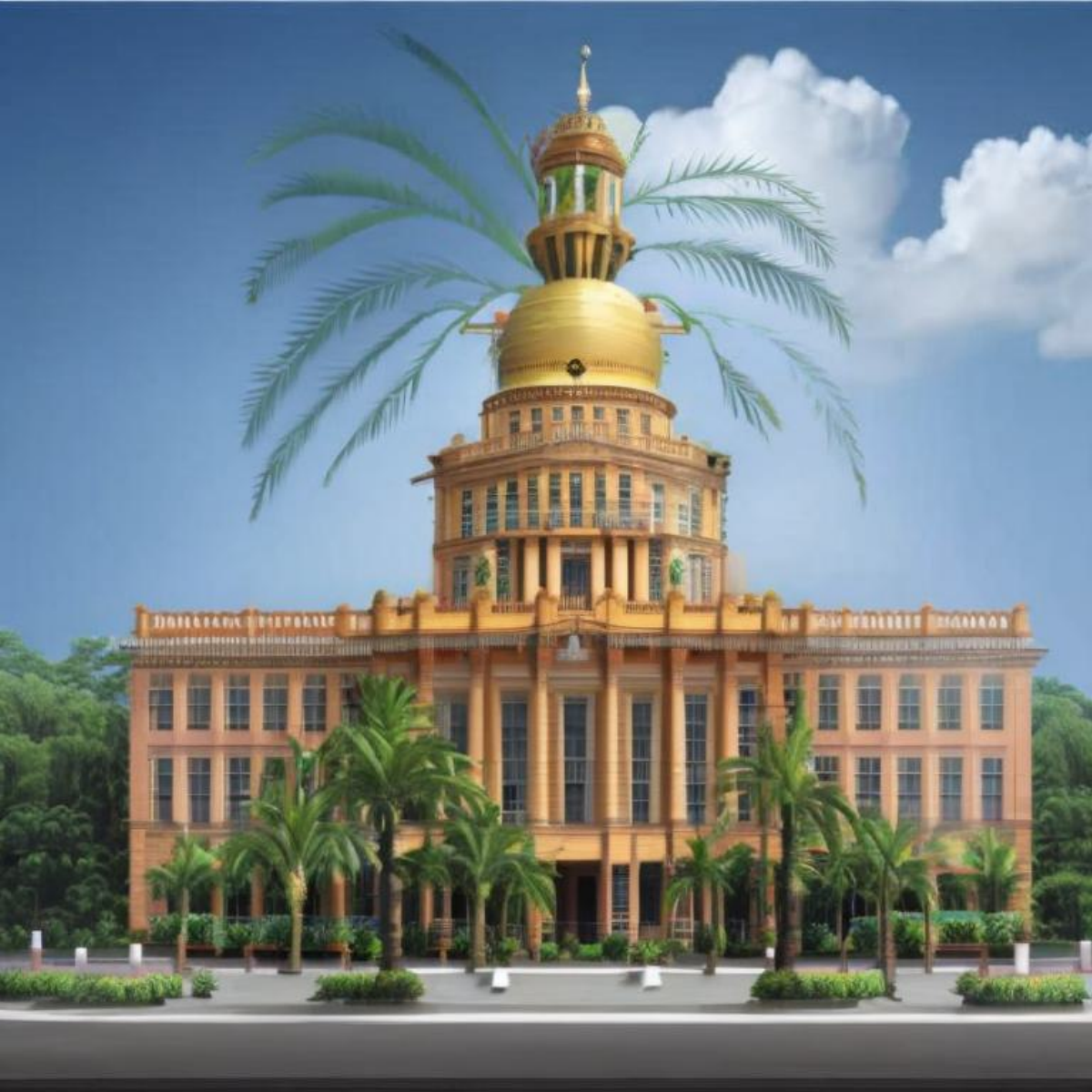} &
\includegraphics[trim=1cm 1cm 1cm 1cm,clip,width=0.2\linewidth]{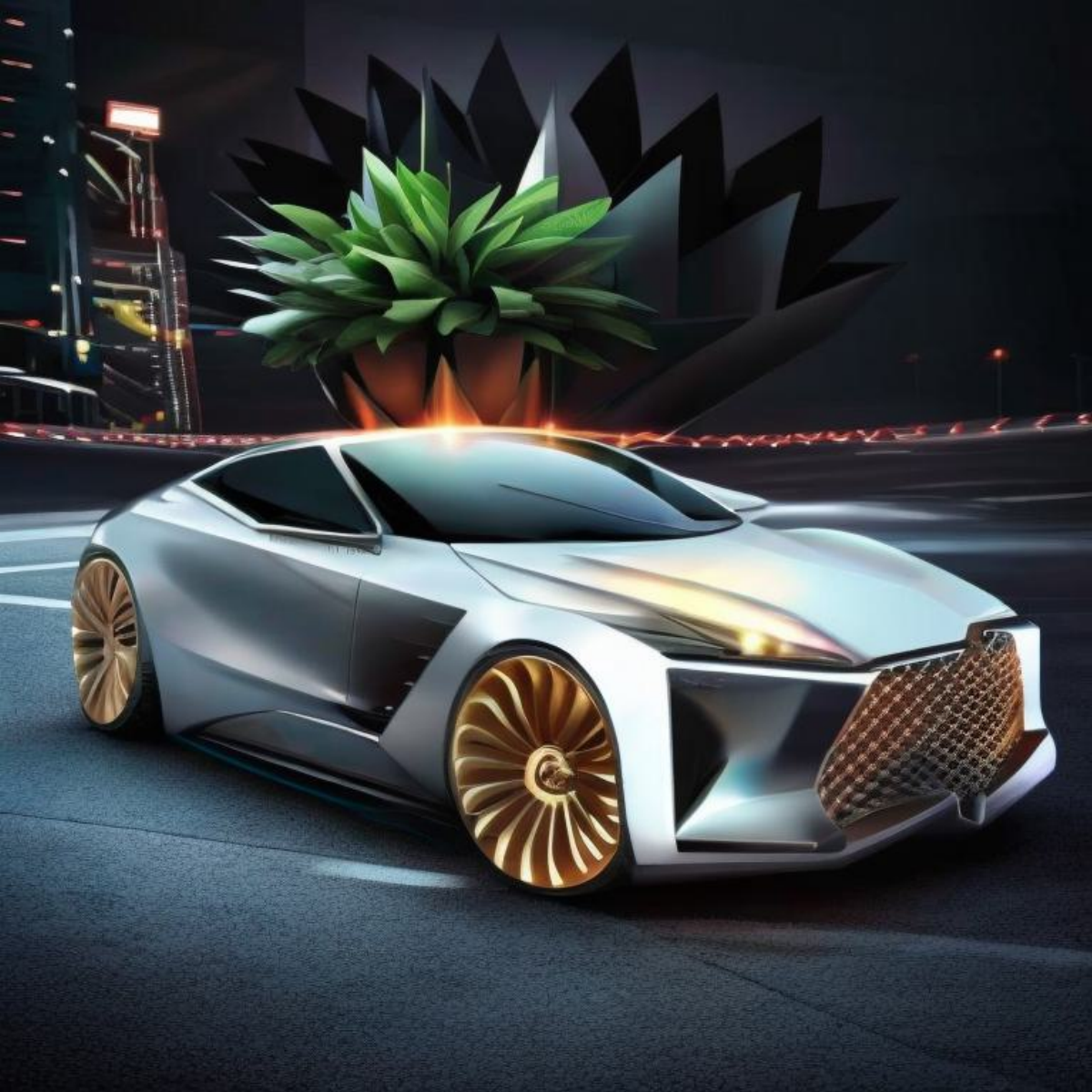} &
\includegraphics[trim=1cm 1cm 1cm 1cm,clip,width=0.2\linewidth]{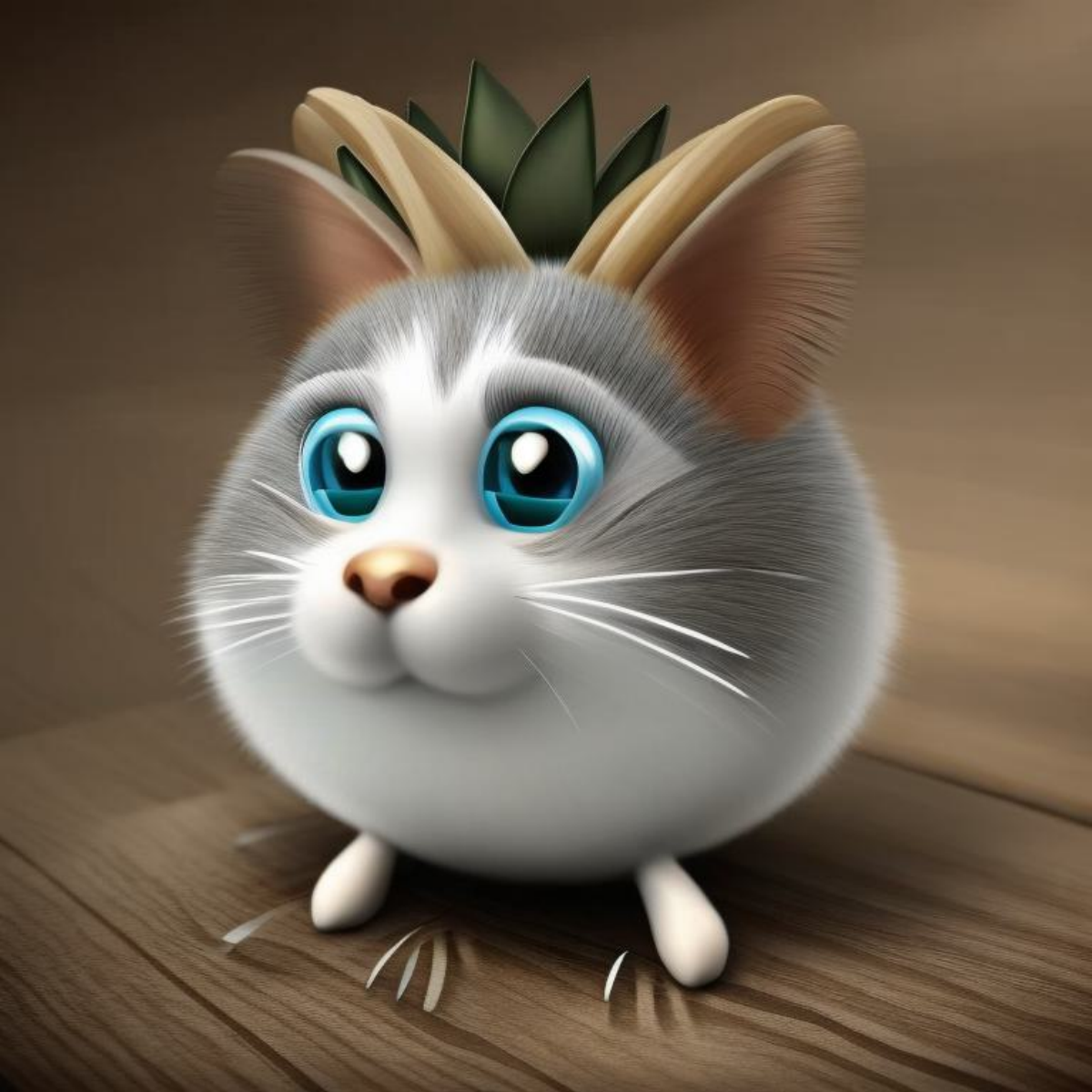} &
\includegraphics[trim=1cm 1cm 1cm 1cm,clip,width=0.2\linewidth]{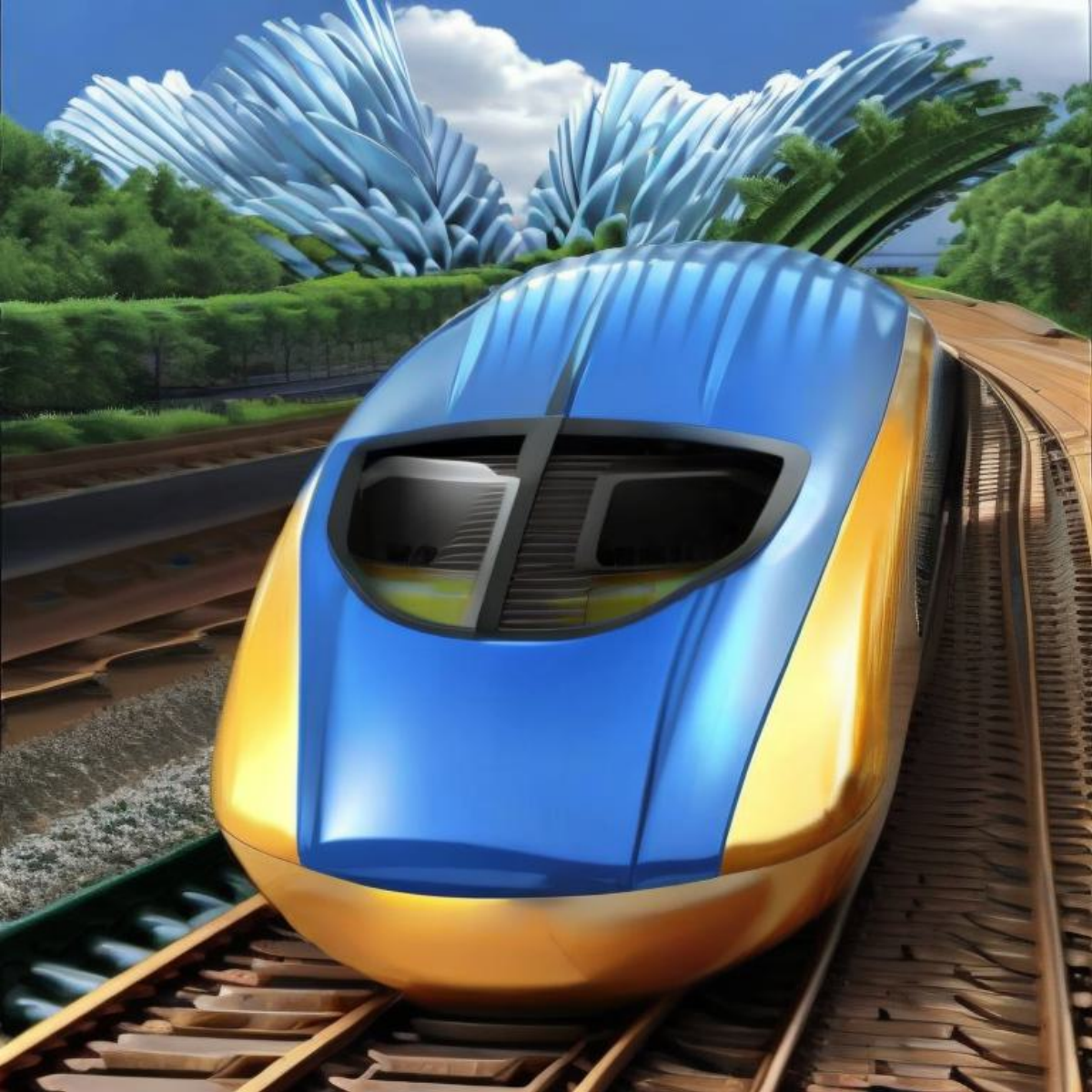} \\
\vspace{1mm} 
\small pineapple-blossom & \small pineapple-building & \small pineapple-car & \small pineapple-cat& \small pineapple-train  \\

\end{tabular}}
\caption{More blending results. Given two concepts, our approach can blend them into a novel object that the model has never seen before, generating high-quality images.}
\label{more_blending_results_2}
\end{figure*}

\Cref{more_blending_results} and \Cref{more_blending_results_2} present several of our blending results, highlighting the strong capabilities of our method for both concept blending and generation tasks.

\section{More Analyses of Blending Results}

Blending is a multifaceted process that occurs across various domains, each exhibiting distinct characteristics and behaviors. In the study of embedding blending, it is evident that different categories impose constraints on the resulting blends. 

\subsection{Results from Different Categories of Concepts}

\begin{table}[ht]
\caption{Mean scores for different categories. }
\centering
\begin{tabular}{ccccc}
\toprule
Category & CLIP-BS($\uparrow$) & DINO-BS($\uparrow$) & CLIP-IQA($\uparrow$) & HPS($\uparrow$) \\
\midrule
Animals & \cellcolor{red!40} \textbf{9.4410} & 0.2460 & 0.5160 & 0.2800 \\
Common objects & 7.9810 & \cellcolor{orange!40} 0.4160 & \cellcolor{orange!40} 0.5270 & \cellcolor{orange!40} 0.3000 \\
Nature & 9.2250 & \cellcolor{red!40} \textbf{0.4190} & 0.3330 & 0.2760 \\
Transports & \cellcolor{orange!40} 9.3100 & 0.2280 & \cellcolor{red!40} \textbf{0.5430} & \cellcolor{red!40} \textbf{0.3060} \\
\bottomrule
\end{tabular}
\label{cate1}
\end{table}

\begin{table}[ht]
\caption{Mean scores for blending category pairs. }
\centering
\begin{tabular}{ccccc}
\toprule
Category Pair & CLIP-BS($\uparrow$) & DINO-BS($\uparrow$) & CLIP-IQA($\uparrow$) & HPS($\uparrow$) \\
\midrule
Animals \& Common & 8.7110 & 0.3308 & 0.5212 & 0.2908 \\
Animals \& Nature & \cellcolor{orange!40} 9.3330 & \cellcolor{orange!40} 0.3323 & 0.4244 & 0.2780 \\
Animals \& Transports & \cellcolor{red!40} \textbf{9.3760} & 0.2369 & \cellcolor{orange!40} 0.5294 & \cellcolor{orange!40} 0.2930 \\
Common \& Nature & 8.6030 & \cellcolor{red!40} \textbf{0.4170} & 0.4298 & 0.2880 \\
Common \& Transports & 8.6450 & 0.3217 & \cellcolor{red!40} \textbf{0.5348} & \cellcolor{red!40} \textbf{0.3030} \\
Nature \& Transports & 9.2680 & 0.3232 & 0.4380 & 0.2910 \\
\bottomrule
\end{tabular}
\label{cate2}
\end{table}

\Cref{cate1} and \Cref{cate2} show the blending scores of four metrics for different category pairs. The CLIP-BS in \Cref{cate1} indicates that similar types of objects, such as animals, nature, and transport, achieve higher scores. However, for common objects, the score is lower due to the larger diversity within this category. Regarding DINO-BS, categories in nature tend to blend more successfully. In the case of HPS, we observe that transport categories generate higher-quality images. In \Cref{cate2}, the pattern is similar to that in \Cref{cate1}, where the best scores align with the original categories. Additionally, compared to nature categories, animals tend to blend better, which we hypothesize is due to biological reasons discussed below.

\subsection{Factors of Blending Concepts}

\textbf{Distance between Concepts. }Distance is a crucial factor that determines the effectiveness of the blending. This principle is illustrated through t-SNE analysis in \Cref{t_SNE_analysis_paragraph}, which demonstrates how the proximity of two concepts within an embedding space influences their ability to blend. When two concepts are close to each other in this space, they blend easily, forming a cohesive blend. However, as the distance between the concepts increases, the blending becomes less seamless. For example, concepts that are close in proximity, such as cats and dogs, tend to blend effortlessly, with the final output lying somewhere between both concepts. When concepts are moderately distant, such as a cactus and a rose, partial blending occurs, with the resulting image leaning more towards one concept while still maintaining some elements of the other. As the distance between concepts becomes even greater, such as in the case of an airplane and a bird, the blending becomes limited, and the final result often favors one concept, with the other remaining a secondary influence. 

\section{Limitations and Future Work}
\label{limitation_sec}

In the case of the Stable Diffusion model, the bias towards different concepts requires adjusting the parameter $\gamma$, a process that is cumbersome and inefficient. Moreover, when blending three or more concepts, it becomes challenging because the UNET architecture can only support two conditions. Therefore, our goal is to explore more stable, cost-effective, and training-free generation methods, while also seeking a more structured approach for blending multiple concepts.

\section{Impact Statement}
\label{impact_statement}
This paper presents a contribution aimed at advancing the field of concept blending. We explicitly state that our work is intended for academic research and commercial applications, provided such uses are authorized by the author. However, we strongly emphasize that we do not condone, nor do we support, the use of our research to generate harmful, unethical, or unlawful content. The generation of unsafe or offensive images that violate ethical standards, legal regulations, or societal norms is strictly prohibited. We use the Safe Checker from Stable Diffusion~\citep{rombach2022high} to prevent the generation of unsafe content. As researchers, we believe it is our responsibility to ensure that our work is used for the betterment of society and in a manner that aligns with fundamental ethical principles. 

\newpage
\newpage

\end{document}